\documentclass{article}

\usepackage{hello}

\usepackage[margin=1in]{geometry}

\usepackage{graphicx}
\usepackage{booktabs}
\usepackage{array}
\usepackage{amsmath}
\usepackage{amssymb}
\usepackage{amsfonts}
\usepackage{multirow}
\usepackage{verbatim}
\usepackage{caption}
\usepackage{longtable}
\usepackage{supertabular}
\usepackage{CJKutf8}
\usepackage[utf8]{inputenc} % optional
\usepackage[T1]{fontenc}
\usepackage[french,vietnamese,mongolian,greek,english]{babel}

\usepackage{enumitem}
\usepackage{tablefootnote}
\usepackage[round,semicolon]{natbib}
\usepackage{colortbl}
\usepackage{xspace}
\usepackage{textcomp}
\usepackage{makecell}
\usepackage{lscape} 
\usepackage{siunitx}
\usepackage{listings}
\usepackage{xcolor}
\definecolor{dt}{gray}{0.7}
\usepackage{pifont}       % \ding{xx}
\usepackage{bbding}       % \Checkmark,\XSolid,... (需要和pifont宏包共同使用)
\usepackage{fontawesome}

\usepackage{scrextend}

\usepackage{tgpagella}
\usepackage{latexsym}
\usepackage[T1]{fontenc}
\usepackage[utf8]{inputenc}
\usepackage{microtype}
\definecolor{mydarkblue}{rgb}{0,0.08,0.45}
\definecolor{citecolor}{HTML}{0071BC}
\usepackage[backref=page,pagebackref=true,colorlinks,citecolor=citecolor,urlcolor=magenta,linkcolor=magenta]{hyperref}
\usepackage{url}            % simple URL typesetting
\usepackage{nicefrac}       % compact symbols for 1/2, etc.
\usepackage{changepage}
\usepackage{xargs}          % Use more than one optional parameter in a new commands
\usepackage{wrapfig,lipsum,booktabs}
\usepackage{longtable}
\usepackage{subcaption}
\usepackage{endnotes}

\usepackage{pgfplots}
\usetikzlibrary{pgfplots.groupplots}
\pgfplotsset{compat=1.3}
\usepackage{tikz}
\usetikzlibrary{patterns}

\usepackage[most]{tcolorbox}

\usepackage[capitalize,noabbrev]{cleveref}
\crefname{section}{Section}{\S\S}
\Crefname{section}{Section}{\S\S}
\crefname{table}{Table}{Tables}
\crefname{figure}{Figure}{Figures}
\crefname{algorithm}{Algorithm}{}
\crefname{equation}{eq.}{}
\crefname{appendix}{Appendix}{}
\crefformat{section}{Section #2#1#3}
\usepackage{multicol}
\usepackage{tcolorbox}

\usepackage{titlesec}
\titleformat*{\section}{\large\bfseries}

\definecolor{battleshipgrey}{rgb}{0.3, 0.3, 0.3}
\definecolor{brilliantrose}{rgb}{1.0, 0.33, 0.64}
\definecolor{americanrose}{rgb}{1.0, 0.01, 0.24}
\definecolor{jweigreen}{rgb}{0,0.45,0.24}
\definecolor{bluegray}{rgb}{0.1, 0.1, 0.4}
\definecolor{ao(english)}{rgb}{0.0, 0.5, 0.0}
\definecolor{blanchedalmond}{rgb}{1.0, 0.92, 0.8}
\definecolor{atomictangerine}{rgb}{1.0, 0.6, 0.4}
\definecolor{chocolate(web)}{rgb}{0.82, 0.41, 0.12}
\definecolor{bananayellow}{rgb}{1.0, 0.88, 0.21}
\definecolor{goldenbrown}{rgb}{0.6, 0.4, 0.08}
\definecolor{aliceblue}{rgb}{0.94, 0.97, 1.0}
\definecolor{beige}{rgb}{0.96, 0.96, 0.86}
\definecolor{babyblue}{rgb}{0.54, 0.81, 0.94}
\definecolor{camel}{rgb}{0.76, 0.6, 0.42}
\definecolor{cinnamon}{rgb}{0.82, 0.41, 0.12}
\definecolor{deepskyblue}{rgb}{0.0, 0.75, 1.0}
\definecolor{frenchblue}{rgb}{0.0, 0.45, 0.73}
\definecolor{classicrose}{rgb}{0.98, 0.8, 0.91}
\definecolor{frenchrose}{rgb}{0.96, 0.29, 0.54}
\definecolor{frenchlilac}{rgb}{0.53, 0.38, 0.56}
\definecolor{frenchbeige}{rgb}{0.65, 0.48, 0.36}
\definecolor{verylightgreen}{RGB}{240, 255, 235}
\definecolor{verylightred}{RGB}{255, 235, 235}
\definecolor{verylightyellow}{RGB}{255, 254, 235}
\definecolor{dt}{gray}{0.7}

\definecolor{forestgreen}{HTML}{2e7d43}
\definecolor{color1}{HTML}{FF9999}
\definecolor{color2}{HTML}{FF6666}
\definecolor{color3}{HTML}{FF3333}
\definecolor{color4}{HTML}{E60000}
\definecolor{color5}{HTML}{B30000}
\definecolor{color6}{HTML}{8CD98C}
\definecolor{color7}{HTML}{53c653}
\definecolor{color8}{HTML}{39ac39}
\definecolor{color9}{HTML}{2d862d}
\definecolor{color10}{HTML}{206020}
\definecolor{color11}{HTML}{cca300}

\usepackage{minitoc}
\usepackage{float}
\newlength\savewidth

\usepackage{amsmath,amsfonts,bm}

\def\eqref#1{equation~\ref{#1}}
\def\1{\bm{1}}

\DeclareMathAlphabet{\mathsfit}{\encodingdefault}{\sfdefault}{m}{sl}
\SetMathAlphabet{\mathsfit}{bold}{\encodingdefault}{\sfdefault}{bx}{n}

\title{
\textbf{
Qwen2-VL: Enhancing Vision-Language Model's Perception of the World at Any Resolution}
}

\author{
\normalsize{}
Peng Wang*\hspace{3mm} 
Shuai Bai*\hspace{3mm} 
Sinan Tan*\hspace{3mm} 
Shijie Wang*\hspace{3mm} 
Zhihao Fan*\hspace{3mm} 
Jinze Bai*$^{\dag}$\hspace{3mm} 
\\
\normalsize{}
Keqin Chen\hspace{3mm} 
Xuejing Liu\hspace{3mm} 
Jialin Wang\hspace{3mm} 
Wenbin Ge\hspace{3mm} 
Yang Fan\hspace{3mm} 
Kai Dang\hspace{3mm} 
Mengfei Du\hspace{3mm} 
\\
\normalsize{}
Xuancheng Ren\hspace{3mm}
Rui Men\hspace{3mm}
Dayiheng Liu\hspace{3mm}
Chang Zhou\hspace{3mm}
Jingren Zhou\hspace{3mm}
Junyang Lin$^{\dag}$
\\ 
\vspace{7mm}
\normalsize{}
\textbf{Qwen Team\hspace{3mm}Alibaba Group}
\vspace{-7mm}
}

\date{}

\begin{document}

\doparttoc % Tell to minitoc to generate a toc for the parts
\faketableofcontents % Run a fake tableofcontents command for the partocs

\maketitle

\begin{abstract}
\noindent
We present the Qwen2-VL Series, an advanced upgrade of the previous Qwen-VL models that redefines the conventional predetermined-resolution approach in visual processing. 
Qwen2-VL introduces the Naive Dynamic Resolution mechanism, which enables the model to dynamically process images of varying resolutions into different numbers of visual tokens. 
This approach allows the model to generate more efficient and accurate visual representations, closely aligning with human perceptual processes. 
The model also integrates Multimodal Rotary Position Embedding (M-RoPE), facilitating the effective fusion of positional information across text, images, and videos.  
We employ a unified paradigm for processing both images and videos, enhancing the model's visual perception capabilities. 
To explore the potential of large multimodal models, Qwen2-VL investigates the scaling laws for large vision-language models (LVLMs). 
By scaling both the model size-with versions at 2B, 8B, and 72B parameters-and the amount of training data, the Qwen2-VL Series achieves highly competitive performance.
Notably, the Qwen2-VL-72B model achieves results comparable to leading models such as GPT-4o and Claude3.5-Sonnet across various multimodal benchmarks, outperforming other generalist models. Code is available at \url{https://github.com/QwenLM/Qwen2-VL}.

\end{abstract}

{\let\thefootnote\relax\footnotetext{$^*$Equal core contribution, $^\dag$Corresponding author}}

\section{Introduction}
In the realm of artificial intelligence, Large Vision-Language Models (LVLMs) represent a significant leap forward, building upon the strong textual processing capabilities of traditional large language models. These advanced models now encompass the ability to interpret and analyze a broader spectrum of data, including images, audio, and video. This expansion of capabilities has transformed LVLMs into indispensable tools for tackling a variety of real-world challenges. Recognized for their unique capacity to condense extensive and intricate knowledge into functional representations, LVLMs are paving the way for more comprehensive cognitive systems. 
By integrating diverse data forms, LVLMs aim to more closely mimic the nuanced ways in which humans perceive and interact with their environment. This allows these models to provide a more accurate representation of how we engage with and perceive our environment

% Recent advancements in the field have seen substantial improvements in LVLMs over a relatively short period \citep{blip2,llava,instructblip,minigpt-4,kosmos,qwenvl,llava1.5,cogvlm,gpt4v,gemini}. These improvements largely follow the \textit{visual encoder$\rightarrow$cross-modal connector$\rightarrow$LLM} \citep{gpt4,llama,llama2,vicuna2023,qwen} paradigm, leveraging next-token prediction training objectives alongside the introduction of high-quality datasets \citep{llava1.5,internlm-xcomposer,sharegpt4v,vlit}, larger model architectures \citep{flamingo}, higher image resolution \citep{otterhd,monkey}, mixture-of-experts (MoE) approaches \citep{cogvlm,mplug-owl2}, model ensembles \citep{sphinx}, and increasingly complex cross-modal connectors \citep{mplug-owl}, among other techniques. These advancements have notably improved the performance of LVLMs in handling complex visual and textual data.

Recent advancements in large vision-language models (LVLMs)~\citep{blip2,llava,instructblip,minigpt-4,kosmos,qwenvl,llava1.5,cogvlm,gpt4v,gemini} have led to significant improvements in a short span. These models~\citep{gpt4,llama,llama2,vicuna2023,qwen} generally follow a common approach of \textit{visual encoder$\rightarrow$cross-modal connector$\rightarrow$LLM}. This setup, combined with next-token prediction as the primary training method and the availability of high-quality datasets~\citep{llava1.5,internlm-xcomposer,sharegpt4v,vlit}, has driven much of the progress. Additional factors like larger model architectures~\citep{flamingo}, higher-resolution images~\citep{otterhd,monkey}, and advanced techniques such as mixture-of-expert models (MoE)~\citep{cogvlm,mplug-owl2}, model ensembles~\citep{sphinx}, and more sophisticated connectors~\citep{mplug-owl} between visual and textual modalities have also played a key role in enhancing LVLMs' ability to process complex visual and textual information more effectively.

However, current large vision-language models (LVLMs) are typically constrained by a fixed image input size. Standard LVLMs encode input images to a fixed resolution (e.g., 224×224), often by either downsampling or upsampling the images \citep{minigpt-4, kosmos}, or by employing a scale-then-padding approach \citep{llava, llava1.5}. While this one-size-fits-all strategy enables processing of images at consistent resolutions, it also limits the model’s ability to capture information at different scales, particularly leading to a significant loss of detailed information in high-resolution images. Consequently, such models fall short of perceiving visual information with the same sensitivity to scale and detail as human vision.

Additionally, most LVLMs rely on a static, frozen CLIP-style \citep{clip} vision encoder, raising concerns about whether the visual representations produced by such pre-trained models are adequate, particularly for complex reasoning tasks and processing intricate details within images. Recent works~\citep{qwenvl, mplug-owl} have attempted to address these limitations by fine-tuning the vision transformer (ViT) during the LVLM training process, which has shown to yield improved results. To further enhance the model's adaptability to varying resolutions, we introduce dynamic resolution training in the LVLM training process. Specifically, we employ a 2D Rotary Position Embedding (RoPE) in the ViT, thus allowing the model to better capture information across different spatial scales.

\begin{figure*}[t]
\centering
\includegraphics[width= 1\linewidth]{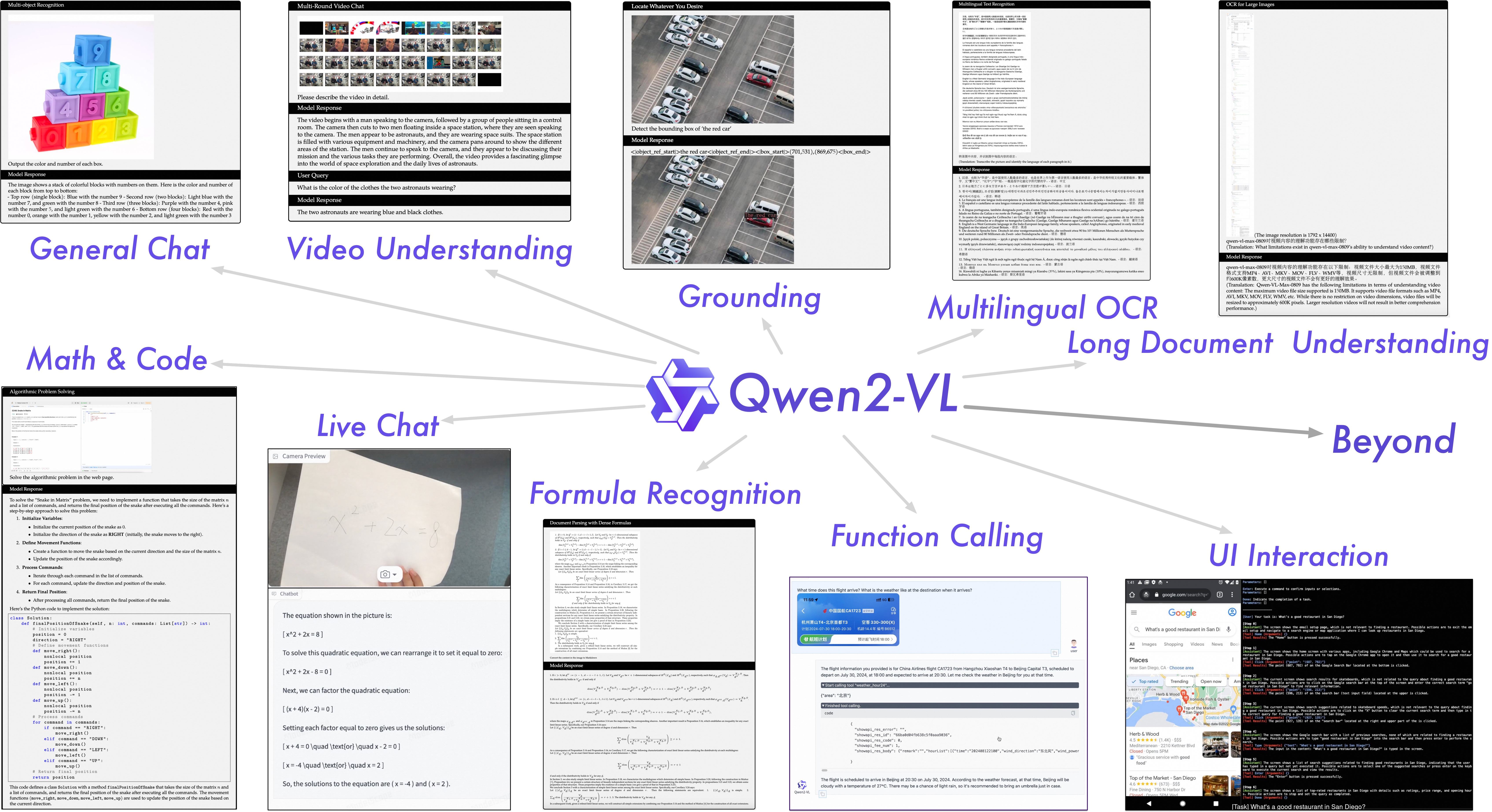}
   \caption{Qwen2-VL capabilities: Multilingual image text understanding, code/math reasoning, video analysis, live chat, agent potential, and more. See Appendix for details.}
\label{fig:example}
\end{figure*}

When it comes to video content, which is essentially a sequence of frames, many existing models continue to treat it as an independent modality. However, understanding the dynamic nature of reality, as manifested in videos, is crucial for models aiming to grasp the complexities of the real world. Unlike text, which is inherently one-dimensional, the real-world environment exists in three dimensions. The use of one-dimensional position embeddings in current models significantly limits their ability to model three-dimensional space and temporal dynamics effectively. To bridge this gap, we have developed Multimodal Rotary Position Embedding (M-RoPE), which employs separate components to represent temporal and spatial information. This enables the model to naturally comprehend dynamic content, such as videos or streaming data, improving its ability to understand and interact with the world.

Furthermore, compared to the scaling of large language models (LLMs), current LVLMs are still in the early stages of exploring the impact of scaling in terms of training data and model parameters. The exploration of scaling laws for LVLMs—how increases in model and data size affect performance—remains an open and promising area of research.

In this work, we introduce the newest addition to the large vision-language models of the Qwen family: Qwen2-VL series, which comprises three open-weight models
with total parameter counts of 2 billion, 8 billion, and 72 billion. As shown in Figure~\ref{fig:example}, the key advances in Qwen2-VL include:

\begin{itemize}

 \item \textbf{State-of-the-art understanding across various resolutions and aspect ratios:} Qwen2-VL achieves leading performance on visual benchmarks, including DocVQA, InfoVQA, RealWorldQA, MTVQA, MathVista, and others.
 
 \item \textbf{Comprehension of extended-duration videos (20 min+):} Qwen2-VL is capable of understanding videos over 20 minutes in length, enhancing its ability to perform high-quality video-based question answering, dialogue, content creation, and more.
 
 \item \textbf{Robust agent capabilities for device operation:} With advanced reasoning and decision-making abilities, Qwen2-VL can be integrated with devices such as mobile phones, robots, etc., enabling autonomous operation based on visual inputs and text instructions.
 
 \item \textbf{Multilingual support:} To serve a global audience, beyond English and Chinese, Qwen2-VL now supports multilingual context understanding within images, including most European languages, Japanese, Korean, Arabic, Vietnamese, and others.
 
\end{itemize}
 
\begin{table}[t]
\centering
\caption{Model descriptions of Qwen2-VL.}
\label{tab:model_description}
\scalebox{0.9}{
\begin{tabular}{c | c | c | m{11cm}}
\toprule
\textbf{Model Name} & \textbf{Vision Encoder} & \textbf{LLM} & \multicolumn{1}{c}{\textbf{Model Description}} \\
\midrule
Qwen2-VL-2B & 675M &1.5B & The \textbf{most efficient} model, designed to run on-device. It delivers adequate performance for most scenarios with limited resources. \\ 
\hline
Qwen2-VL-7B & 675M & 7.6B & The \textbf{performance-optimized} model in terms of cost, significantly upgraded for text recognition and video understanding capabilities. It delivers significant performance across a broad range of visual tasks. \\
\hline
Qwen2-VL-72B & 675M &72B & The \textbf{most capable} model, further improvements in visual reasoning, instruction-following, decision-making, and agent capabilities. It delivers optimal performance on most complex tasks. \\
\bottomrule
\end{tabular}
}
\end{table}

\begin{figure*}[t]
\centering
\includegraphics[width= 1\linewidth]{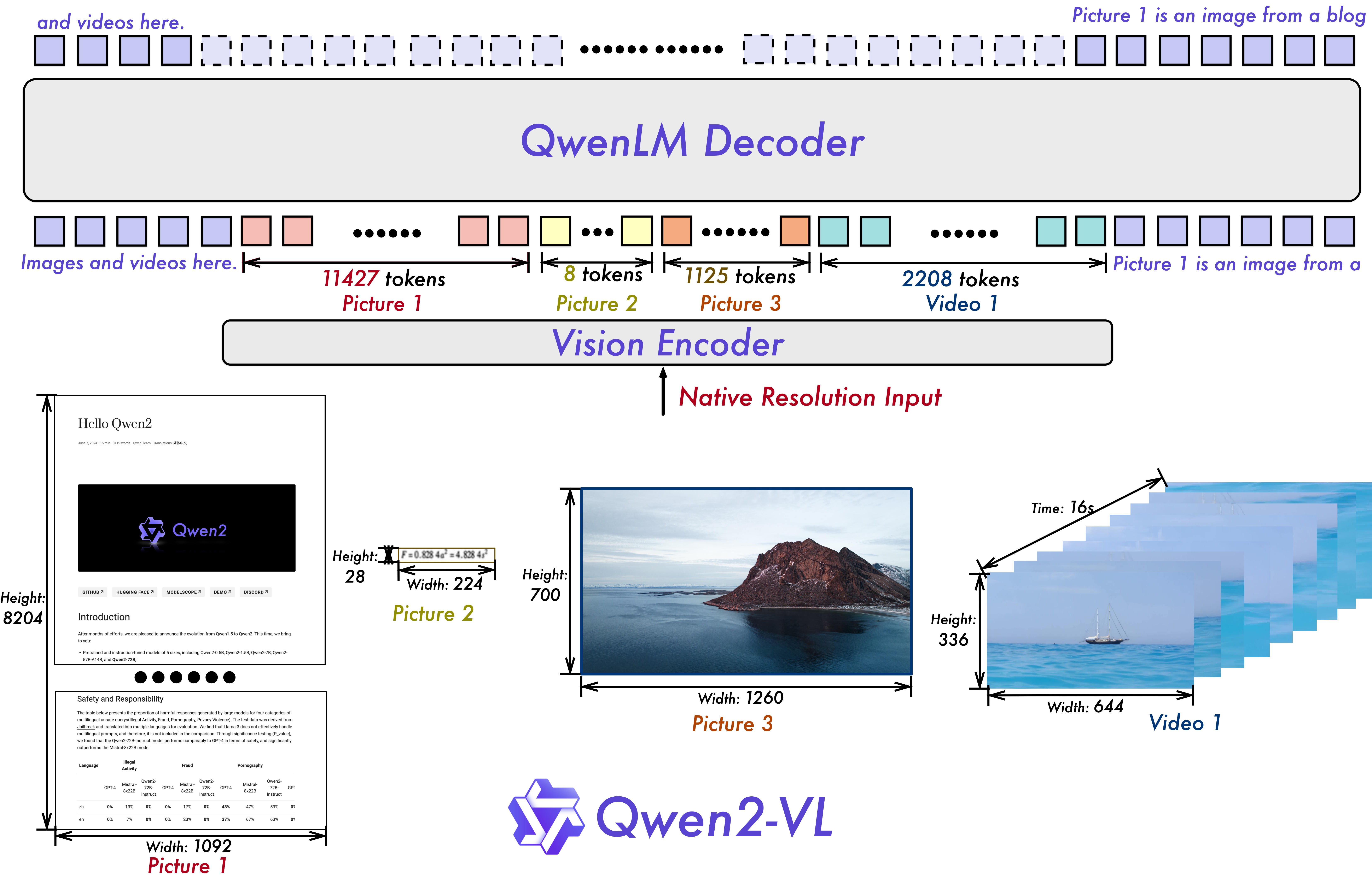}
   \caption{Qwen2-VL is capable of accurately identifying and comprehending the content within images, regardless of their clarity, resolution, or extreme aspect ratios.}
\label{fig:model}
\end{figure*}

\section{Approach}
The Qwen2-VL series consists of models of 3 sizes, which are Qwen2-VL-2B, Qwen2-VL-7B and Qwen2-VL-72B. Table~\ref{tab:model_description} lists the hyper-parameters and important information.
Notably, Qwen2-VL employs a 675M parameter ViT across various-sized LLMs, ensuring that the computational load of the ViT remains constant regardless of the scale of the LLM.
 
\subsection{Model Architecture}
Figure~\ref{fig:model} illustrates the comprehensive structure of Qwen2-VL. We have retained the Qwen-VL~\citep{qwenvl} framework, which integrates vision encoders and language models. 
For various scale adaptations, we have implemented a Vision Transformer (ViT)~\citep{vit} with approximately 675 million parameters, adept at handling both image and video inputs.
In terms of language processing, we have opted for the more powerful Qwen2~\citep{qwen2} series of language models.
To further enhance the model’s ability to effectively perceive and comprehend visual information in videos, we introduced several key upgrades:

\paragraph{Naive Dynamic Resolution}

\begin{figure*}[t]
\centering
\includegraphics[width= 1\linewidth]{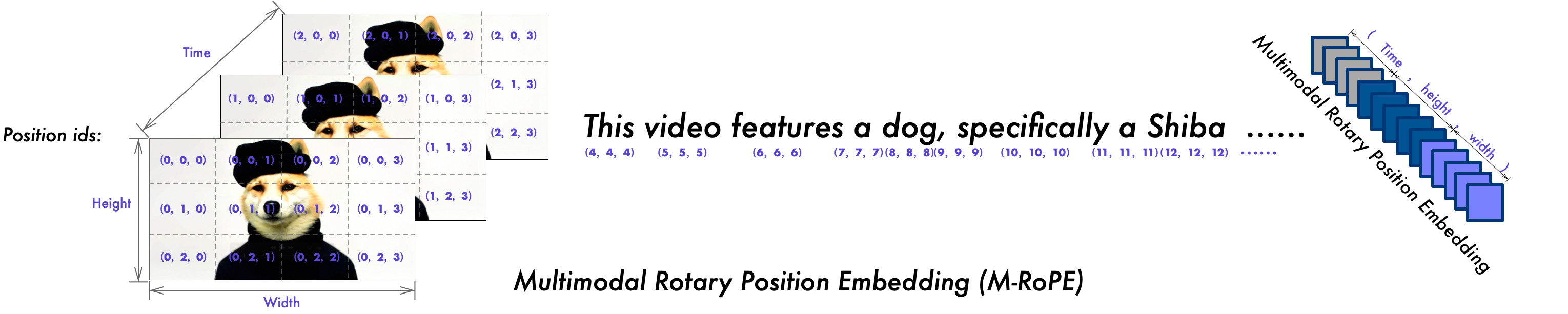}
   \caption{A demonstration of M-RoPE. By decomposing rotary embedding into temporal, height, and width components, M-RoPE can explicitly model the positional information of text, images, and video in LLM.}
\label{fig:mrope}
\end{figure*}

A key architectural improvement in Qwen2-VL is the introduction of naive dynamic resolution support~\citep{navit}. Unlike Qwen-VL, Qwen2-VL can now process images of any resolution, dynamically converting them into a variable number of visual tokens.\footnote{This technology was previously implemented in the internal iterations, Qwen-VL Plus and Qwen-VL MAX. We have further upgraded it in Qwen2-VL.}
To support this feature, we modified ViT by removing the original absolute position embeddings and introducing 2D-RoPE~\citep{rope,2drope} to capture the two-dimensional positional information of images.
At the inference stage, images of varying resolutions are packed into a single sequence, with the packed length controlled to limit GPU memory usage. 
Furthermore, to reduce the visual tokens of each image, a simple MLP layer is employed after the ViT to compress adjacent $2\times2$ tokens into a single token, with the special <|vision\_start|> and <|vision\_end|> tokens placed at the beginning and end of the compressed visual tokens.
As a result, an image with a resolution of $224\times224$, encoded with a ViT using patch\_size=14, will be compressed to 66 tokens before entering LLM.

\paragraph{Multimodal Rotary Position Embedding (M-RoPE)}

Another key architectural enhancement is the innovation of Multimodal Rotary Position Embedding (M-RoPE). 
Unlike the traditional 1D-RoPE in LLMs, which is limited to encoding one-dimensional positional information, M-RoPE effectively models the positional information of multimodal inputs.
This is achieved by deconstructing the original rotary embedding into three components: temporal, height, and width.
For text inputs, these components utilize identical position IDs, making M-RoPE functionally equivalent to 1D-RoPE~\citep{ropetie}. 
When processing images, the temporal IDs of each visual token remain constant, while distinct IDs are assigned to the height and width components based on the token's position in the image. 
For videos, which are treated as sequences of frames, the temporal ID increments for each frame, while the height and width components follow the same ID assignment pattern as images. 
In scenarios where the model's input encompasses multiple modalities, position numbering for each modality is initialized by incrementing the maximum position ID of the preceding modality by one.
An illustration of M-RoPE is shown in Figure~\ref{fig:mrope}.
M-RoPE not only enhances the modeling of positional information but also reduces the value of position IDs for images and videos, enabling the model to extrapolate to longer sequences during inference.

\paragraph{Unified Image and Video Understanding}

Qwen2-VL employs a mixed training regimen incorporating both image and video data, ensuring proficiency in image understanding and video comprehension. 
To preserve video information as completely as possible, we sampled each video at two frames per second. 
Additionally, we integrated 3D convolutions~\citep{i3d} with a depth of two to process video inputs, allowing the model to handle 3D tubes instead of 2D patches, thus enabling it to process more video frames without increasing the sequence length~\citep{vivit}. 
For consistency, each image is treated as two identical frames. 
To balance the computational demands of long video processing with overall training efficiency, we dynamically adjust the resolution of each video frame, limiting the total number of tokens per video to 16384. 
This training approach strikes a balance between the model's ability to comprehend long videos and training efficiency.

\subsection{Training}
Following Qwen-VL~\citep{qwenvl}, we adopt a three-stage training methodology. In the first stage, we focus exclusively on training the Vision Transformer (ViT) component, utilizing a vast corpus of image-text pairs to enhance semantic understanding within the Large Language Model (LLM). In the second stage, we unfreeze all parameters and train with a wider range of data for more comprehensive learning.  In the final stage, we lock the ViT parameters and perform exclusive fine-tuning of the LLM using instructional datasets.

The model is pre-trained on a diverse dataset that includes image-text pairs, optical character recognition (OCR) data, interleaved image-text articles, visual question answering datasets, video dialogues, and image knowledge datasets. Our data sources primarily comprise cleaned web pages, open-source datasets, and synthetic data. The cutoff date for our data knowledge is June 2023. This diverse data composition is instrumental in developing a robust multimodal understanding capability.

During the initial pre-training phase, Qwen2-VL is exposed to a corpus of around 600 billion tokens. The LLM component of Qwen2-VL is initialized using the parameters from Qwen2~\citep{qwen2}, while the vision encoder of Qwen2-VL is initialized with the ViT derived from DFN. However, the fixed position embedding in the original DFN’s ViT~\citep{fang2023data} is replaced by RoPE-2D. This pre-training phase primarily focuses on learning image-text relationships, textual content recognition within images through OCR, and image classification tasks. Such foundational training is instrumental in enabling the model to develop a robust understanding of core visual-textual correlations and alignments.

The second pre-training phase marks a significant progression, involving an additional 800 billion tokens of image-related data. This stage introduces a higher volume of mixed image-text content, facilitating a more nuanced understanding of the interplay between visual and textual information. The incorporation of visual question answering datasets refines the model's capacity to respond to image-related queries. Moreover, the inclusion of multitasking datasets is pivotal in developing the model's ability to navigate diverse tasks concurrently, a skill of paramount importance when dealing with complex, real-world datasets. Concurrently, purely textual data continues to play a crucial role in maintaining and advancing the model's linguistic proficiency.

Throughout the pre-training stages, Qwen2-VL processes a cumulative total of 1.4 trillion tokens. Specifically, these tokens encompass not only text tokens but also image tokens. During the training process, however, we only provide supervision for the text tokens. This exposure to extensive and diverse linguistic and visual scenarios ensures that the model develops a deep understanding of the intricate relationships between visual and textual information, thereby laying a robust foundation for various multimodal tasks.

During the instruction fine-tuning phase, we employ the ChatML~\citep{chatml} format to construct instruction-following data. This dataset encompasses not only pure text-based dialogue data but also multimodal conversational data. The multimodal components include image question-answering, document parsing, multi-image comparison, video comprehension, video stream dialogue, and agent-based interactions. Our comprehensive approach to data construction aims to enhance the model's capability to understand and execute a wide range of instructions across various modalities. By incorporating diverse data types, we seek to develop a more versatile and robust language model capable of handling complex, multimodal tasks in addition to traditional text-based interactions.

\subsubsection{Data Format.}
In line with Qwen-VL, Qwen2-VL also employs special tokens to distinguish vision and text inputs. Tokens <|vision\_start|> and <|vision\_end|> are inserted at the start and end of the image feature sequence to demarcate the image content. 

\paragraph{Dialogue Data.}
In terms of dialogue format, we construct our instruction tuning dataset using the ChatML format, where each interaction's statement is marked with two special tokens (<|im\_start|> and <|im\_end|>) to facilitate dialogue termination. The sections marked in blue indicate the supervised parts.

\begin{tcolorbox}[colback=black!5!white,colframe=black!75!black,title=The Dataset Format Example of ChatML]
\textcolor{blue}{<|im\_start|>}user

<|vision\_start|>Picture1.jpg<|vision\_end|><|vision\_start|>Picture2.jpg<|vision\_end|>What do the two pictures have in common?\textcolor{blue}{<|im\_end|>}

\textcolor{blue}{<|im\_start|>}assistant

\textcolor{blue}{Both pictures are of SpongeBob SquarePants. <|im\_end|>}

\textcolor{blue}{<|im\_start|>}user

What is happening in the video?<|vision\_start|>video.mp4<|vision\_end|>\textcolor{blue}{<|im\_end|>}

\textcolor{blue}{<|im\_start|>}assistant

\textcolor{blue}{The protagonist in the video is frying an egg.<|im\_end|>}
\end{tcolorbox}

\paragraph{Visual Grounding.}
To endow the model with visual grounding capabilities, bounding box coordinates are normalized within [0, 1000) and represented as "$(X_{\text{top left}}, Y_{\text{top left}}), (X_{\text{bottom right}}, Y_{\text{bottom right}})$".
Tokens <|box\_start|> and <|box\_end|> are utilized to demarcate bounding box text. To accurately link bounding boxes with their textual descriptions, we introduce tokens <|object\_ref\_start|> and <|object\_ref\_end|> to indicate the content that the bounding box references, thereby allowing the model to effectively interpret and generate precise descriptions of specific regions.

\begin{tcolorbox}[colback=black!5!white,colframe=black!75!black,title=Referring Grounding]
<|vision\_start|>Picture1.jpg<|vision\_end|>

<|object\_ref\_start|>the eyes on a giraffe<|object\_ref\_end|>\textcolor{blue}{<|box\_start|>(176,106),(232,160) <|box\_end|> }
\end{tcolorbox}

\paragraph{Visual Agent.}
To develop Qwen2-VL as a general-purpose VL-Agent, we treat various agent tasks, such as UI Operations, Robotic Control, Games, and Navigation, as sequential decision-making problems, enabling Qwen2-VL to accomplish tasks through multi-step action execution. For each task, we first define a set of permissible actions and keywords pattern (underline) for function call~\citep{qwen-agent}. Qwen2-VL then analyzes the observations, performs reasoning and planning, executes the selected actions, and interacts with the environment to acquire new observations. This cycle repeats iteratively until the task is successfully completed. By integrating various tools and leveraging the vision perception capabilities of large vision-language models (LVLMs), Qwen2-VL is able to iteratively execute increasingly complex tasks involving real-world visual interactions.

\begin{tcolorbox}[colback=black!5!white,colframe=black!75!black,title=Visual Agent]
\textcolor{blue}{<|im\_start|>}system

You are a helpful assistant.

\# Actions

\#\# You have the following actions.

\#\#\# Tap 

\textcolor{black}{Tap: A gentle tap that commands, chooses, or navigates through a smartphone's user interface. Parameters: [\{"name": "point", "description": "The specific spot of interest on the monitor, denoted by the coordinates (x, y) where x and y range from 0 to 1000.", "required": True\}]}

\#\#\# Home 

\textcolor{black}{Home: Go to phone's home screen. Parameters: []}

\#\#\# Other Actions ...

\#\# Continuously take action until the task is completed.

\underline{*FUNCTION*:} The action to take, should be one of \{Actions\}.

\underline{*ARGS*:} The input of the action.

\underline{*RESULT*:} Action results.

\underline{*RETURN*:} Reply based on action results. \textcolor{blue}{<|im\_end|>}

\textcolor{blue}{<|im\_start|>}user

Find a pizza restaurant nearby in Map. <|vision\_start|>Screenshot\_1.jpg<|vision\_end|> \textcolor{blue}{<|im\_end|>}

\textcolor{blue}{<|im\_start|>}assistant 

\textcolor{blue}{Several e-mails are displaying on the phone screen. To open Map, I need go back to the home screen and find the corresponding app icon.}

\textcolor{blue}{\underline{*FUNCTION*}: Home \underline{*ARGS*:} \{\}}

\textcolor{blue}{\underline{*RESULT*:}} <|vision\_start|>Screenshot\_2.jpg<|vision\_end|>

\textcolor{blue}{\underline{*RETURN*:} I return to the home screen. Next, I need to find the icon of Map and tap on it.}

\textcolor{blue}{\underline{*FUNCTION*:} Tap \underline{*ARGS*:} \{"point": (348,291)\}}

\textcolor{blue}{\underline{*RESULT*:}} <|vision\_start|>Screenshot\_3.jpg<|vision\_end|>

\textcolor{blue}{\underline{*RETURN*:} [Thinking for the next action.]}

\textcolor{blue}{[Other subsequent actions.] ......}

\textcolor{blue}{I have found the pizza restaurant nearby in Map.} \textcolor{blue}{<|im\_end|>}
\end{tcolorbox}

\subsection{Multimodal Model Infrastructure}
The Qwen2-VL models were trained on Alibaba Cloud's PAI-Lingjun Intelligent Computing Service~\citep{pai-lingjun} with its scalable computing, auto resuming and straggler detection.

\paragraph{Storage.} We use Alibaba Cloud's ultra-speed CPFS (Cloud Parallel File Storage)~\citep{cpfs} to build a storage system of Qwen2-VL pre-training and post-training. We decoupled the text data and vision data storage. We simply store text data on CPFS and use mmap for efficient access. For vision data, we use Alibaba Cloud's OSS (Object Storage Service)~\citep{oss} for persistent storage. During training, we accessed vision data through OSS's python-client concurrently and tuned the concurrency and retrying parameters to avoid reaching the QPS (queries per second) limit. We also found that video data decoding is a main bottleneck, especially for long videos. After several attempts with open-source~\citep{ffmpeg} and in-house software failed, we opted for a caching decoding technique. Checkpointing saves each GPU’s optimizer and model states on CPFS.

\paragraph{Parallelism.}
We use 3D parallelism which combines data parallelism (DP)~\citep{pytorchddp}, tensor parallelism (TP)~\citep{alexnet,megatron1} and pipeline parallelism (PP)~\citep{gpipe,megatron2,bfspp} to scale Qwen2-VL model training. We also leverage deepspeed's zero-1 redundancy optimizer~\citep{zero1} to shard states for memory saving. Sequence parallelism (SP)~\citep{megatron3} with selective checkpointing activation~\citep{checkpointing} was leveraged to reduce memory usage. When enabling TP training, we always shard the vision encoder and large language models together but not the vision merger due to its relatively few parameters. We found the TP training would result in different model shared-weights due to the convolution operator's non-deterministic behavior~\footnote{\href{https://pytorch.org/docs/stable/notes/randomness.html}{https://pytorch.org/docs/stable/notes/randomness.html}}. We resolved this issue by performing offline reduction of the shared weights, thereby avoiding an additional \textbf{all-reduce} communication step. This approach resulted in only a minimal impact on performance. We leverage 1F1B PP~\citep{megatron2} for Qwen2-VL 72B training. We combine the vision encoder, vision adapter and several LLM's decoder layers into one stage, and evenly split the remaining decoder layers. Note that the vision and text sequence lengths are dynamic for each data point. We \textbf{broadcast} the dynamic sequence lengths before initiating the 1F1B process and access the shape information using batch indices. We also implemented an interleaved 1F1B PP~\citep{megatron2} but found it is slower than the standard 1F1B setting.

\paragraph{Software.}
We use PyTorch~\citep{pytorch,pytorch2} version 2.1.2 with CUDA 11.8~\citep{cuda} for training. Additionally, we leverage flash-attention~\citep{flash1, flash2, flash3} for efficient training in both the vision encoder and the LLM. We also utilize fused operators~\citep{apex} such as LayerNorm~\citep{layernorm}, RMSNorm~\citep{rmsnorm}, and Adam~\citep{adamw}. Besides this, we leverage the overlap of communication and computation during matrix multiplication in our training process.

\section{Experiments}

In this section, we first evaluate the model's performance by conducting a comparative analysis across a variety of visual benchmarks, demonstrating the advantages of our approach. Subsequently, we carry out a detailed examination of specific capabilities, including general visual perception, document understanding, multilingual recognition in images, video comprehension, and agent abilities. Finally, we present an ablation study to investigate several key components of our approach.

\begin{table}[t] 
\centering 
\caption{Performance Comparison of Qwen2-VL Models and State-of-the-art.} 
\label{tab:comparison}
\scalebox{0.68}{
\begin{tabular}{l|c|c|c|c|c|c} 
\toprule 
Benchmark & Previous SoTA  & Claude-3.5 Sonnet & GPT-4o &Qwen2-VL-72B&Qwen2-VL-7B& Qwen2-VL-2B\\ 
\midrule 
MMMU\textsubscript{val}~\citep{yue2023mmmu} & 66.1~\citep{grok2} & 68.3 & \textbf{69.1} & 64.5 & 54.1 & 41.1 \\ 
DocVQA\textsubscript{test}~\citep{docvqa} & 94.1~\citep{internvl2} & 95.2 & 92.8 & \textbf{96.5} & 94.5 & 90.1 \\ 
InfoVQA\textsubscript{test}~\citep{docvqa} & 82.0~\citep{internvl2} & - & - & \textbf{84.5} & 76.5 & 65.5 \\ 
AI2D~\citep{kembhavi2016diagram} & 87.6~\citep{internvl2} & 80.2(94.7) & 84.6(94.2) & \textbf{88.1} & 83.0 & 74.7 \\ 
ChartQA\textsubscript{test}~\citep{masry2022chartqa} & 88.4~\citep{internvl2} & \textbf{90.8} & 85.7 & 88.3 & 83.0 & 73.5 \\ 
TextVQA\textsubscript{val}~\citep{textvqa} & 84.4~\citep{internvl2} & - & - & \textbf{85.5} & 84.3 & 79.7 \\ 
OCRBench~\citep{liu2024ocrbenchhiddenmysteryocr} & 852~\citep{minicpm-v} & 788 & 736 & \textbf{877} & 866 & 809 \\ 
MTVQA~\citep{tang2024mtvqa} & 23.2~\citep{gemini} & 25.7 & 27.8 & \textbf{30.9} & 25.6 & 18.1 \\ 
VCR\textsubscript{en easy}~\citep{zhang2024vcr} & 84.7~\citep{internvl2} & 63.9 & 91.6 & \textbf{91.9} & 89.7 & 81.5 \\ 
VCR\textsubscript{zh easy}~\citep{zhang2024vcr} & 22.1~\citep{internvl2} & 1.0 & 14.9 & \textbf{65.4} & 59.9 & 46.2 \\ 
RealWorldQA~\citep{grok15} & 72.2~\citep{internvl2} & 60.1 & 75.4 & \textbf{77.8} & 70.1 & 62.9 \\ 
MME\textsubscript{sum}~\citep{fu2023mme} & 2414.7~\citep{internvl2} & 1920.0 & 2328.7 & \textbf{2482.7} & 2326.8 & 1872.0 \\ 
MMBench-EN\textsubscript{test}~\citep{MMBench} & \textbf{86.5}~\citep{internvl2} & 79.7 & 83.4 & \textbf{86.5} & 83.0 & 74.9 \\ 
MMBench-CN\textsubscript{test}~\citep{MMBench} & 86.3~\citep{internvl2} & 80.7 & 82.1 & \textbf{86.6} & 80.5 & 73.5 \\ 
MMBench-V1.1\textsubscript{test}~\citep{MMBench} & 85.5~\citep{internvl2} & 78.5 & 82.2 & \textbf{85.9} & 80.7 & 72.2 \\ 
MMT-Bench\textsubscript{test}~\citep{mmtbench} & 63.4~\citep{internvl1.5} & - & 65.5 & \textbf{71.7} & 63.7 & 54.5 \\ 
MMStar~\citep{chen2024we} & 67.1~\citep{internvl2} & 62.2 & 63.9 & \textbf{68.3} & 60.7 & 48.0 \\ 
MMVet\textsubscript{GPT-4-Turbo}~\citep{yu2024mm} & 67.5~\citep{gpt4v} & 66.0 & 69.1 & \textbf{74.0} & 62.0 & 49.5 \\ 
HallBench\textsubscript{avg}~\citep{guan2023hallusionbench} & 55.2~\citep{internvl2} & 49.9 & 55.0 & \textbf{58.1} & 50.6 & 41.7 \\ 
MathVista\textsubscript{testmini}~\citep{mathvista} & 69.0~\citep{grok2} & 67.7 & 63.8 & \textbf{70.5} & 58.2 & 43.0 \\ 
MathVision~\citep{mathvision} & 30.3~\citep{gpt4} & - & \textbf{30.4} & 25.9 & 16.3 & 12.4 \\ 
MMMU-Pro~\citep{mmmupro} & 46.9~\citep{gemini} & 51.5 & \textbf{51.9} & 46.2 & 43.5 & 37.6 \\ 
\bottomrule 
\end{tabular} 
}
\end{table}

\begin{table}[t]
\centering
\caption{Performance of Qwen2-VL and GPT-4o on internal multilingual OCR benchmarks.}
\label{tab:additional_performance_comparison_multilingual_ocr}
\scalebox{1}{
\begin{tabular}{l|cccccccc}
\toprule
\textbf{Language} & Korean & Japanese & French & German & Italian & Russian & Vietnamese & Arabic \\
\midrule
GPT-4o & 87.8 & 88.3 & 89.7 & 88.3 & 74.1 & 96.8 & 72.0 & \textbf{75.9} \\
Qwen2-VL-72B & \textbf{94.5} & \textbf{93.4} & \textbf{94.1} & \textbf{91.5} & \textbf{89.8} & \textbf{97.2} & \textbf{73.0} & 70.7 \\
\bottomrule
\end{tabular}
}
\end{table}

\begin{table}[t]
\centering
\caption{Performance of Qwen2-VL and other models on video benchmarks.}
\label{tab:additional_performance_comparison}
\scalebox{0.75}{
\begin{tabular}{l|cccccc}
\toprule
\textbf{Benchmark} &Previous SoTA & Gemini 1.5-Pro & GPT-4o & Qwen2-VL-72B & Qwen2-VL-7B & Qwen2-VL-2B \\
\midrule
MVBench~\citep{li2024mvbench} & 69.6 & - & - & \textbf{73.6} & 67.0 & 63.2 \\
PerceptionTest\textsubscript{test}~\citep{patraucean2024perception} & 66.9 & - & - & \textbf{68.0} & 62.3 & 53.9 \\
EgoSchema\textsubscript{test}~\citep{mangalam2023egoschema} & 62.0 & 63.2 & 72.2 & \textbf{77.9} & 66.7 & 54.9 \\
Video-MME\textsubscript{(wo/w subs)}~\citep{fu2024video} & 66.3/69.6 & \textbf{75.0}/\textbf{81.3} & 71.9/77.2 & 71.2/77.8 & 63.3/69.0 & 55.6/60.4 \\
\bottomrule
\end{tabular}
}
\end{table}

\begin{table}[t]
\centering
\caption{Performance Comparison of Qwen2-VL-72B across various agent benchmarks and GPT-4o. SR, GC, TM and EM are short for success rate, goal-condition success, type match and exact match. ALFRED, R2R and REVERIE are performance in valid-unseen.}
\label{tab:comparison_v3}
\scalebox{0.85}{
\begin{tabular}{l|l|c|c|c|c}
\toprule
 & Benchmark & Metric & Previous SoTA & GPT-4o & \textbf{Qwen2-VL-72B} \\
\midrule
\multirow{2}{*}{General} & \multirow{2}{*}{FnCall} & TM  & -   & 90.2  & \textbf{93.1} \\
                         &                           & EM  & -   & 50.0  & \textbf{53.2} \\
\midrule
\multirow{2}{*}{UI Operations} &\multirow{2}{*}{AITZ~\citep{zhang2024android}}                      & TM  & 83.0~\citep{hong2023cogagent}  & 70.0  & \textbf{89.6}  \\
                         &                           & EM  & 47.7~\citep{zhan2023you}  & 35.3  & \textbf{72.1}  \\
\midrule
\multirow{4}{*}{Card Games}   & Number Line~\citep{zhai2024fine}           & SR  & 89.4~\citep{zhai2024fine} & 91.5  & \textbf{100.0} \\
                         & BlackJack~\citep{zhai2024fine}                 & SR  & 40.2~\citep{zhai2024fine}  & 34.5  & \textbf{42.6}  \\
                         & EZPoint~\citep{zhai2024fine}                 & SR  & 50.0~\citep{zhai2024fine}  & 85.5  & \textbf{100.0} \\
                         & Point24~\citep{zhai2024fine}                   & SR  & 2.6~\citep{llava}  & 3.0   & \textbf{4.5}   \\
\midrule
\multirow{2}{*}{Robotic Control} &\multirow{2}{*}{ALFRED~\citep{shridhar2020alfred}}  & SR   & 67.7~\citep{lu2023thinkbot}   & -     & \textbf{67.8} \\
                         &                              & GC   & 75.3~\citep{lu2023thinkbot} & -     & \textbf{75.8} \\
\midrule
\multirow{2}{*}{Navigation}     & R2R~\citep{anderson2018vision}    & SR   & \textbf{79.0}~\citep{chen2022think}           & 43.7 & 51.7 \\
                         & REVERIE~\citep{qi2020reverie} & SR   & \textbf{61.0}~\citep{sigurdsson2023rrex}           & 31.6& 31.0 \\
\bottomrule
\end{tabular}
}
\end{table}

\subsection{Compare to SOTAs}
We evaluate the visual capabilities of our model through various visual benchmarks, video tasks, and agent-based assessments. Qwen2-VL demonstrates highly competitive performance at the same scale, achieving new state-of-the-art (SoTA) results. Overall, our 72B model consistently delivers top-tier performance across most evaluation metrics, frequently surpassing even closed-source models such as GPT-4o~\citep{gpt4o} and Claude 3.5-Sonnet~\citep{sonnet3_5}. Notably, it exhibits a significant advantage in document understanding tasks. However, in the MMMU~\citep{yue2023mmmu} benchmark, our model still lags behind GPT-4o to some extent, indicating that Qwen2-VL-72B has room for improvement when handling more complex and challenging problem sets.

\subsection{Quantitative Results} % \subsection{the numbers}
In this section, we present an extensive evaluation of the Qwen2-VL series across an array of datasets, offering a comprehensive understanding of the model's capabilities in various aspects.

\subsubsection{General Visual Question Answering}

To rigorously assess our models' capabilities in general visual question answering tasks, we conduct extensive evaluations across a diverse array of state-of-the-art benchmarks: RealWorldQA \citep{grok15}, MMStar \citep{chen2024we}, MMVet \citep{yu2024mm}, MMT-Bench \citep{mmtbench}, MMBench \citep{MMBench}, MMbench-1.1 \citep{MMBench}, MME \citep{fu2023mme}, and HallusionBench \citep{guan2023hallusionbench}. The Qwen2-VL series exhibits exceptional performance across these benchmarks, with the 72B model consistently achieving or surpassing state-of-the-art results, while the 7B and 2B variants also demonstrate robust capabilities. On RealWorldQA, which evaluates real-world spatial comprehension, Qwen2-VL-72B achieves a score of 77.8, surpassing both the previous state-of-the-art (72.2) and formidable baselines such as GPT-4o (75.4), thus demonstrating superior understanding of physical environments. For MMStar, a benchmark designed to assess genuine multimodal capabilities through visually indispensable samples, Qwen2-VL-72B attains 68.3, outperforming the previous best of 67.1 and highlighting its proficiency in integrating visual and textual information. On MMVet, which evaluates the integration of core vision-language capabilities across 16 complex multimodal tasks, Qwen2-VL-72B achieves a remarkable 74.0, significantly outperforming strong competitors including GPT-4V (67.5) and showcasing its versatility in addressing diverse multimodal challenges. In the MMT-Bench evaluation, which assesses advanced reasoning and instruction following across 32 core meta-tasks and 162 subtasks in multimodal understanding, Qwen2-VL-72B achieves 71.7, markedly surpassing the previous best (63.4) and demonstrating its prowess in applying expert knowledge and executing deliberate visual recognition, localization, reasoning, and planning. On MMBench, which evaluates fine-grained abilities across 20 dimensions, Qwen2-VL-72B exhibits strong performance, achieving 86.5 on the English test set, matching the state-of-the-art, and 86.6 on the Chinese test set, establishing a new benchmark. For MME, which measures a wide spectrum of perception and cognition abilities across 14 subtasks, Qwen2-VL-72B achieves a cumulative score of 2482.7, significantly outperforming the previous best (2414.7), underscoring its advanced capabilities in both visual perception and high-level cognition tasks.

These comprehensive results underscore the Qwen2-VL series' exceptional proficiency in general visual question answering tasks. The models demonstrate advanced capabilities in real-world spatial comprehension, genuine multimodal integration, complex reasoning, instruction following, and a broad range of perception and cognition tasks. The consistent superior performance across diverse benchmarks, particularly the outstanding results of the 72B model, positions the Qwen2-VL series as a leading solution in the field of visual question answering. Our models excel in handling visually indispensable tasks, integrating core vision-language capabilities, and demonstrating expertise across diverse multimodal scenarios, ranging from fundamental perception tasks to complex reasoning and planning. This exhaustive evaluation highlights the Qwen2-VL series' versatility and effectiveness in addressing the multifaceted challenges posed by state-of-the-art multimodal benchmarks, thereby setting a new standard for large vision-language models.

\subsubsection{Document and Diagrams Reading}
We tested our model's OCR and document and diagram comprehension on DocVQA~\citep{docvqa}, ChartQA~\citep{masry2022chartqa},InfoVQA~\citep{docvqa}, TextVQA~\citep{textvqa},AI2D~\citep{kembhavi2016diagram} datasets. The DocVQA/InfoVQA/ChartQA dataset focuses on the model's ability to comprehend text in documents/high-resolution infographics/charts, while the TextVQA dataset examines the ability to comprehend text in naturalistic images. The OCRBench dataset is a a dataset of mixed tasks, which focuses on mathematical formula parsing and information extraction in addition to the text-based VQA. The AI2D dataset focuses on multiple-choice questions on scientific diagrams containing text. In addition, we also tested the OCR and formula recognition capabilities of our model on OCRBench~\citep{liu2024ocrbenchhiddenmysteryocr}, as well as the multilingual OCR capabilities of our model on the MTVQA~\citep{tang2024mtvqa} dataset.

The experimental results show that our model achieves SoTA level in several metrics, including DocVQA, InfoVQA, TextVQA and OCRBench, demonstrating that our model has good comprehension of textual content in images from multiple domains.

\subsubsection{Multilingual Text Recognition and Understanding}

In particular, our model surpasses all existing general-purpose LVLMs in multilingual OCR. Our model not only outperforms existing LVLMs (including proprietary models such as GPT-4o, Claude 3.5 Sonnet, etc.) on the public-available MTVQA dataset, it also outperforms GPT-4o on the in-house internal benchmark across all foreign languages except Arabic (Table \ref{tab:additional_performance_comparison_multilingual_ocr}).

\subsubsection{Mathematical Reasoning}
We've conducted experiments on the MathVista~\citep{mathvista} and MathVision~\citep{mathvision} datasets to assess mathematical reasoning capabilities. MathVista is a comprehensive benchmark featuring 6,141 diverse examples of mathematical and visual tasks. The MathVision dataset comprises 3,040 math problems embedded in visual contexts from actual math competitions, covering 16 mathematical disciplines and varying in difficulty across five levels. These challenges underscore the necessity for LVLMs to exhibit strong visual comprehension, a deep understanding of mathematics, and sound logical reasoning skills. The Qwen2-VL series has demonstrated superior performance on MathVista, achieving a 70.5 outperforming other LVLMs. Additionally, it has set a new open-source benchmark on MathVision with 25.9.

\subsubsection{Referring Expression Comprehension}

Regarding visual localization task, we evaluate Qwen2-VL on RefCOCO, RefCOCO+, and RefCOCOg datasets~\citep{refcoco,refcocog}. The results, as depicted in Table~\ref{tab:grounding}, demonstrate that Qwen2-VL attains top-tier results among generalist models. Benefiting from a more rational structure design, Qwen2-VL is able to perceive details in high-resolution images, leading to significant improvements over Qwen-VL.
The superiority of these models in comparison to both generalist and specialized models highlights their potential for advancing the field of visual localization and their capacity for real-world implementation in tasks requiring precise visual understanding.

\begin{table}[t] 
\centering 
\caption{Performance Comparison on Referring Expression Comprehension Task.} 
\label{tab:grounding}
\scalebox{0.85}{
\begin{tabular}{l|l|cccccccc}
\toprule
\multirow{2}{*}{\textbf{Type}}
& \multirow{2}{*}{\textbf{Model}} & \multicolumn{3}{c}{\textbf{RefCOCO}} & \multicolumn{3}{c}{\textbf{RefCOCO+}} & \multicolumn{2}{c}{\textbf{RefCOCOg}} \\
\cmidrule(lr){3-5}\cmidrule(lr){6-8}\cmidrule(lr){9-10}
 & & val & test-A & test-B & val & test-A & test-B & val & test \\
\cmidrule(lr){1-10}
\multirow{11}{*}{\textit{Generalist}}
& OFA-L~\citep{wang2022ofa}       &	80.0 	&	83.7 	&	76.4 	&	68.3 	&	76.0 	&	61.8 	&	67.6 	&	67.6 	\\
& Shikra~\citep{shikra}       &	87.0 	&	90.6 	&	80.2 	&	81.6 	&	87.4 	&	72.1 	&	82.3 	&	82.2 	\\
& Qwen-VL~\citep{qwenvl}      &	89.4 	&	92.3 	&	85.3 	&	83.1 	&	88.3 	&	77.2 	&	85.6 	&	85.5 	\\
& Ferretv2~\citep{ferretv2}   &	92.6 	&	95.0 	&	88.9 	&	87.4 	&	92.1 	&	81.4 	&	89.4 	&	90.0 	\\
& CogVLM~\citep{cogvlm}       &	92.8 	&	94.8 	&	89.0 	&	88.7 	&	92.9 	&	83.4 	&	89.8 	&	90.8 	\\
% & InternVL2\textsubscript{1b}~\citep{internvl2}    & 83.6 & 88.7 & 79.8 & 76.0 & 83.6 & 67.7 & 80.2 & 79.9 \\
& InternVL2\textsubscript{2b}~\citep{internvl2}    & 82.3 & 88.2 & 75.9 & 73.5 & 82.8 & 63.3 & 77.6 & 78.3 \\
% & InternVL2\textsubscript{4b}~\citep{internvl2}    & 88.5 & 91.2 & 83.9 & 81.2 & 87.2 & 73.8 & 84.6 & 84.6 \\
& InternVL2\textsubscript{8b}~\citep{internvl2}    & 87.1 & 91.1 & 80.7 & 79.8 & 87.9 & 71.4 & 82.7 & 82.7 \\
% & InternVL2\textsubscript{26b}~\citep{internvl2}   & 91.2 & 93.3 & 87.4 & 86.8 & 91.0 & 81.2 & 88.5 & 88.6 \\
% & InternVL2\textsubscript{40b}~\citep{internvl2}   & 93.0 & 94.7 & 89.2 & 88.5 & 92.8 & 83.6 & 90.3 & 90.6 \\
& InternVL2\textsubscript{76b}~\citep{internvl2}   & 92.2 & 94.8 & 88.4 & 88.8 & 93.1 & 82.8 & 89.5 & 90.3 \\
& Qwen2-VL\textsubscript{2b}     &	87.6 	&	90.6 	&	82.3 	&	79.0 	&	84.9 	&	71.0 	&	81.2 	&	80.3 	\\
& Qwen2-VL\textsubscript{7b}     &	91.7 	&	93.6 	&	87.3 	&	85.8 	&	90.5 	&	79.5 	&	87.3 	&	87.8 	\\
& Qwen2-VL\textsubscript{72b}    & \textbf{93.2} & \textbf{95.3} & 90.7 & \textbf{90.1} & \textbf{93.8} & \textbf{85.6} & \textbf{89.9} & \textbf{90.4} \\
\cmidrule(lr){1-10}
\multirow{3}{*}{\textit{Specialist}}
& G-DINO-L~\citep{grounding_dino}    &	90.6 	&	93.2 	&	88.2 	&	82.8 	&	89.0 	&	75.9 	&	86.1 	&	87.0 	\\
& UNINEXT-H~\citep{uninext}    &	92.6 	&	94.3 	&	\textbf{91.5}	&	85.2 	&	89.6 	&	79.8 	&	88.7 	&	89.4 	\\
& ONE-PEACE~\citep{one-peace}    &	92.6 	&	94.2 	&	89.3 	&	88.8 	&	92.2 	&	83.2 	&	89.2 	&	89.3 	\\
\bottomrule
\end{tabular}
}
\end{table}

\subsubsection{Video Understanding}

We evaluate our models on various video understanding tasks, with related benchmarks covering short videos of a few seconds to long videos of up to one hour. Table~\ref{tab:additional_performance_comparison} presents the performance of Qwen2-VL and baseline models. 
Overall, Qwen2-VL demonstrates strong results across 2B, 7B, and 72B sizes, with Qwen2-VL-72B achieving the best performance on MVBench~\citep{li2024mvbench}, PerceptionTest~\citep{patraucean2024perception}, and EgoSchema~\citep{mangalam2023egoschema}. 
This showcases Qwen2-VL's superior capabilities in video understanding tasks, and scaling up Qwen2-VL yields significant improvements. 
For the challenging Video-MME benchmark~\citep{fu2024video}, which includes videos up to one hour, it is noteworthy that we limited the maximum number of frames extracted per video to 768 during evaluation, potentially impacting performance on longer videos.
Future work will focus on extending Qwen2-VL to support longer sequences, thereby accommodating longer videos.

\subsubsection{Visual Agent}
Qwen2-VL is evaluated first for its ability to interact with the environment via function calls and then for its capacity to complete complex sequential decision tasks through multiple rounds of interaction. The implementation is based on the Qwen-Agent framework~\citep{qwen-agent}.

\paragraph{Function Calling}
Unlike function calling in LLMs~\citep{berkeley-function-calling-leaderboard,srinivasan2023nexusraven,chen2023t}, function calling in LVLMs often involves extracting information from visual cues. Due to the absence of public benchmarks for evaluating the capabilities of LVLMs in function calling, we constructed our internal evaluation dataset.

To construct the evaluation dataset, we undertook the following procedures~\citep{chen2023t}: Scene Categorization, Image Collection, Image Content Extraction, and Question/Functions/Arguments Generation. Firstly, we classified scenes into categories based on different visual applications. Subsequently, we downloaded and meticulously selected high-quality, representative images from the internet for each category. Thereafter, utilizing an advanced LVLM~\citep{qwenvl}, we analyzed each image to extract key visual elements and textual information. Finally, based on the content information from the images, we used an advanced LLM~\citep{qwen2} to generate a series of questions that required specific functions to answer, along with specifying the input parameters needed for these function calls.

Similar to the function calling evaluation method in LLMs~\citep{berkeley-function-calling-leaderboard}, we designed two metrics to evaluate the accuracy of the function selection and the correctness of the arguments input. Specifically, Type Match(TM), is calculated as the ratio of times the model successfully invoked the correct function to the total number of calls attempted. Exact Match(EM), for each function calling, we checked whether the arguments passed to the function exactly matched those recorded in the image's content information, calculating this correctness ratio.

As shown in Table~\ref{tab:comparison_v3}, the performance of Qwen2-VL in both Type Match(93.1 vs. 90.2) and Exact Match(53.2 vs. 50.0) over GPT-4o substantiates the efficacy of Qwen2-VL's capability in function calling, thereby underscoring its significant potential for application expansion through external tool integration.

The evaluation results demonstrated that GPT-4o underperformed, primarily due to two factors: in scenarios where uncertainty arises, GPT-4o demonstrates a conservative approach by avoiding using external tools. The Optical Character Recognition (OCR) capability of GPT-4o is outperformed by Qwen2-VL, particularly in the context of Chinese characters.

\paragraph{UI Operations/Games/Robotics/Navigation}

To assess Qwen2-VL’s ability to generally handle complex tasks, we conduct evaluations across multiple VL agent tasks, including mobile operations~\citep{zhang2024android,rawles2024androidinthewild,lu2024gui,rawles2024androidworld}, robotic control~\citep{kolve2017ai2,shridhar2020alfred,inoue2022prompter,lu2023thinkbot,jiang2022vima,huang2023instruct2act}, card games~\citep{zhai2024fine}, and vision-language navigation~\citep{anderson2018vision,qi2020reverie}. As these tasks need multiple actions to complete tasks, we keep the history (observation, action) through Qwen2-VL supports a 32K context length, then append each new observation image after every action, enabling continuous reasoning about subsequent steps. 

\textbf{UI Operations:} we evaluate Qwen2-VL using the AITZ task~\citep{zhang2024android}, which constructs a core clean test set derived from AITW~\citep{rawles2024androidinthewild}. Based on common operation patterns of phone, we define actions such as tap, input and swipe~\citep{rawles2024androidinthewild} for Qwen2-VL to interact with on-screen icons for task completion. For example, when Qwen2-VL is tasked with finding a pizza restaurant nearby by Google Maps, it should input "pizza" in the search term, swipe to select the appropriate restaurant, and tap the corresponding link. Following the AITZ setting, we report both type match (correctness of tap, input, or swipe) and exact match (correctness of tap location, input text, or swipe direction). With the support of grounding capability on UI, Qwen2-VL surpasses GPT-4 and previous SoTA~\citep{zhang2024android,zhan2023you}.

\textbf{Robotic Control:} we evaluate Qwen2-VL on the ALFRED task~\citep{shridhar2020alfred} in AI2THOR~\citep{kolve2017ai2}. The task requires agent to perform complex household tasks, such as toasting bread and slicing an apple to prepare a meal. To work in the virtual environment, we define high-level actions (GotoLocation, Pickup, PutDown, Open, Close, Clean, Heat, Cool, Slice)~\citep{shridhar2020alfworld} as the action set. Moreover, agent needs to localize objects for manipulation (e.g., it can only pick up an apple if the apple is recognized). To improve the accuracy of manipulation, we integrate SAM~\citep{kirillov2023segment}. ALFRED task reports task success rate (SR) (e.g., preparing dinner) and sub-goal completion metrics (GC) (e.g., whether the bread is toasted or the apple is sliced). Qwen2-VL slightly outperforms the previously specialized model ThinkBot~\citep{lu2023thinkbot} on the valid-unseen set.

\textbf{Card Games:} we leverage the card game environment from RL4VLM~\citep{zhai2024fine} to assess Qwen2-VL's performance in a series of card-based games: Number Line, BlackJack, EZPoint, and Point24. Each game presents distinct challenges: (1) reaching a target number using +1 or -1 operations, (2) drawing or holding cards to compete against the dealer, (3) applying basic arithmetic operations to reach a total of 12, and (4) using arithmetic operations to achieve a total of 24. We report the success rate of the tasks. They not only evaluate agent capabilities but also require strong OCR skills to recognize these cards and understand the progression of the game. Qwen2-VL demonstrates superior performance across all tasks.

\textbf{Vision-Language Navigation:} we evaluate Qwen2-VL on the Vision-and-Language Navigation (VLN) task using the R2R~\citep{anderson2018vision} and REVERIE~\citep{qi2020reverie}. In VLN, the model must autonomously determine the next location based on instruction, current observations. We report the success rate (SR) of VLM in reaching the predetermined destination for this task. The performance of Qwen2-VL is comparable to that of GPT-4o, but both models fall significantly behind current specialized VLN models~\citep{chen2022think,sigurdsson2023rrex}. We attribute this gap to the incomplete and unstructured map information generated by the model from multiple images. Accurately modeling maps and locations in a 3D environment remains a major challenge for multimodal models.

\subsection{Ablation Study}

In this section, we present ablation studies on image dynamic resolution, M-RoPE, and model scale. These experiments aim to provide insights into the impact of these key components on our model's performance.

\subsubsection{Dynamic Resolution}

\begin{table}[t] 
\centering 
    \caption{Qwen2-VL-7B under fixed/dynamic image tokens. Adjusting image sizes only results in small perturbations in performance, demonstrating the robustness to varying image sizes. Moreover, the dynamic resolution strategy achieves top-tier performance while consuming fewer tokens on average, demonstrating the efficiency of our model.}
    \label{tab:fix_dyn_resolution}
    \scalebox{0.9}{
        \begin{tabular}{ c | c | c | c | c | c }
        \toprule
        \textbf{Strategy} & \textbf{Average Image Tokens} & \textbf{InfoVQA\textsubscript{val}} & \textbf{RealWorldQA}  & \textbf{OCRBench} & \textbf{MMMU} \\
        \hline
        \multirow{4}{*}{Fixed Image Tokens}
        & 64  & 28.85 & 56.47 & 572 & 53.33  \\
        & 576 & 65.72 & 65.88 & 828 & 52.78  \\
        & 1600 & 74.99 & 69.54 & 824 & 52.89 \\
        & 3136 & 77.27 & 70.59 & 786 & 53.44 \\
        \hline
        Dynamic Image Tokens& 1924 & 75.89 & 70.07 & 866 & 53.44 \\
        \bottomrule
        \end{tabular}
        }
\end{table}

\begin{figure*}[t]
\centering
\includegraphics[width= 1\linewidth]{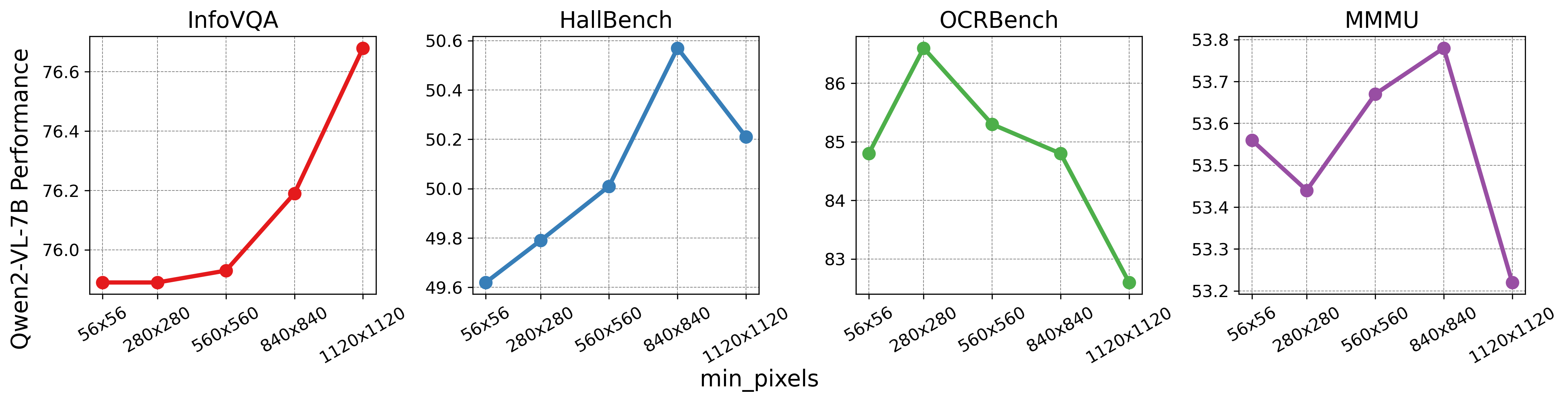}
   \caption{Qwen2-VL-7B with different min\_pixels. Small images are upscaled to surpass a specified min\_pixels threshold before input into the model. Increasing the image size within a reasonable range shows enhanced performance on perceptual tasks like InfoVQA, HallusionBench, and OCRBench.}
\label{fig:minpixels_resolution}
\end{figure*}

As shown in Table~\ref{tab:fix_dyn_resolution}, we compare the performance between dynamic resolution and fixed resolution. For fixed resolution, we resize the images to ensure a constant number of image tokens being input to the model, rather than resizing to a specific height and width, as this would distort the original aspect ratio. For dynamic resolution, we only set min\_pixels$=100\times28\times28$ and max\_pixels$=16384\times28\times28$, allowing the number of image tokens depend primarily on the image's native resolution. It can be observed that adjusting image sizes only results in small perturbations in performance, demonstrating the model robustness to varying image sizes. Moreover, dynamic resolution approach is more efficient. We can observe that no single fixed resolution achieves optimal performance across all benchmarks. In contrast, the dynamic resolution approach consistently achieves top-tier performance while consuming fewer tokens on average.

Additionally, we observe that merely increasing the image size does not always lead to improved performance. It is more important to choose an appropriate resolution for different images. As detailed in Figure~\ref{fig:minpixels_resolution}, we upscale small images to surpass a specified min\_pixels threshold. Evaluations on upscaled images shows enhanced performance on perceptual tasks like InfoVQA, HallusionBench, and OCRBench. We attribute these gains to increased computational load. However, for OCRBench, a too-high min\_pixels value leads to a severe performance decline. This is likely because OCRBench contains numerous extremely small images, and excessive enlargement causes these images to deviate from the training data distribution, turning them into out-of-distribution samples. In contrast, the effect of increasing min\_pixels on the MMMU benchmark is negligible. We hypothesize that the performance bottleneck in MMMU is more related to the model's reasoning capability rather than image resolution.

\subsubsection{M-RoPE}

\begin{table}[t] 
\centering 
\caption{Ablation studies of M-RoPE. Compared to 1D-RoPE, using M-RoPE achieves better performance in downstream tasks, particularly in video benchmarks. RWQ means RealworldQA.} 
\label{tab:abla_mrope}
\scalebox{0.75}{
\begin{tabular}{l|c|c|c|c|c|c|c|c|c|c|c} 
\toprule 
  & \multicolumn{8}{c|}{Image Benchmarks} &\multicolumn{3}{c}{Video Benchmarks} \\
\midrule
  & MathVista & MMB & MMStar & RWQ & DocVQA & ChartQA & InfoVQA & TextVQA & PerceptionTest &NextQA & STAR \\ 
\midrule 
1D-RoPE & 39.2 & 58.6 & \textbf{36.7} & \textbf{54.5} & 82.5 & 68.0 & \textbf{50.8} & 71.3 & 46.6 & 43.9 & 55.5 \\ 
M-RoPE & \textbf{43.4} & \textbf{60.6} & \textbf{36.7} & 53.7 & \textbf{82.8} & \textbf{68.4} & 50.3 & \textbf{71.8} & \textbf{47.4} & \textbf{46.0} & \textbf{57.9} \\
\bottomrule 
\end{tabular} 
}
\end{table}

\begin{figure*}[t]
\centering
\includegraphics[width=1\linewidth]{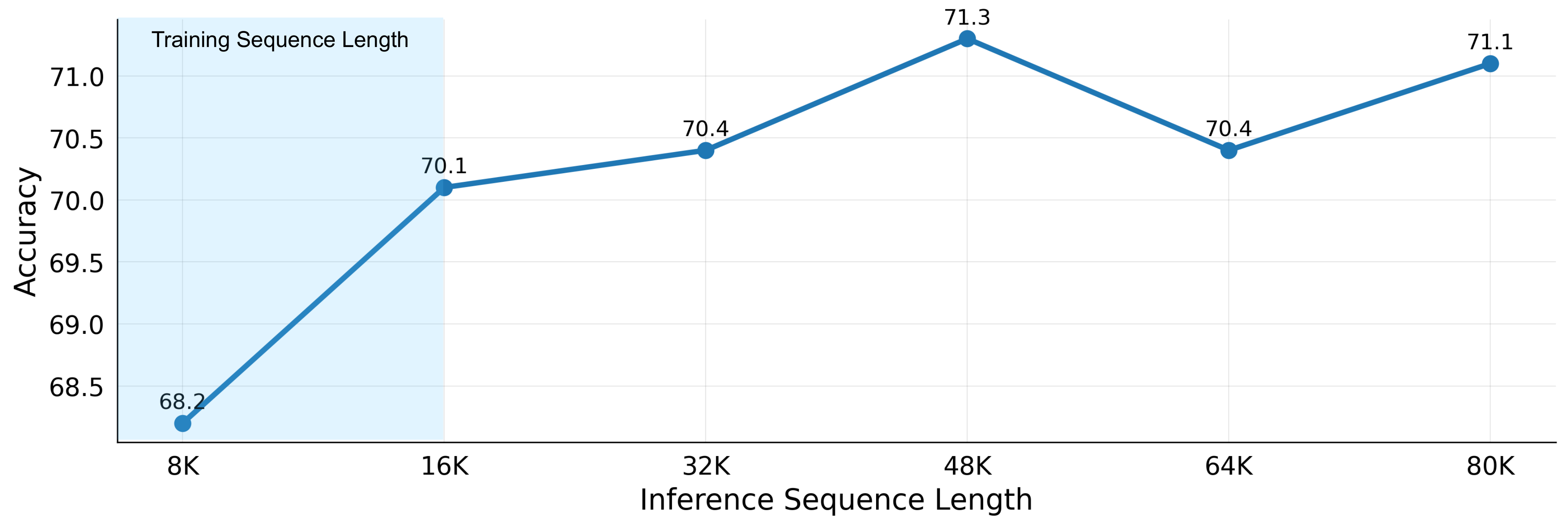}
   \caption{Evaluate the length extrapolation capability of Qwen2-VL-72B on Video-MME Medium Video. With the help of M-RoPE, the model demonstrated robust performance when the inference length exceeded the maximum training length of 16384 tokens.}
\label{fig:mrope_extropolate}
\end{figure*}

In this subsection, we demonstrate the effectiveness of M-RoPE. First, we validate its capability on various downstream tasks. We employ Qwen2-1.5B and ViT-L as the backbone and report the results of the pre-trained models. As shown in Table~\ref{tab:abla_mrope}, compared to 1D-RoPE, using M-RoPE achieves better performance in downstream tasks, particularly in video benchmarks. Furthermore, we assess the length extrapolation capability of M-RoPE on Video-MME medium-length videos. Figure~\ref{fig:mrope_extropolate} illustrates the performance of Qwen2-VL-72B at different inference lengths. Leveraging M-RoPE, the model demonstrates robust results across various inference lengths. Notably, despite limiting the maximum tokens per video to 16K during training, the model still exhibits exceptional performance at a maximum inference length of 80K tokens.

\subsubsection{Model Scaling}
\begin{figure*}[t]
\centering
\includegraphics[width= 1\linewidth]{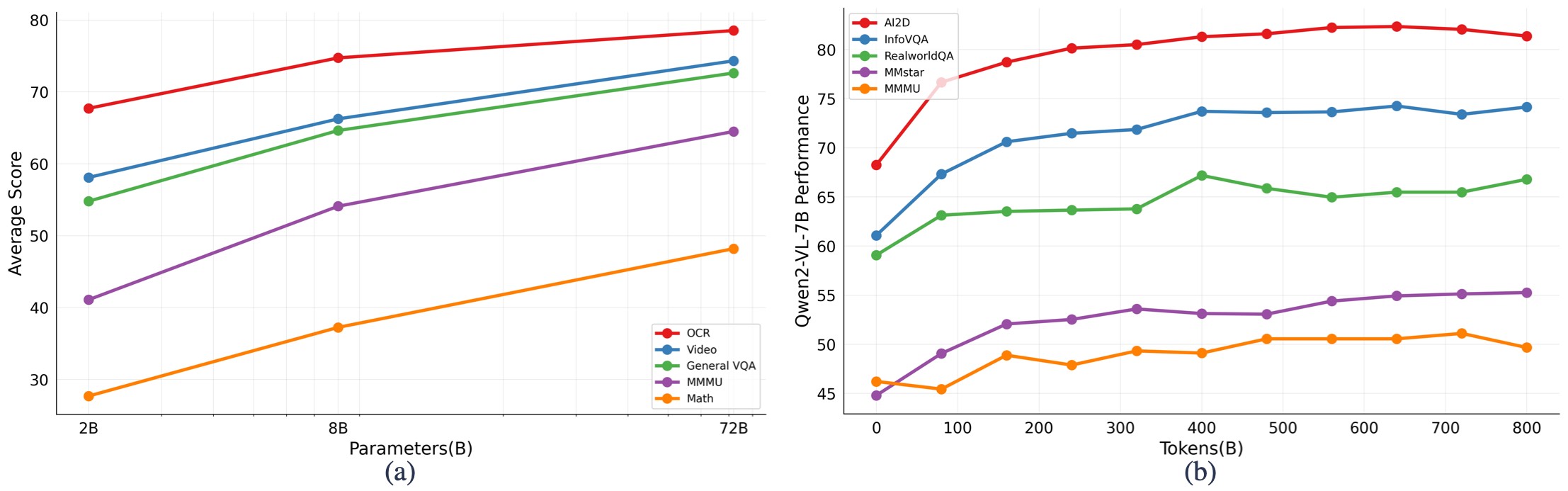}
   \caption{Model Performance Scaling Across Capabilities and Training Progress. As model size and the volume of training data increase, performance consistently improves across a range of capabilities and benchmarks.}
\label{fig:scale}
\end{figure*}

We evaluate the performance of models of varying scales across multiple capability dimensions. Specifically, we categorize these dimensions into complex college-level problem-solving, mathematical abilities, document and table comprehension, general scenario question-answering, and video comprehension. The overall capability of a model is assessed by averaging its scores across different benchmarks associated with each dimension.

In particular, we use the MMMU~\citep{yue2023mmmu} benchmark to represent college-level problem-solving ability, while the average scores from MathVista~\citep{mathvista} and MathVision~\citep{mathvision} serve as indicators of mathematical ability. For general scenario question-answering, we compute the average score across the RealWorldQA~\citep{grok15}, MMBench-V1.1~\citep{MMBench}, MMT-Bench~\citep{mmtbench}, HallBench~\citep{guan2023hallusionbench}, MMVet~\citep{yu2024mm}, and MMStar~\citep{chen2024we} benchmarks. Document and table comprehension capability is reflected through the average score from benchmarks like DocVQA~\citep{docvqa}, InfoVQA~\citep{docvqa}, ChartQA~\citep{masry2022chartqa}, TextVQA~\citep{textvqa}, OCRBench~\citep{liu2024ocrbenchhiddenmysteryocr}, and MTVQA~\citep{tang2024mtvqa}. Lastly, video comprehension ability is measured by averaging scores across MVBench~\citep{li2024mvbench}, PerceptionTest~\citep{patraucean2024perception}, EgoSchema~\citep{mangalam2023egoschema}, and Video-MME~\citep{fu2024video}.

As illustrated in Figure~\ref{fig:scale}(a), there is a consistent improvement in performance with increasing model size, particularly with respect to mathematical abilities, which show a positive correlation with the number of model parameters. On the other hand, for optical character recognition (OCR)-related tasks, even smaller-scale models exhibit relatively strong performance.

As shown in Figure~\ref{fig:scale}(b), we visualize the relationship between model performance and the number of training tokens during the second stage of pretraining for Qwen2-VL-7B. As the number of training tokens increases, the model performance improves; however, performance on vision question answering (VQA) tasks exhibits some fluctuation. In contrast, for tasks such as AI2D~\citep{kembhavi2016diagram} and InfoVQA~\citep{docvqa}—both of which involve understanding textual and graphical information in images—the model performance shows steady improvement as training data is augmented.

\section{Conclusion}

We have presented the Qwen2-VL series, the versatile large vision-language models, including three open-weight models with total parameter counts of 2, 8, and 72 billion. Qwen2-VL matches the performance of top-tier models like GPT-4o and Claude3.5-Sonnet in a range of multimodal scenarios, surpassing all other open-weight LVLM models. Qwen2-VL series introduces naive dynamic resolution and multimodal rotary position embedding (M-RoPE) to fuse information across modals effectively and be capable of understanding videos over 20 minutes in length. With advanced reasoning and decision-making abilities, Qwen2-VL can be integrated with devices such as mobile phones, robots, etc. Furthermore, Qwen2-VL now supports understanding multilingual texts within images, including most European languages, Japanese, Korean, Arabic, Vietnamese, and others.

We have made the Qwen2-VL model weights openly accessible, which enables researchers and developers to harness the full potential in a variety of applications and research projects. We aim to advance AI technologies and enhance their beneficial effects on society by dedicating ourselves to these endeavors.

\section*{Acknowledgements}

We express our gratitude to Juan Zhu, Fan Hong, Jie Zhang, Yong Li of  Alibaba Cloud’s PAI team~\citep{pai-lingjun} for supporting the training infrastructure of Qwen2-VL. This work was also supported by Qwen LLM team~\citep{qwen2}, and we especially thank Na Ni, Yichang Zhang, Jianxin Ma, Bowen Yu, Zheren Fu for their data contribution and insightful discussion.

\bibliographystyle{plainnat} 
\bibliography{references}
\clearpage
\appendix

\section{Model Capabilities and Qualitative Examples}
In this section, we present some practical examples of our Qwen2-VL.

\subsection{General Chat and OCR}

The Qwen2-VL models are now more adept at accurately describing and identifying complex information within images, as well as providing detailed background and answering related questions. Besides, the text processing capabilities of the Qwen2-VL models have seen significant improvements, particularly concerning the recognition of Chinese and English text within images.

% 1_3.json
\begin{figure}[htbp]
\centering
\begin{tcolorbox}[colback=black!5!white,colframe=black!75!black,title=Multi-object Recognition]
\includegraphics[width=9cm]{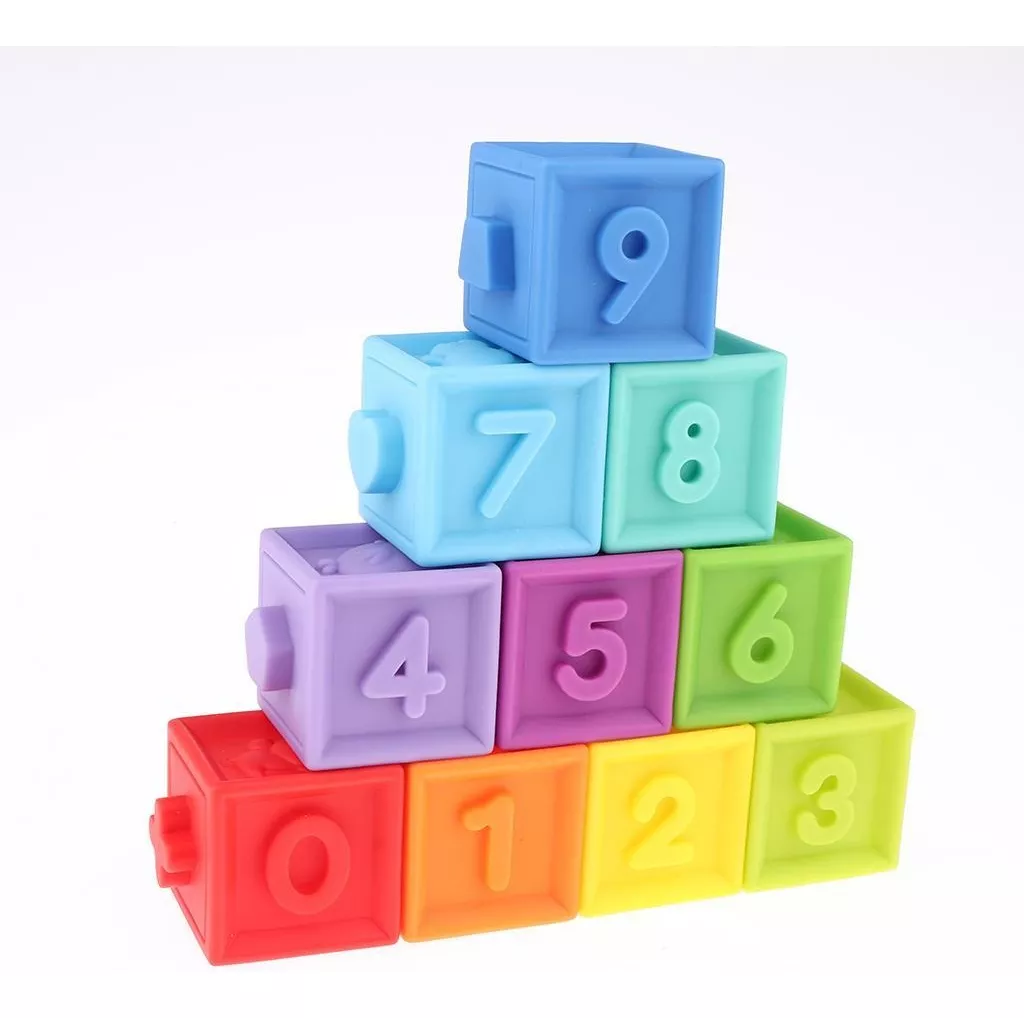}

Output the color and number of each box.
\tcbsubtitle{Model Response}
The image shows a stack of colorful blocks with numbers on them. Here is the color and number of each block from top to bottom:

- Top row (single block): Blue with the number 9
- Second row (two blocks): Light blue with the number 7, and green with the number 8
- Third row (three blocks): Purple with the number 4, pink with the number 5, and light green with the number 6
- Bottom row (four blocks): Red with the number 0, orange with the number 1, yellow with the number 2, and light green with the number 3
\label{mt_format}
\end{tcolorbox}
\caption{When presented with an image of cubes of different colors, the models identify their layout and the color of each cube.}
\end{figure}

% 1_4.json

\begin{figure}[htbp]
\centering
\begin{tcolorbox}[colback=black!5!white,colframe=black!75!black,title=Plant identification]
\includegraphics[width=15cm]{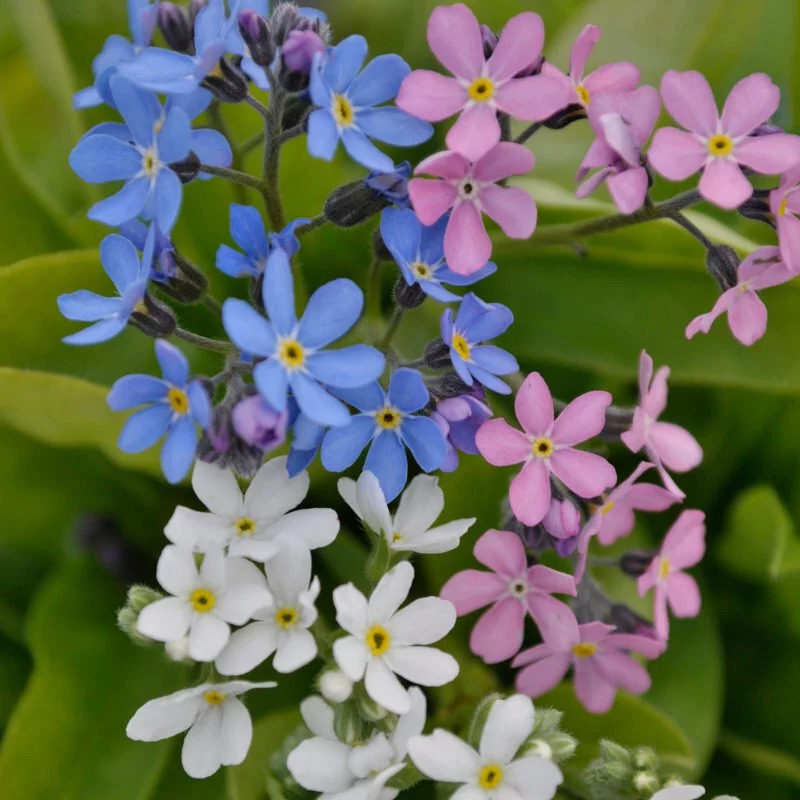}

What kind of flower is this?
\tcbsubtitle{Model Response}
The flowers in the picture are forget-me-nots (Myosotis). These small, delicate flowers are known for their vibrant colors, typically blue, pink, or white, and are often used in bouquets and gardens for their charming appearance.
\label{mt_format}
\end{tcolorbox}
\caption{The model displays an adeptness in recognizing flowers in photographs.}
\end{figure}

% 1_6.json

\begin{figure}[htbp]
\centering
\begin{tcolorbox}[colback=black!5!white,colframe=black!75!black,title=Document Parsing with Dense Formulas]
\includegraphics[width=7cm]{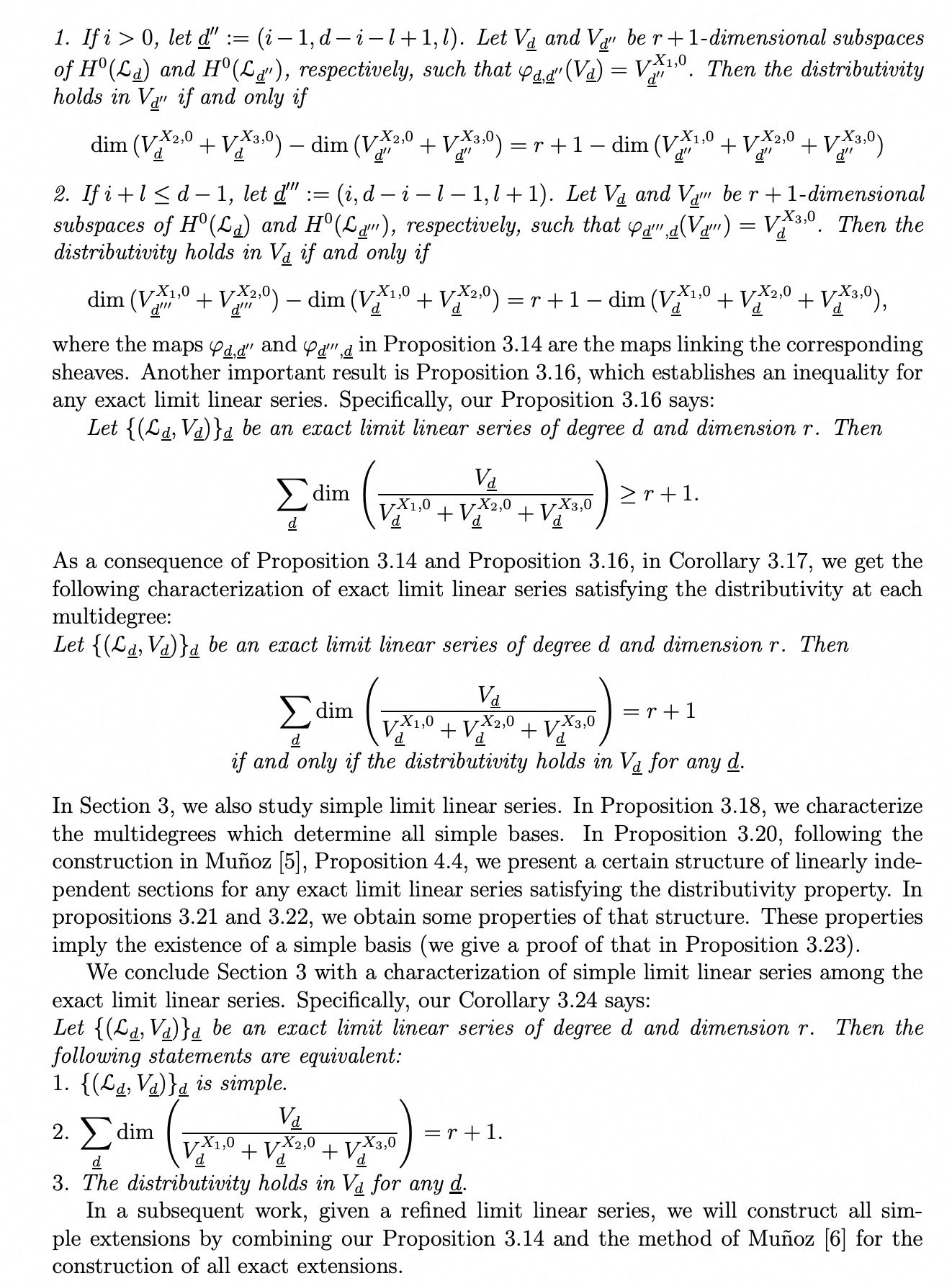}

{\tiny Convert the content in the image to Markdown}
\tcbsubtitle{Model Response}

{\tiny 1. If $i > 0$, let $\underline{d}'' := (i-1, d-i-l+1, l)$. Let $V_{\underline{d}}$ and $V_{\underline{d}''}$ be $r+1$-dimensional subspaces of $H^0(\mathcal{L}_{\underline{d}})$ and $H^0(\mathcal{L}_{\underline{d}''})$, respectively, such that $\varphi_{\underline{d}, \underline{d}''}(V_{\underline{d}}) = V_{\underline{d}''}^{X_1, 0}$. Then the distributivity holds in $V_{\underline{d}''}$ if and only if
$$
\dim (V_{\underline{d}}^{X_2, 0} + V_{\underline{d}}^{X_3, 0}) - \dim (V_{\underline{d}''}^{X_2, 0} + V_{\underline{d}''}^{X_3, 0}) = r + 1 - \dim (V_{\underline{d}''}^{X_1, 0} + V_{\underline{d}''}^{X_2, 0} + V_{\underline{d}''}^{X_3, 0})
$$

2. If $i + l \leq d - 1$, let $\underline{d}''' := (i, d-i-l-1, l+1)$. Let $V_{\underline{d}}$ and $V_{\underline{d}'''}$ be $r+1$-dimensional subspaces of $H^0(\mathcal{L}_{\underline{d}})$ and $H^0(\mathcal{L}_{\underline{d}'''})$, respectively, such that $\varphi_{\underline{d}''', \underline{d}}(V_{\underline{d}'''}) = V_{\underline{d}}^{X_3, 0}$. Then the distributivity holds in $V_{\underline{d}}$ if and only if
$$
\dim (V_{\underline{d}'''}^{X_1, 0} + V_{\underline{d}'''}^{X_2, 0}) - \dim (V_{\underline{d}}^{X_1, 0} + V_{\underline{d}}^{X_2, 0}) = r + 1 - \dim (V_{\underline{d}}^{X_1, 0} + V_{\underline{d}}^{X_2, 0} + V_{\underline{d}}^{X_3, 0}),
$$
where the maps $\varphi_{\underline{d}, \underline{d}''}$ and $\varphi_{\underline{d}''', \underline{d}}$ in Proposition 3.14 are the maps linking the corresponding sheaves. Another important result is Proposition 3.16, which establishes an inequality for any exact limit linear series. Specifically, our Proposition 3.16 says:

Let $\{(\mathcal{L}_{\underline{d}}, V_{\underline{d}})\}_{\underline{d}}$ be an exact limit linear series of degree $d$ and dimension $r$. Then
$$
\sum_{\underline{d}} \dim \left( \frac{V_{\underline{d}}}{V_{\underline{d}}^{X_1, 0} + V_{\underline{d}}^{X_2, 0} + V_{\underline{d}}^{X_3, 0}} \right) \geq r + 1.
$$

As a consequence of Proposition 3.14 and Proposition 3.16, in Corollary 3.17, we get the following characterization of exact limit linear series satisfying the distributivity at each multidegree:

Let $\{(\mathcal{L}_{\underline{d}}, V_{\underline{d}})\}_{\underline{d}}$ be an exact limit linear series of degree $d$ and dimension $r$. Then
$$
\sum_{\underline{d}} \dim \left( \frac{V_{\underline{d}}}{V_{\underline{d}}^{X_1, 0} + V_{\underline{d}}^{X_2, 0} + V_{\underline{d}}^{X_3, 0}} \right) = r + 1
$$
if and only if the distributivity holds in $V_{\underline{d}}$ for any $\underline{d}$.

In Section 3, we also study simple limit linear series. In Proposition 3.18, we characterize the multidegrees which determine all simple bases. In Proposition 3.20, following the construction in Muñoz [5], Proposition 4.4, we present a certain structure of linearly independent sections for any exact limit linear series satisfying the distributivity property. In propositions 3.21 and 3.22, we obtain some properties of that structure. These properties imply the existence of a simple basis (we give a proof of that in Proposition 3.23).

We conclude Section 3 with a characterization of simple limit linear series among the exact limit linear series. Specifically, our Corollary 3.24 says:

Let $\{(\mathcal{L}_{\underline{d}}, V_{\underline{d}})\}_{\underline{d}}$ be an exact limit linear series of degree $d$ and dimension $r$. Then the following statements are equivalent:
1. $\{(\mathcal{L}_{\underline{d}}, V_{\underline{d}})\}_{\underline{d}}$ is simple.
2. $\sum_{\underline{d}} \dim \left( \frac{V_{\underline{d}}}{V_{\underline{d}}^{X_1, 0} + V_{\underline{d}}^{X_2, 0} + V_{\underline{d}}^{X_3, 0}} \right) = r + 1$.
3. The distributivity holds in $V_{\underline{d}}$ for any $\underline{d}$.

In a subsequent work, given a refined limit linear series, we will construct all simple extensions by combining our Proposition 3.14 and the method of Muñoz [6] for the construction of all exact extensions.}

\label{literary_writing}
\end{tcolorbox}
\vspace{-1em}
\caption{Literary writing in multiple languages based on visual stimuli.} 
\end{figure}

% 1_7.json

\begin{figure}[htbp]
\centering
\vspace{-1cm}
\begin{tcolorbox}[colback=black!5!white,colframe=black!75!black,title=Multilingual Text Recognition]
\includegraphics[width=6cm]{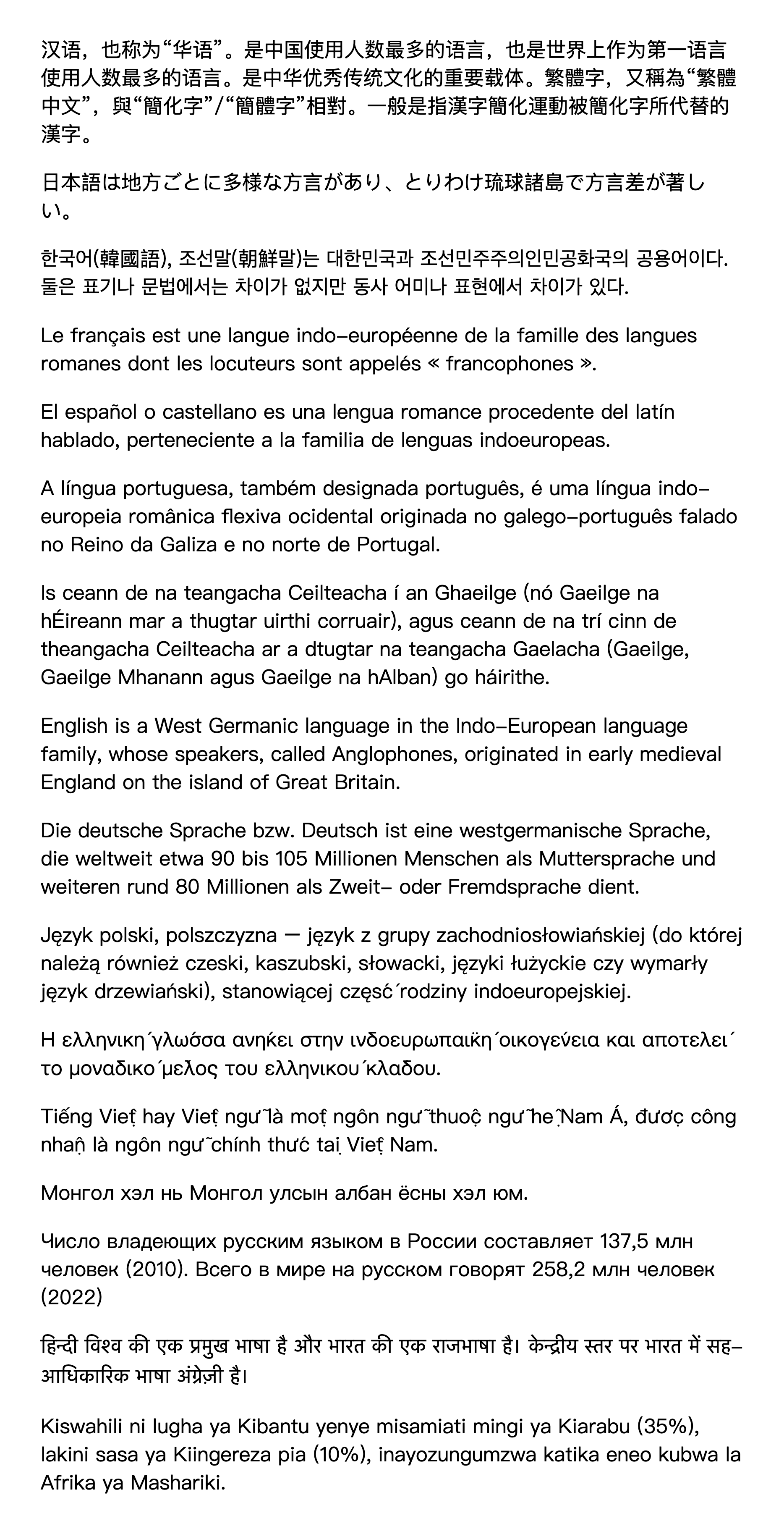}

\begin{CJK*}{UTF8}{gbsn}
{\scriptsize 转录图中内容，并识别图中每段内容的语言。}

{\scriptsize (Translation: Transcribe the picture and identify the language of each paragraph in it.)}
\end{CJK*}

\tcbsubtitle{Model Response}

\begin{CJK*}{UTF8}{gbsn}

\scriptsize 1. 汉语，也称为“华语”。是中国使用人数最多的语言，也是世界上作为第一语言使用人数最多的语言。是中华优秀传统文化的重要载体。繁体字，又稱為“繁體中文”，與“簡化字”/“簡體字”相對。一般是指漢字簡化運動被簡化字所代替的漢字。
   - 语言：中文

\scriptsize 2. 日本語は地方ごとに多様な方言があり、とりわけ琉球諸島で方言差が著しい。
   - 语言：日语
\end{CJK*}

\begin{CJK*}{UTF8}{mj}
\scriptsize 3. 한국어(韓國語), 조선말(朝鮮말)는 대한민국과 조선민주주의인민공화국의 공용어이다. 둘은 표기나 문법에서는 차이가 없지만 동사 어미나 표현에서 차이가 있다.
\end{CJK*}
\begin{CJK*}{UTF8}{gbsn}
\scriptsize    - 语言：韩语
\end{CJK*}

\begin{CJK*}{UTF8}{gbsn}
\scriptsize 4. Le français est une langue indo-européenne de la famille des langues romanes dont les locuteurs sont appelés « francophones ».
   - 语言：法语

\scriptsize 5. El español o castellano es una lengua romance procedente del latín hablado, perteneciente a la familia de lenguas indoeuropeas.
   - 语言：西班牙语

\scriptsize 6. A língua portuguesa, também designada português, é uma língua indo-europeia românica flexiva ocidental originada no galego-português falado no Reino da Galiza e no norte de Portugal.
   - 语言：葡萄牙语

\scriptsize 7. Is ceann de na teangacha Ceilteacha í an Ghaeilge (nó Gaeilge na hÉireann mar a thugtar uirthi corruair), agus ceann de na trí cinn de theangacha Ceilteacha ar a dtugtar na teangacha Gaelacha (Gaeilge, Gaeilge Mhanann agus Gaeilge na hAlban) go háirithe.
   - 语言：爱尔兰语

\scriptsize 8. English is a West Germanic language in the Indo-European language family, whose speakers, called Anglophones, originated in early medieval England on the island of Great Britain.
   - 语言：英语

\scriptsize 9. Die deutsche Sprache bzw. Deutsch ist eine westgermanische Sprache, die weltweit etwa 90 bis 105 Millionen Menschen als Muttersprache und weiteren rund 80 Millionen als Zweit- oder Fremdsprache dient.
   - 语言：德语

\scriptsize 10. Język polski, polszczyzna — język z grupy zachodniosłowiańskiej (do której należą również czeski, kaszubski, słowacki, języki łużyckie czy wymarły język drzewiański), stanowiącej część rodziny indoeuropejskiej.
    - 语言：波兰语
\end{CJK*}

\begin{otherlanguage}{greek}
\scriptsize 11. Η ελληνική γλώσσα ανήκει στην ινδοευρωπαϊκή οικογένεια και αποτελεί το μοναδικό μέλος του ελληνικού κλάδου.
\end{otherlanguage}
\begin{CJK*}{UTF8}{gbsn}
\scriptsize     - 语言：希腊语
\end{CJK*}

\begin{otherlanguage}{vietnamese}
\scriptsize 12. Tiếng Việt hay Việt ngữ là một ngôn ngữ thuộc ngữ hệ Nam Á, được công nhận là ngôn ngữ chính thức tại Việt Nam.
\end{otherlanguage}
\begin{CJK*}{UTF8}{gbsn}
\scriptsize    - 语言：越南语
\end{CJK*}

\begin{otherlanguage}{mongolian}
\scriptsize 13. Монгол хэл нь Монгол улсын албан ёсны хэл юм.
\end{otherlanguage}
\begin{CJK*}{UTF8}{gbsn}
\scriptsize     - 语言：蒙古语
\end{CJK*}

%\begin{otherlanguage}{russian}
%\scriptsize 14. Число владеющих русским языком в России составляет 137,5 млн человек (2010). Всего в мире на русском говорят 258,2 млн человек (2022)
%\end{otherlanguage}
\begin{CJK*}{UTF8}{gbsn}
\scriptsize     - 语言：俄语

%15. हिन्दी विश्व की एक प्रमुख भाषा है और भारत की एक राजभाषा है। केन्द्रीय स्तर पर भारत में सह-आधिकारिक भाषा अंग्रेज़ी है।
%    - 语言：印地语

\scriptsize 16. Kiswahili ni lugha ya Kibantu yenye misamiati mingi ya Kiarabu (35\%), lakini sasa ya Kiingereza pia (10\%), inayozungumzwa katika eneo kubwa la Afrika ya Mashariki.
    - 语言：斯瓦希里语

\end{CJK*}
\label{mt_format}
\end{tcolorbox}
\caption{The model displays an adeptness in recognizing multilingual texts in images.}
\end{figure}

\begin{figure}[htbp]
\centering
\begin{tcolorbox}[colback=black!5!white,colframe=black!75!black,title=Text Information Recognition \& Processing]
\includegraphics[width=5cm]{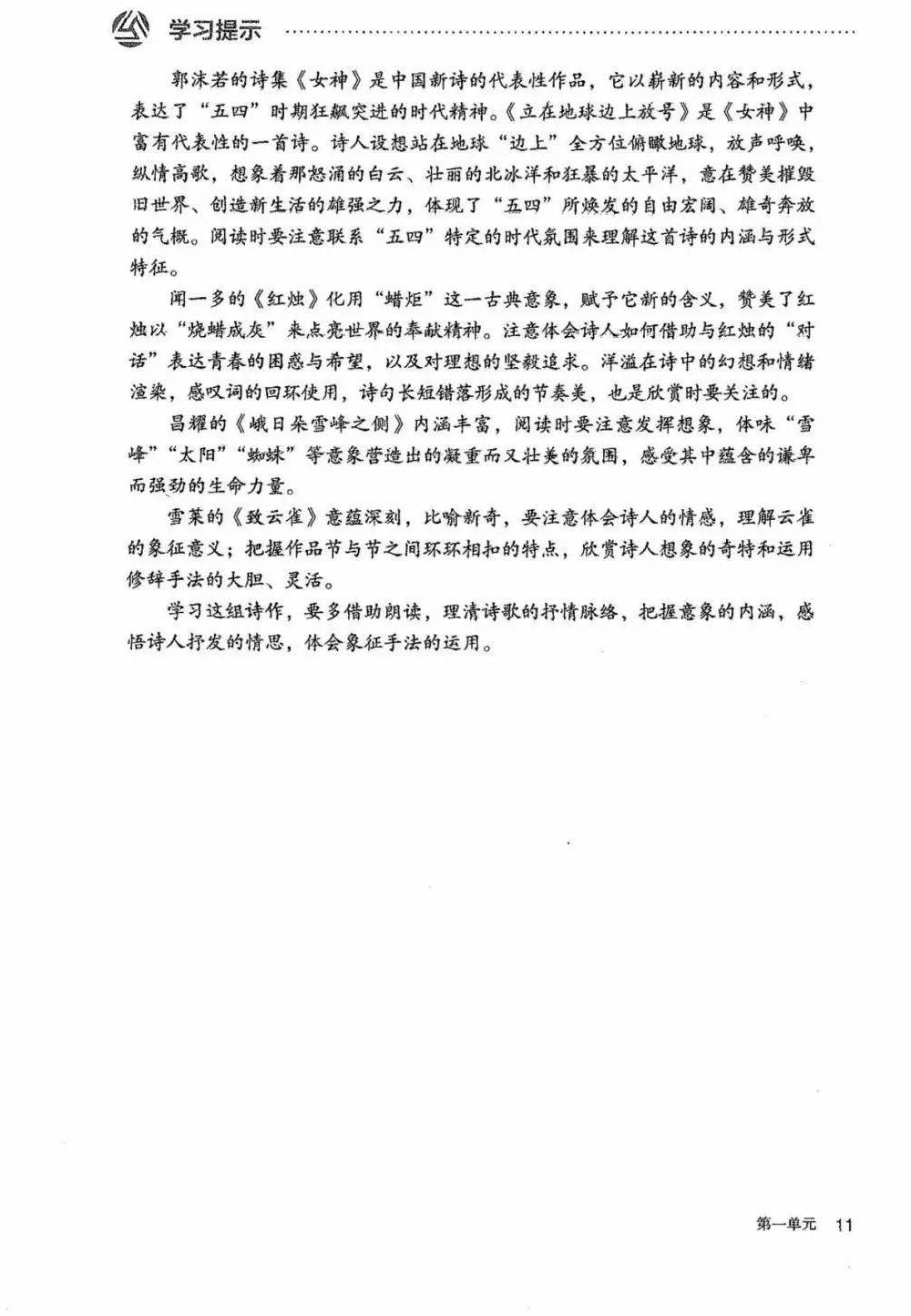}

Translate all the text in the picture into English and standardize the format
\tcbsubtitle{Model Response}
{\small The text in the image reads:

Learning Prompt

Guo Moruo's collection of poems, "The Goddess", is a representative work of modern Chinese poetry. It expresses new content and forms, embodying the spirit of the May Fourth Movement. "Standing on the Edge of the Earth" is a representative poem from "The Goddess". The poet imagines standing on the edge of the earth to observe it from all angles, shouting and singing, imagining the surging clouds, magnificent Arctic ice, and violent Pacific Ocean, praising the power to destroy the old world and create a new life. This poem embodies the free, and bold spirit of the May Fourth Movement. When reading, pay attention to connecting with the specific context of the May Fourth Movement to understand the content and form of this poem.  \\

Wen Yiduo's "Red Candle" uses the classical image of "candle" to give it a new meaning, praising the red candle for its dedication to lighting up the world. Pay attention to how the poet uses the dialogue between "red candle" and "red candle" to express the confusion and hope of youth, as well as the firm pursuit of ideals. The use of fantasy and emotional rhetoric, the use of rhetorical devices such as alliteration and assonance, and the formation of rhythmic beauty through the arrangement of long and short sentences in the poem are also important aspects to appreciate. \\

Chang Yao's "Snow Capped Peak on the Side of the Sun" is rich in content. When reading, pay attention to the development of imagination and taste the heavy and majestic atmosphere created by the images of "snow-capped peak", "sun", and "spider". Feel the powerful contained in it. \\

Shelley's "To a Skylark" has profound implications and novel imagery. Pay attention to the poet's emotions and understand the symbolic meaning of the skylark. Grasp the characteristics of the interlocking between sections of the work and appreciate the poet's use of bold and flexible imagery. \\

When learning this group of poems, rely more on reading aloud to clarify the emotional context of the poem and grasp the connotation of the image. Feel the emotions expressed by the poet and understand the use of symbolism.}
\label{Information}
\end{tcolorbox}
\caption{Recognition of dense Chinese text. The model is capable of directly converting lengthy, closely packed Chinese characters into standard English with accuracy and fluency.}
\end{figure}

% 1_1.json

\begin{figure}[htbp]
\centering
\begin{tcolorbox}[colback=black!5!white,colframe=black!75!black,title=Multilingual Text Recognition]
\includegraphics[width=15cm]{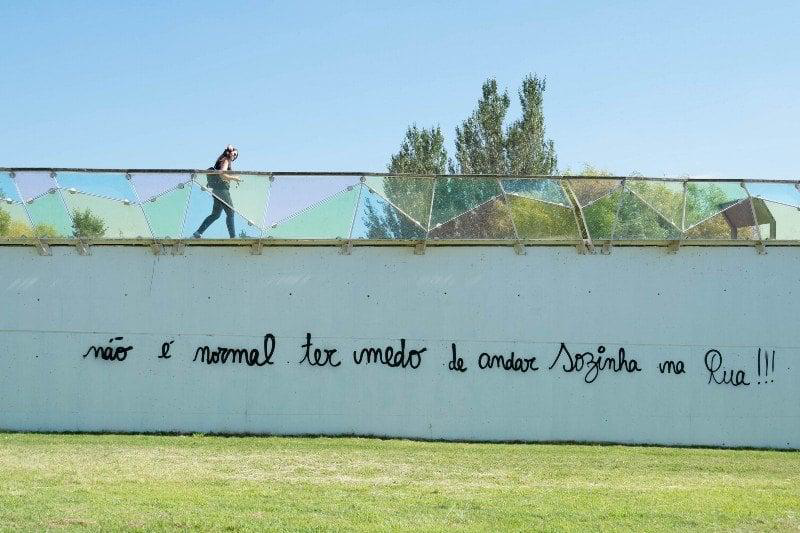}

What is the text?
\tcbsubtitle{Model Response}

The text on the wall is in Portuguese and reads: ``não é normal ter medo de andar sozinha na rua!!!'' This translates to: ``It is not normal to be afraid to walk alone on the street!!!''

\label{mt_format}
\end{tcolorbox}
\caption{The model displays an adeptness in recognizing multilingual texts in images. Image source: \citep{yang2023dawn}}
\end{figure}

% 1_5.json

\begin{figure}[htbp]
\centering
\begin{tcolorbox}[colback=black!5!white,colframe=black!75!black,title=Multilingual Text Recognition]
\includegraphics[width=12cm]{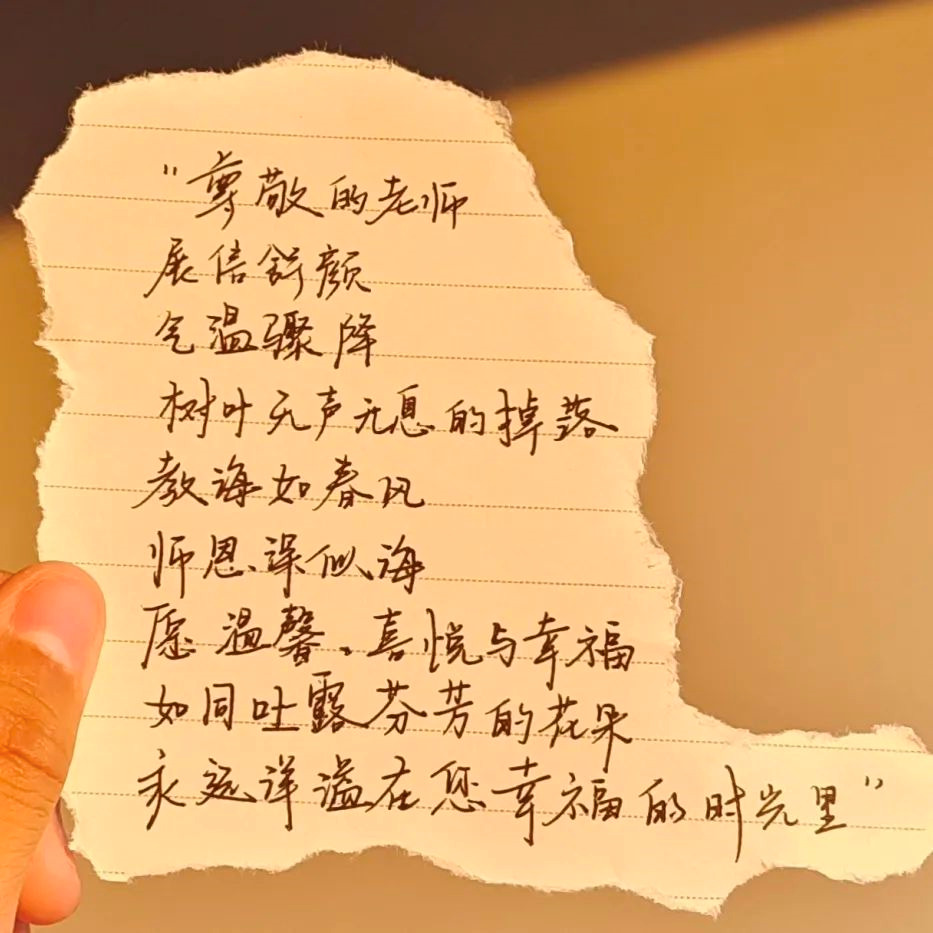}

\begin{CJK*}{UTF8}{gbsn}
识别图中所有文字，无需添加标点。

(Translation: Recognize all the text in the image without adding punctuation.)
\end{CJK*}
\tcbsubtitle{Model Response}

\begin{CJK*}{UTF8}{gbsn}
尊敬的老师

展信舒颜

气温骤降

树叶无声无息的掉落

教诲如春风

师恩深似海

愿温馨、喜悦与幸福

如同吐露芬芳的花朵

永远洋溢在您幸福的时光里
\end{CJK*}

\label{mt_format}
\end{tcolorbox}
\caption{The model displays an adeptness in recognizing multilingual texts in images.}
\end{figure}

\clearpage

\subsection{Information extraction and Visual Reasoning}

A notable advancement in the Qwen2-VL models is their enhanced visual reasoning capability. This advancement allows the models to interpret and comprehend complex representations such as flowcharts, diagrams, and other symbolic systems. 

\begin{figure}[htbp]
\centering
\begin{tcolorbox}[colback=black!5!white,colframe=black!75!black,title=Mathematical Problem Solving]
\includegraphics[width=10cm]{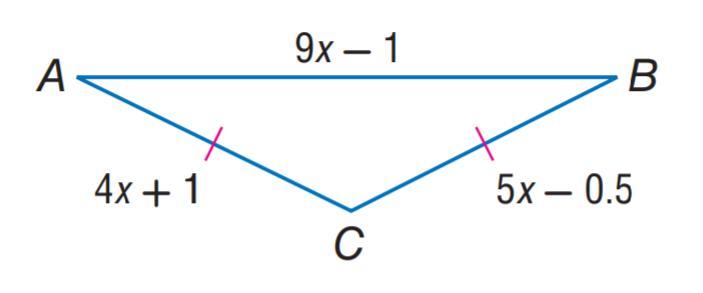}

Find the length of $AC$ in the isosceles triangle $ABC$.
\tcbsubtitle{Model Response}
Since triangle $ABC$ is isosceles, we know that $AC = BC$. Therefore, we can set the expressions for $AC$ and $BC$ equal to each other and solve for $x$

$4x + 1 = 5x - 0.5$ 

Subtracting $4x$ from both sides, we get:

$1 = x - 0.5$

Adding $0.5$ to both sides, we get:

$1.5 = x$

Now that we know the value of $x$, we can substitute it back into the expression for $AC$ to find its length:

$AC = 4x + 1$

$AC = 4(1.5) + 1$

$AC = 6 + 1$

$AC = 7$

Therefore, the length of $AC$ is $7$.
\label{}
\end{tcolorbox}
\caption{The model displays an adeptness in solving mathematical problems. Image source: \cite{lu2021inter}}
\end{figure}

\begin{figure}[htbp]
\centering
\begin{tcolorbox}[colback=black!5!white,colframe=black!75!black,title=Mathematical Problem Solving]
\includegraphics[width=10cm]{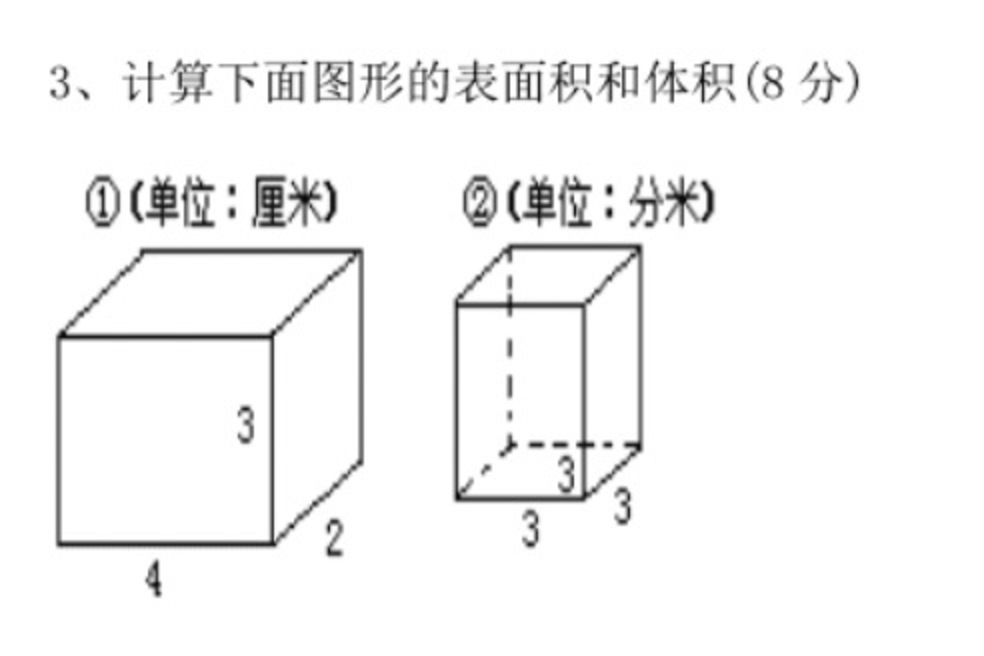}

Read the image and solve it step by step.
\tcbsubtitle{Model Response}
{\small
The image shows two geometric figures, a rectangular prism and a cube, with their dimensions labeled. The problem asks to calculate the surface area and volume of these figures. \\

1.For the rectangular prism:

Length (l) = 4 cm

Width (w) = 2 cm

Height (h) = 3 cm 

Surface Area (SA) of a rectangular prism = 2lw + 2lh + 2wh 

Volume (V) of a rectangular prism = lwh 

Substituting the given values:   

SA = 2(4cm)(2cm) + 2(4cm)(3cm) + 2(2cm)(3cm) 

SA = 16cm² + 24cm² + 12cm² SA = 52cm²

V = (4cm)(2cm)(3cm) V = 24cm³ \\

2.For the cube:

Side length (a) = 3 dm

Surface Area (SA) of a cube = 6a² Volume (V) of a cube = a³

Substituting the given value: 

SA = 6(3dm)² SA = 6(9dm²) SA = 54dm²

V = (3dm)³ V = 27dm³ \\

So, the surface area and volume for each figure are as follows:

For the rectangular prism:

Surface Area = 52 cm²

Volume = 24 cm³ \\

For the cube:

Surface Area = 54 dm²

Volume = 27 dm³}
\label{Visual_Referring_Prompting}
\end{tcolorbox}
\caption{Solving a math problem. The model calculates the surface area and volume of these figures with step-by-step explanation }
\end{figure}

\begin{tcolorbox}[colback=black!5!white,colframe=black!75!black,title=Algorithmic Problem Solving]
\includegraphics[width=15cm]{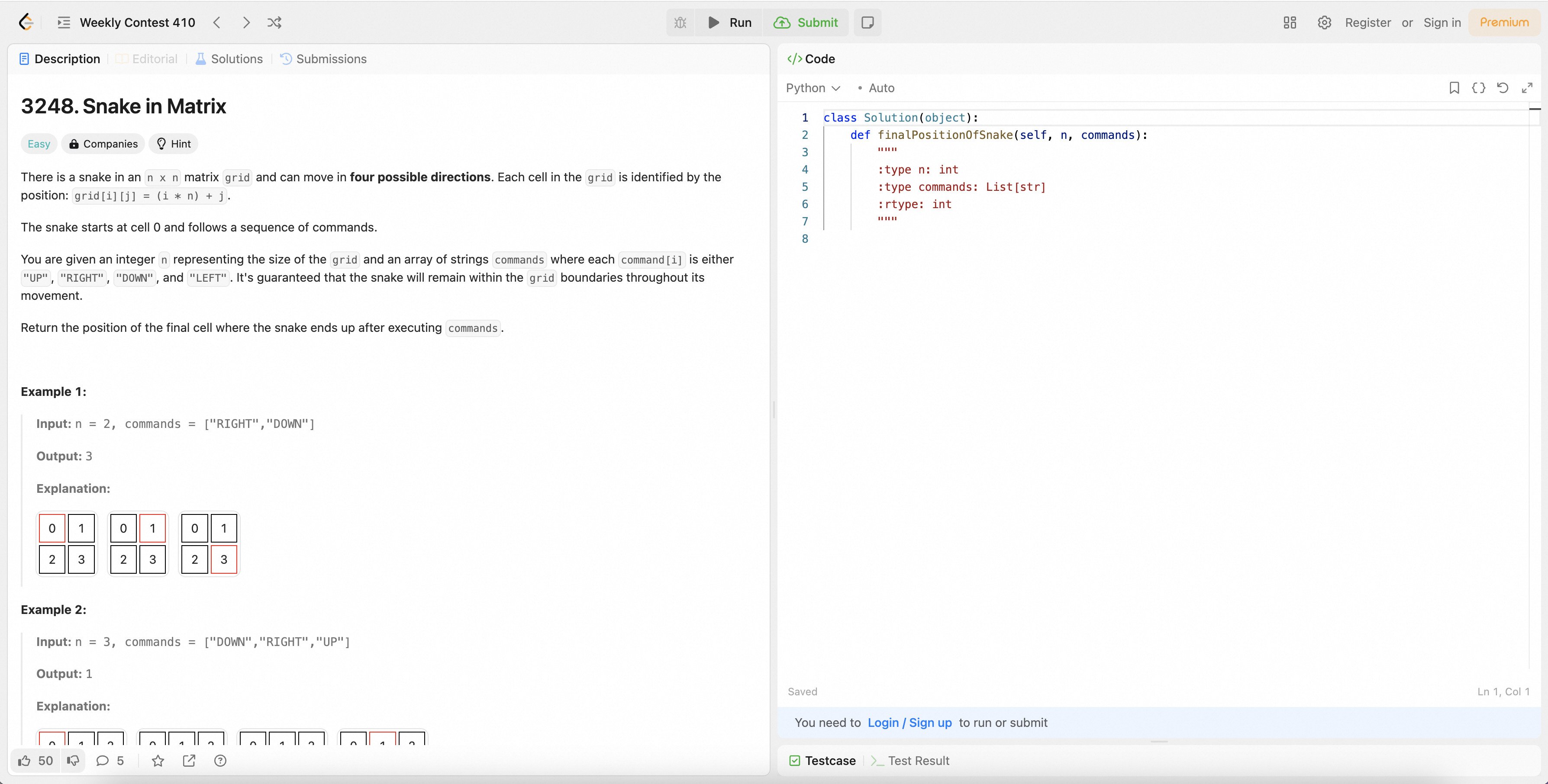}

Solve the algorithmic problem in the web page.
\end{tcolorbox}

\begin{figure}
\small
\centering
\vspace{-0.5cm}
\begin{tcolorbox}[colback=black!5!white,colframe=black!75!black,title=Model Response]
To solve the ``Snake in Matrix'' problem, we need to implement a function that takes the size of the matrix $n$ and a list of commands, and returns the final position of the snake after executing all the commands. Here's a step-by-step approach to solve this problem:
\begin{enumerate}
    \item \textbf{Initialize Variables}:
    \begin{itemize}
        \item Initialize the current position of the snake as $0$.
        \item Initialize the direction of the snake as \textbf{RIGHT} (initially, the snake moves to the right).
    \end{itemize}
    
    \item \textbf{Define Movement Functions}:
    \begin{itemize}
        \item Create a function to move the snake based on the current direction and the size of the matrix $n$.
        \item Update the position of the snake accordingly.
    \end{itemize}
    
    \item \textbf{Process Commands}:
    \begin{itemize}
        \item Iterate through each command in the list of commands.
        \item For each command, update the direction and position of the snake.
    \end{itemize}
    
    \item \textbf{Return Final Position}:
    \begin{itemize}
        \item After processing all commands, return the final position of the snake.
    \end{itemize}
\end{enumerate}
Here's the Python code to implement the solution:
\begin{lstlisting}[language=Python]
class Solution:
    def finalPositionOfSnake(self, n: int, commands: List[str]) -> int:
        # Initialize variables
        position = 0
        direction = "RIGHT"
        # Define movement functions
        def move_right():
            nonlocal position
            position += 1
        def move_down():
            nonlocal position
            position += n
        def move_left():
            nonlocal position
            position -= 1 
        def move_up():
            nonlocal position
            position -= n         
        # Process commands
        for command in commands:
            if command == "RIGHT":
                move_right()
            elif command == "DOWN":
                move_down()
            elif command == "LEFT":
                move_left()
            elif command == "UP":
                move_up()            
        # Return final position
        return position
\end{lstlisting}
This code defines a class \texttt{Solution} with a method \texttt{finalPositionOfSnake} that takes the size of the matrix $n$ and a list of commands, and returns the final position of the snake after executing all the commands. The movement functions (\texttt{move\_right}, \texttt{move\_down}, \texttt{move\_left}, \texttt{move\_up}) are used to update the position of the snake based on the current direction.
\label{}
\end{tcolorbox}
\caption{The model displays an adeptness in solving algorithmic problems.}
\end{figure}

\begin{figure}
\centering
\begin{tcolorbox}[colback=black!5!white,colframe=black!75!black,title=Web Content Recognition]
\includegraphics[width=10cm]{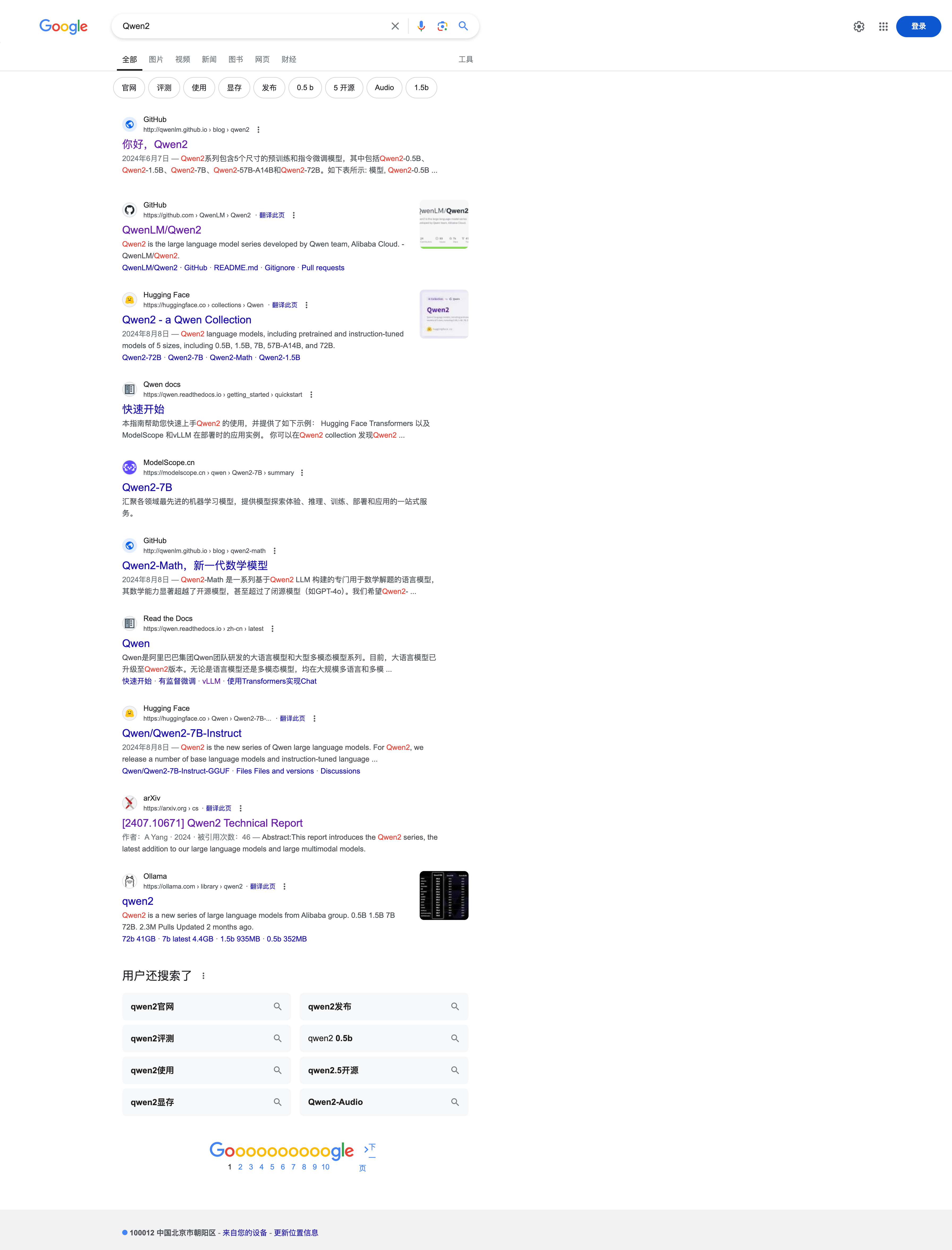}

Extract all page titles in the screenshot of search result.
\tcbsubtitle{Model Response}
\small
The page titles in the screenshot of the search result are:
\begin{CJK*}{UTF8}{gbsn}
\begin{enumerate}
    \item 你好，Qwen2
    \item QwenLM/Qwen2
    \item Qwen2 - a Qwen Collection
    \item 快速开始
    \item Qwen2-7B
    \item Qwen2-Math，新一代数学模型
    \item Qwen
    \item Qwen/Qwen2-7B-Instruct
    \item \texttt{[2407.10671] Qwen2 Technical Report}
    \item qwen2
\end{enumerate}
\end{CJK*}
\label{}
\end{tcolorbox}
\caption{The model displays an adeptness in recognize content from the web page.}
\end{figure}

\begin{figure}
\small
\centering
\begin{tcolorbox}[colback=black!5!white,colframe=black!75!black,title=OCR and Mathematical Problem Solving]
\includegraphics[width=7cm]{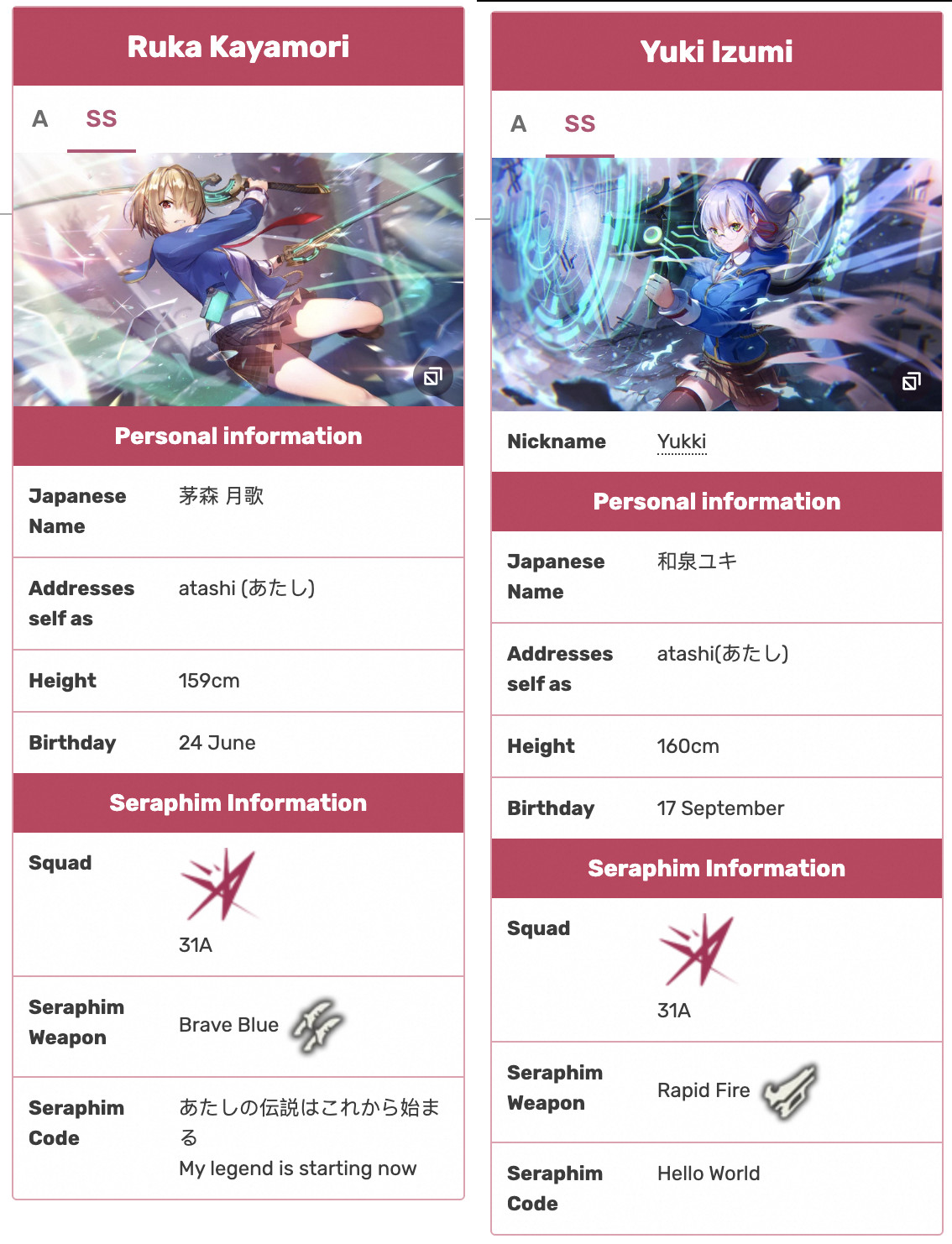}

Please tell me how many days are between Ruka's birthday and Yuki's birthday? Think carefully step by step.
\tcbsubtitle{Model Response}
To determine the number of days between Ruka's and Yuki's birthdays, we need to calculate the difference between their respective birth dates.

\begin{enumerate}
    \item Ruka's birthday is on June 24.
    \item Yuki's birthday is on September 17.
\end{enumerate}

Let's break it down step by step:

\begin{enumerate}
    \item \textbf{From June 24 to June 30:}
    \begin{itemize}
        \item June 24 to June 30 is 6 days.
    \end{itemize}
    
    \item \textbf{From July 1 to August 31:}
    \begin{itemize}
        \item July has 31 days.
        \item August has 31 days.
        \item Total days in July and August: 31 + 31 = 62 days.
    \end{itemize}
    
    \item \textbf{From September 1 to September 17:}
    \begin{itemize}
        \item September 1 to September 17 is 17 days.
    \end{itemize}
\end{enumerate}

Now, we add all these days together:

\begin{itemize}
    \item June 24 to June 30: 6 days
    \item July 1 to August 31: 62 days
    \item September 1 to September 17: 17 days
\end{itemize}

Total days = 6 + 62 + 17 = 85 days

Therefore, there are 85 days between Ruka's and Yuki's birthdays.
\label{}
\end{tcolorbox}
\caption{The model displays an adeptness in OCR and mathematical reasoning.}
\end{figure}

\begin{figure}
\centering
\small
\begin{tcolorbox}[colback=black!5!white,colframe=black!75!black,title=OCR and Format Following]
\includegraphics[width=6cm]{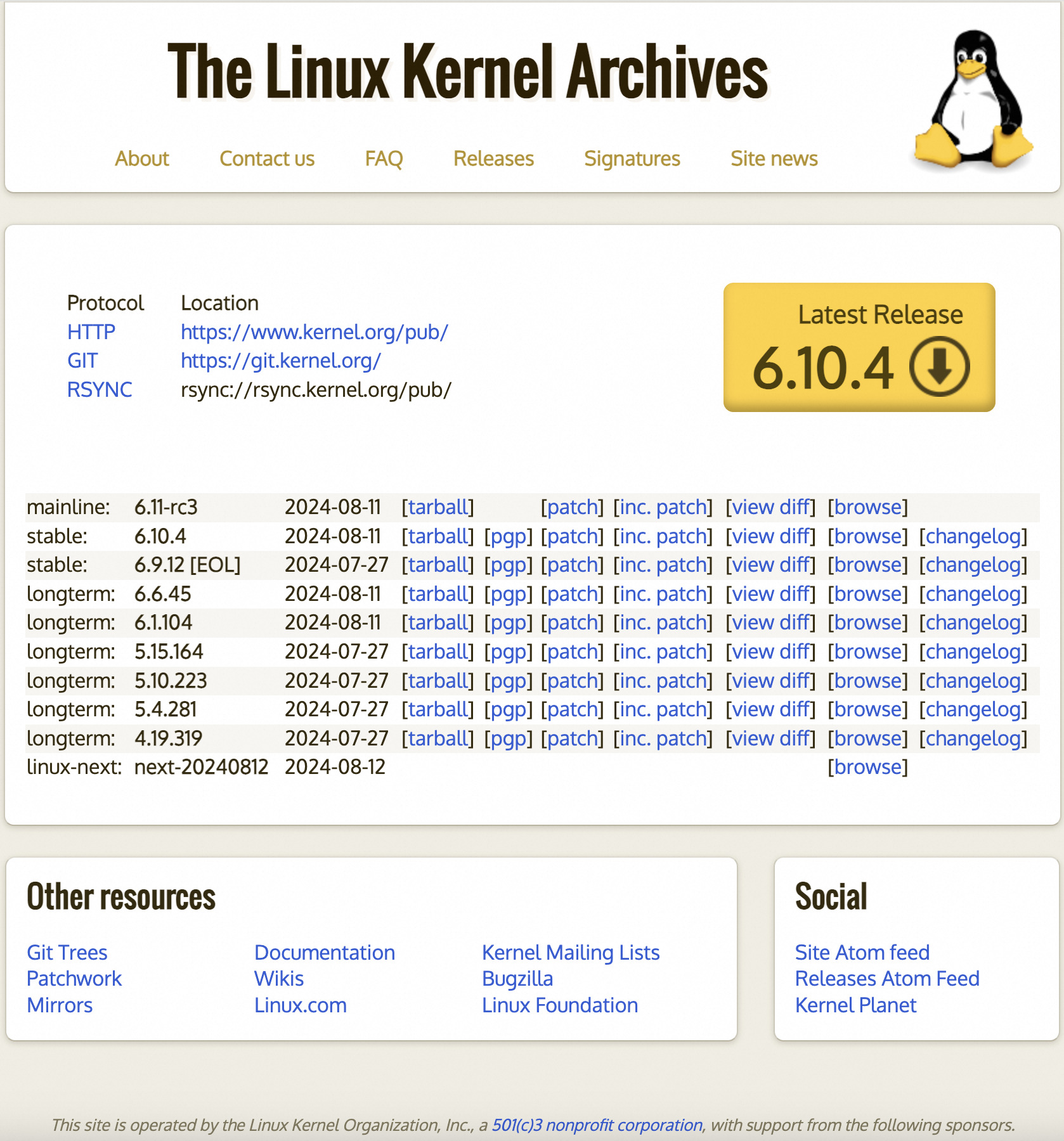}

What are the Linux versions and their release dates in the picture? Return results as a JSON list.
\tcbsubtitle{Model Response}
\begin{lstlisting} 
[
  {
    "version": "6.11-rc3", "release_date": "2024-08-11"
  },
  {
    "version": "6.10.4", "release_date": "2024-08-11"
  },
  {
    "version": "6.9.12 [EOL]", "release_date": "2024-07-27"
  },
  {
    "version": "6.6.45", "release_date": "2024-08-11"
  },
  {
    "version": "6.1.104", "release_date": "2024-08-11"
  },
  {
    "version": "5.15.164", "release_date": "2024-07-27"
  },
  {
    "version": "5.10.223", "release_date": "2024-07-27"
  },
  {
    "version": "5.4.281", "release_date": "2024-07-27"
  },
  {
    "version": "4.19.319", "release_date": "2024-07-27"
  },
  {
    "version": "next-20240812", "release_date": "2024-08-12"
  }
]
\end{lstlisting}
\label{}
\end{tcolorbox}
\caption{The model displays an adeptness in OCR and following formats.}
\end{figure}

\begin{figure}
\centering
\begin{tcolorbox}[colback=black!5!white,colframe=black!75!black,title=OCR and Format Following]
\includegraphics[width=12cm]{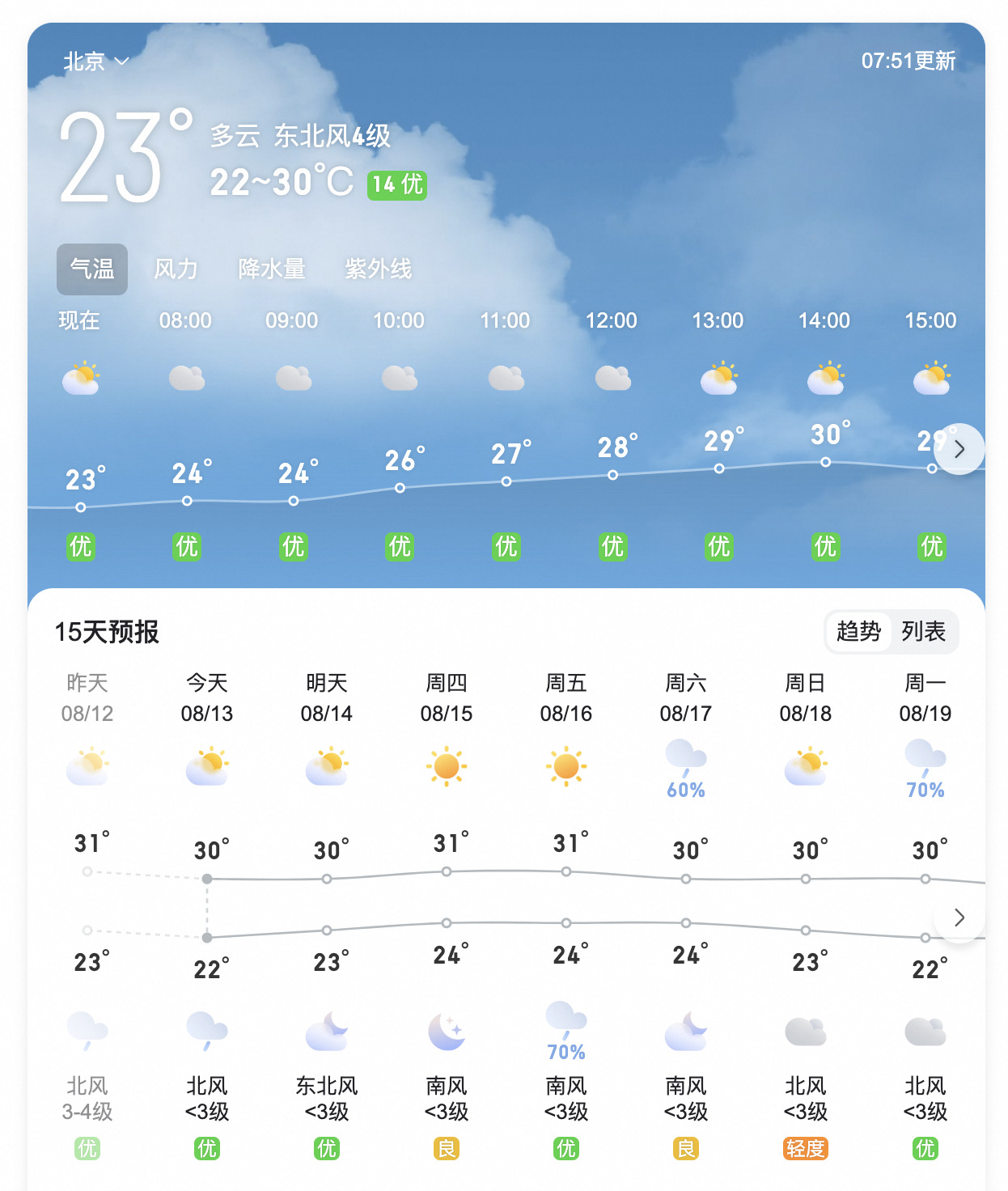}

\begin{CJK*}{UTF8}{gbsn}
将图中的每日的最高气温和最低气温整理成表格。
(Translation: Organize the daily highest and lowest temperatures from the image into a table.)
\end{CJK*}
\tcbsubtitle{Model Response}
\begin{CJK*}{UTF8}{gbsn}
以下是图中每日的最高气温和最低气温整理成的表格：

\begin{tabular}{lcc}
\toprule
日期 & 最高气温 & 最低气温 \\
\midrule
08/12 & 31°C & 23°C \\
08/13 & 30°C & 22°C \\
08/14 & 30°C & 23°C \\
08/15 & 31°C & 24°C \\
08/16 & 31°C & 24°C \\
08/17 & 30°C & 24°C \\
08/18 & 30°C & 23°C \\
08/19 & 30°C & 22°C \\
\bottomrule
\end{tabular}

希望这对你有帮助！
\end{CJK*}
\label{}
\end{tcolorbox}
\caption{The model displays an adeptness in OCR and following formats.}
\end{figure}

\begin{figure}
\centering
\vspace{-0.5cm}
\begin{tcolorbox}[colback=black!5!white,colframe=black!75!black,title=OCR for Large Images]
\includegraphics[width=2.1cm]{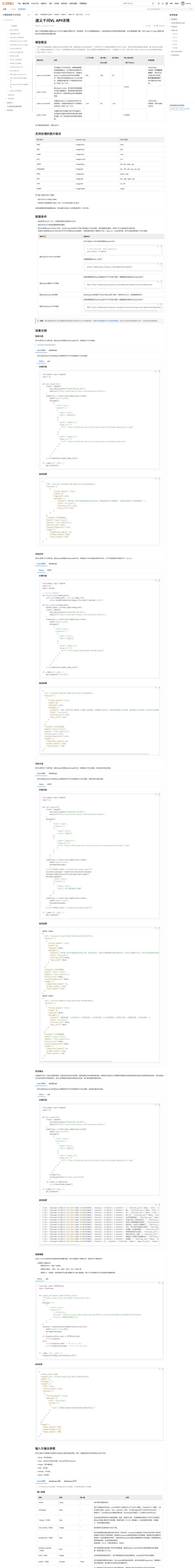}
\includegraphics[width=12.5cm]{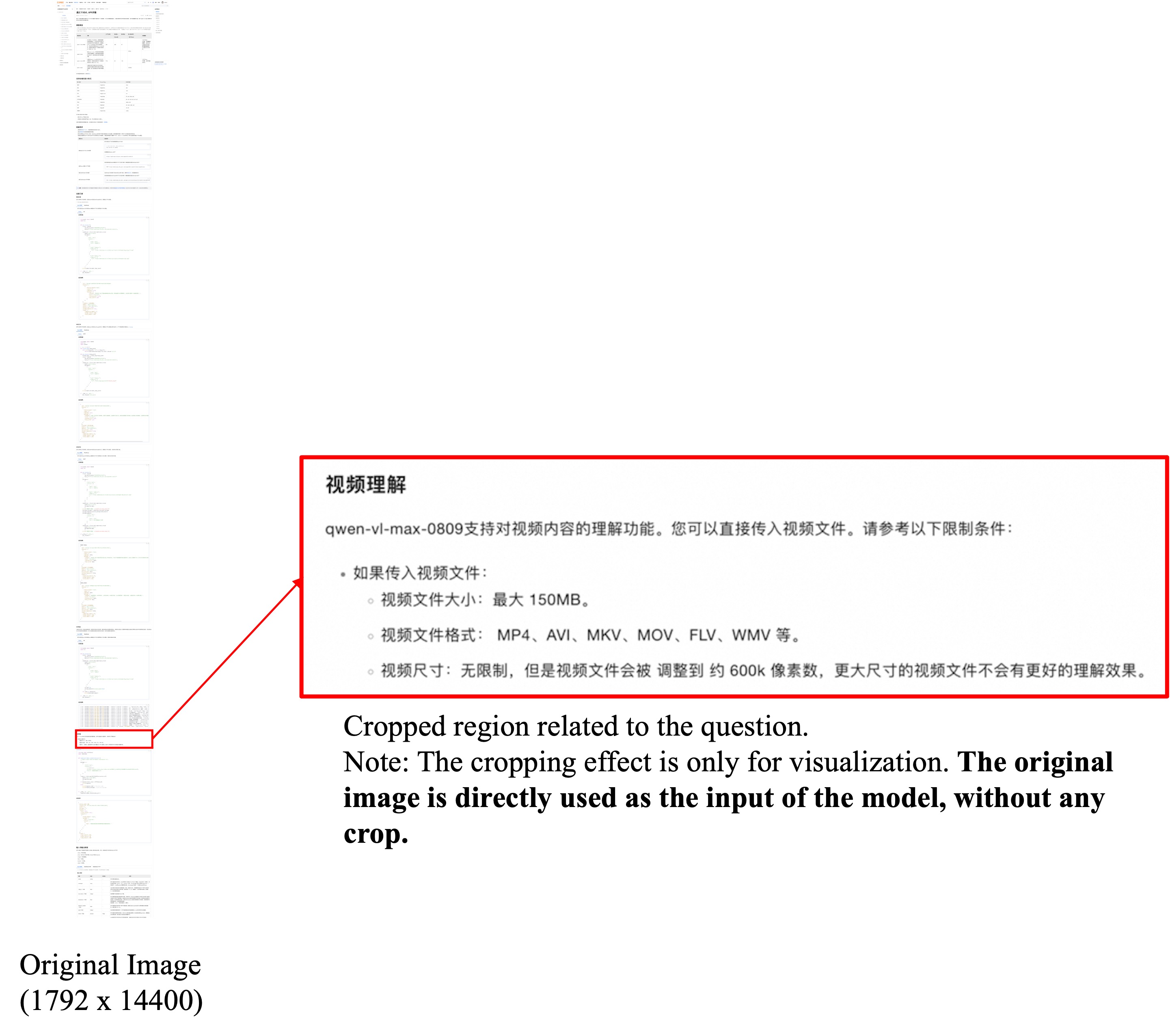}
%(The image resolution is 1792 x 14400)

\begin{CJK*}{UTF8}{gbsn}
qwen-vl-max-0809对视频内容的理解功能存在哪些限制？

(Translation: What limitations exist in qwen-vl-max-0809's ability to understand video content?)
\end{CJK*}
\tcbsubtitle{Model Response}
\begin{CJK*}{UTF8}{gbsn}
qwen-vl-max-0809对视频内容的理解功能存在以下限制：视频文件大小最大为150MB，视频文件格式支持MP4、AVI、MKV、MOV、FLV、WMV等，视频尺寸无限制，但视频文件会被调整到约600K像素数，更大尺寸的视频文件不会有更好的理解效果。
(Translation: qwen-vl-max-0809 has the following limitations on the video content understanding: the maximum video file size is 150MB, the video file formats supported are MP4, AVI, MKV, MOV, FLV, WMV, etc. There is no limitation on the video size, but the video file will be resized to about 600K pixels, and video files of larger sizes won't have better video understanding results.)
\end{CJK*}
\label{}
\end{tcolorbox}
\caption{The model displays an adeptness in large images OCR.}
\end{figure}

\clearpage
\subsection{Video Understanding}

\begin{figure}[htbp]
\centering
\begin{tcolorbox}[colback=black!5!white,colframe=black!75!black,title=Multi-Round Video Chat]
\includegraphics[width=15cm]{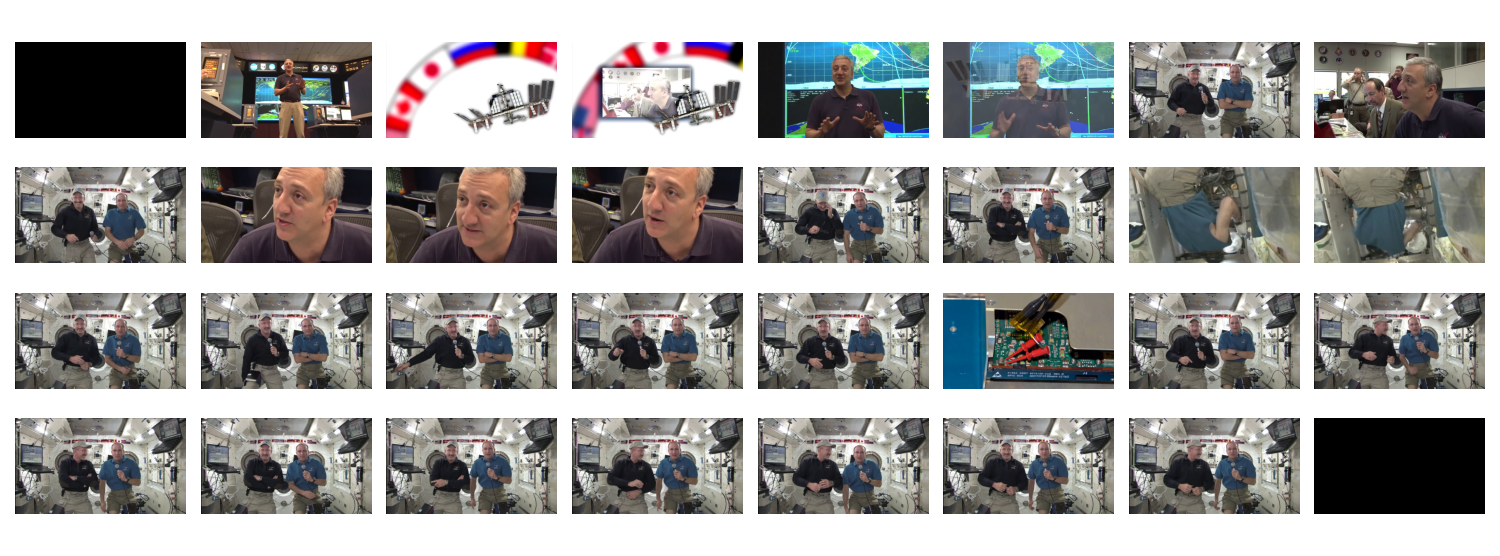}

\begin{CJK*}{UTF8}{gbsn}
Please describe the video in detail.
\end{CJK*}
\tcbsubtitle{Model Response}

The video begins with a man speaking to the camera, followed by a group of people sitting in a control room. The camera then cuts to two men floating inside a space station, where they are seen speaking to the camera. The men appear to be astronauts, and they are wearing space suits. The space station is filled with various equipment and machinery, and the camera pans around to show the different areas of the station. The men continue to speak to the camera, and they appear to be discussing their mission and the various tasks they are performing. Overall, the video provides a fascinating glimpse into the world of space exploration and the daily lives of astronauts.

\tcbsubtitle{User Query}
What is the color of the clothes the two astronauts wearing?

\tcbsubtitle{Model Response}
The two astronauts are wearing blue and black clothes.

\label{mt_format}
\end{tcolorbox}
\caption{The model displays an adeptness in recognizing multi-round video chat.}
\end{figure}

\begin{figure}[htbp]
\centering
\vspace{-2.7em}
\begin{tcolorbox}[colback=black!5!white,colframe=black!75!black,title=Multi-Video Understanding]
\includegraphics[width=8cm]{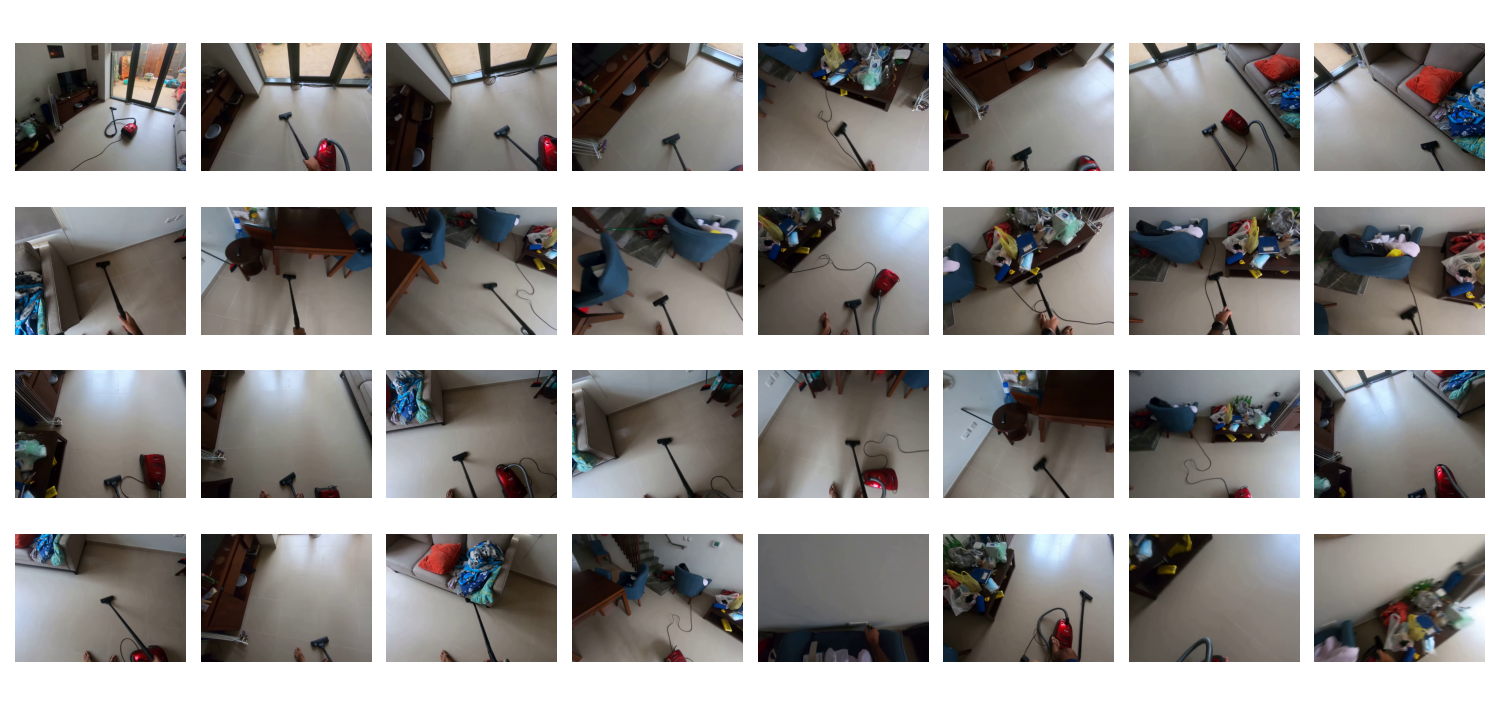}

\begin{CJK*}{UTF8}{gbsn}
\small 视频中的人在做什么？

\small (Translation: What are the people in the video doing?)
\end{CJK*}
\tcbsubtitle{Model Response}

\begin{CJK*}{UTF8}{gbsn}
\small 视频中的人正在使用吸尘器清理地板。

\small (Translation: The man in the video is using a vacuum cleaner to clean the floor.)
\end{CJK*}

\tcbsubtitle{User Query}
\begin{CJK*}{UTF8}{gbsn}
\small 如果要进一步清洁地面的话，接下来该怎么做？

\small (Translation: What should I do next if I want to clean the floor further?)
\end{CJK*}

\tcbsubtitle{Model Response}
\begin{CJK*}{UTF8}{gbsn}
\small 如果要进一步清洁地面，可以使用拖把或湿布擦拭地面，以去除吸尘器无法清除的污渍和灰尘。

\small (Translation: For further cleaning of the floor, wipe the floor with a mop or damp cloth to remove stains and dirt that the vacuum cleaner cannot remove.)
\end{CJK*}

\tcbsubtitle{User Query}
\includegraphics[width=8cm]{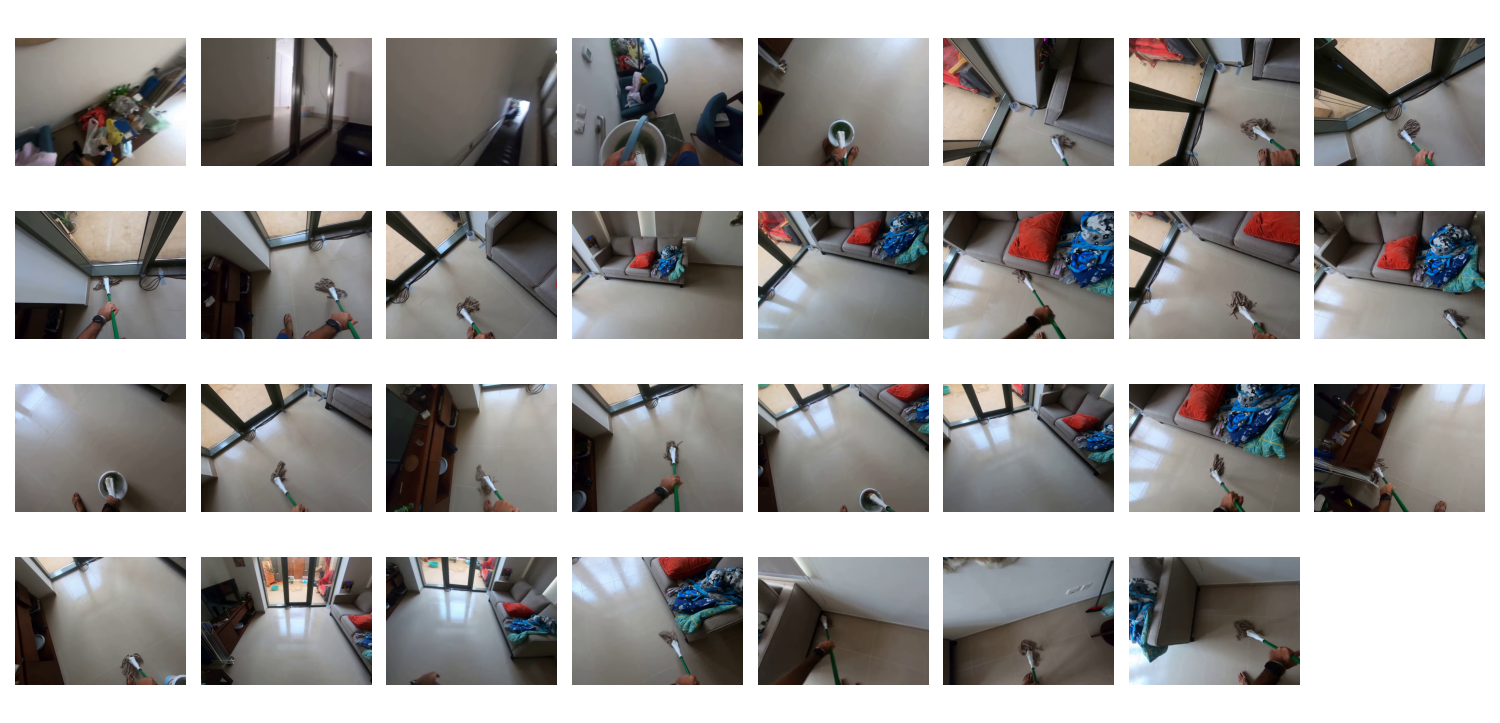}

\begin{CJK*}{UTF8}{gbsn}
\small 已经用拖把清洁了地面，评价一下清洁工作？

\small (Translation: Already cleaned the floor with a mop and please comment on the cleaning?)
\end{CJK*}

\tcbsubtitle{Model Response}
\begin{CJK*}{UTF8}{gbsn}
\small 从视频中可以看出，清洁工作已经完成，地面已经变得干净整洁。

\small (Translation: As you can see from the video, the cleaning has been completed and the floor has been made clean and tidy.)
\end{CJK*}

\tcbsubtitle{User Query}
\begin{CJK*}{UTF8}{gbsn}
\small 通过你看到的，接下来还能做些什么？

\small (Translation: By what you see, what else can you do next?)
\end{CJK*}

\tcbsubtitle{Model Response}
\begin{CJK*}{UTF8}{gbsn}
\small 接下来可以将拖把清洗干净，晾干备用。

\small (Translation: Next you can clean the mop, dry it and set it aside.)
\end{CJK*}

\label{mt_format}
\end{tcolorbox}
\caption{The model displays an adeptness in recognizing multi-video understanding.}
\end{figure}

\subsection{Visual Agent Capability}
The Qwen2-VL also excels in location and agent tasks.

\begin{figure}[htbp]
\centering
\begin{tcolorbox}[colback=black!5!white,colframe=black!75!black,title=Locate Whatever You Desire]
\includegraphics[width=9cm]{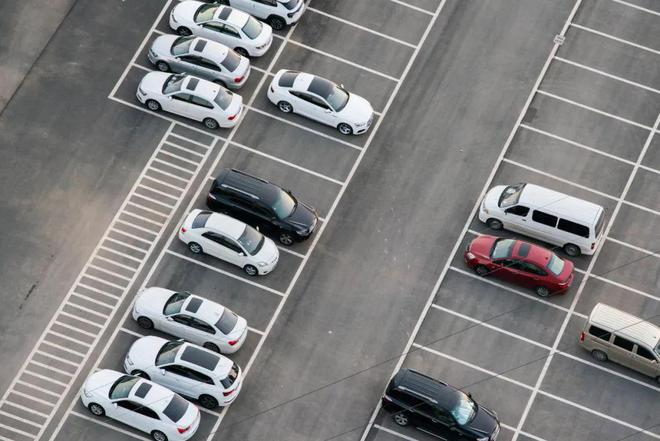}

Detect the bounding box of 'the red car'
\tcbsubtitle{Model Response}

<|object\_ref\_start|>the red car<|object\_ref\_end|><|box\_start|>(701,531),(869,675)<|box\_end|>

\includegraphics[width=9cm]{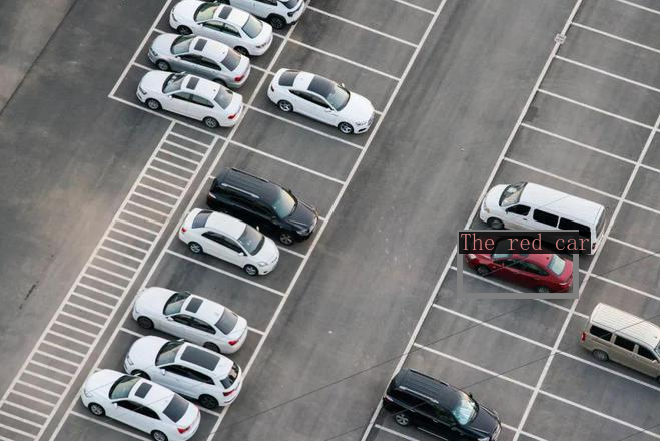}

\label{Locate}
\end{tcolorbox}
\caption{Our models were able to locate specific elements within images, such as identifying the red car accurately. }
\end{figure}

\begin{figure}[htbp]
\centering
\begin{tcolorbox}[colback=black!5!white,colframe=black!75!black,title=Visual Referring Prompting]
\includegraphics[width=12cm]{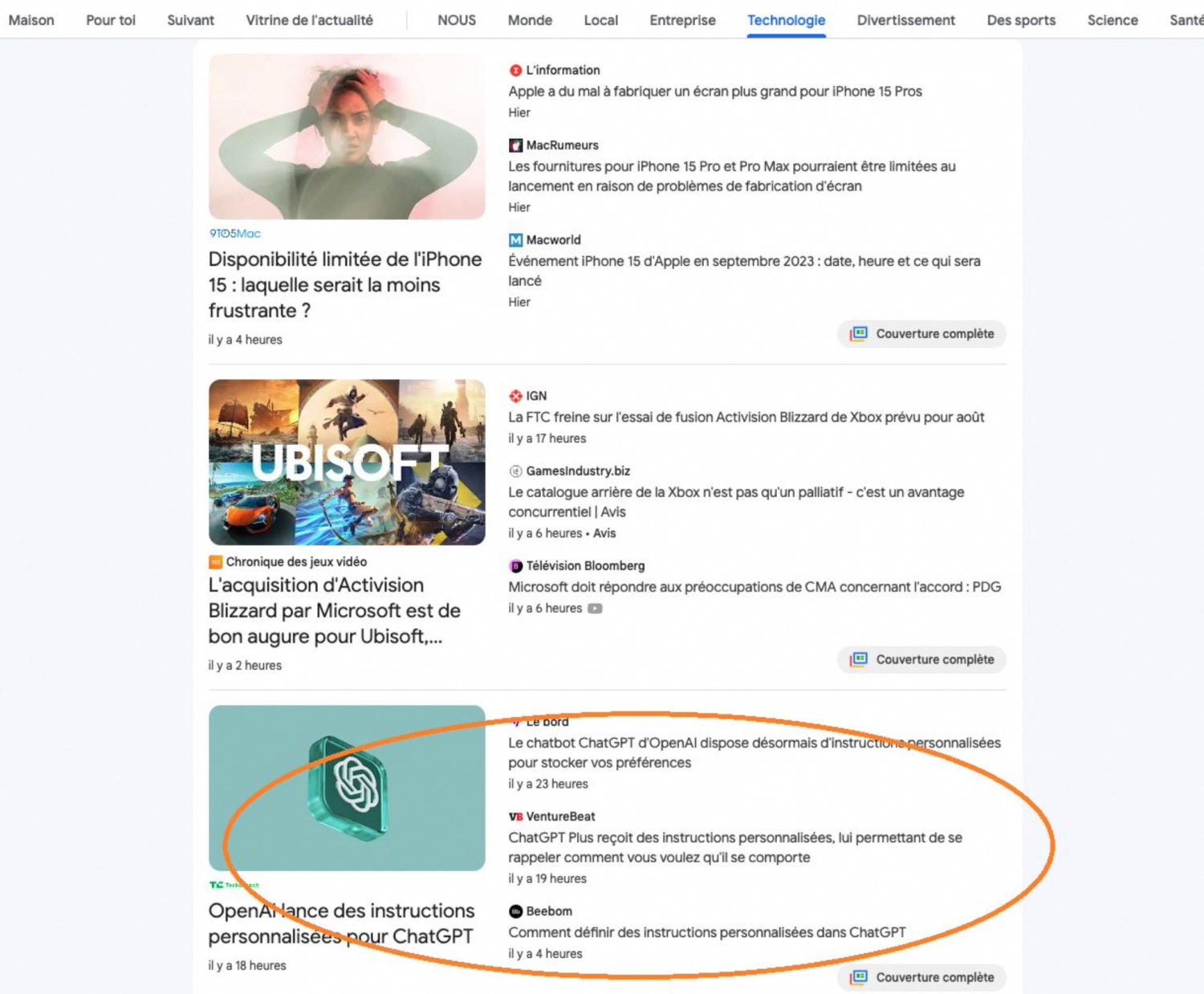}

What does the pointed news talk about?
\tcbsubtitle{Model Response}

The pointed news with a green logo of OpenAI and a title "Le chatbot ChatGPT d'OpenAI dispose désormais d'instructions personnalisées pour stocker vos préférences" is talking about the new feature of ChatGPT that can now store personalized instructions to better serve users.
\label{Visual_Referring_Prompting}
\end{tcolorbox}
\caption{Our model is capable of annotating and posing questions regarding the content present in webpage screenshots, demonstrating its potential as a visual agent. Image source: ~\citep{yang2023dawn}}
\end{figure}

\begin{figure}[htbp]
\centering
\begin{tcolorbox}[colback=black!5!white,colframe=black!75!black,title=Function Calling - Basic]
\includegraphics[width=8cm]{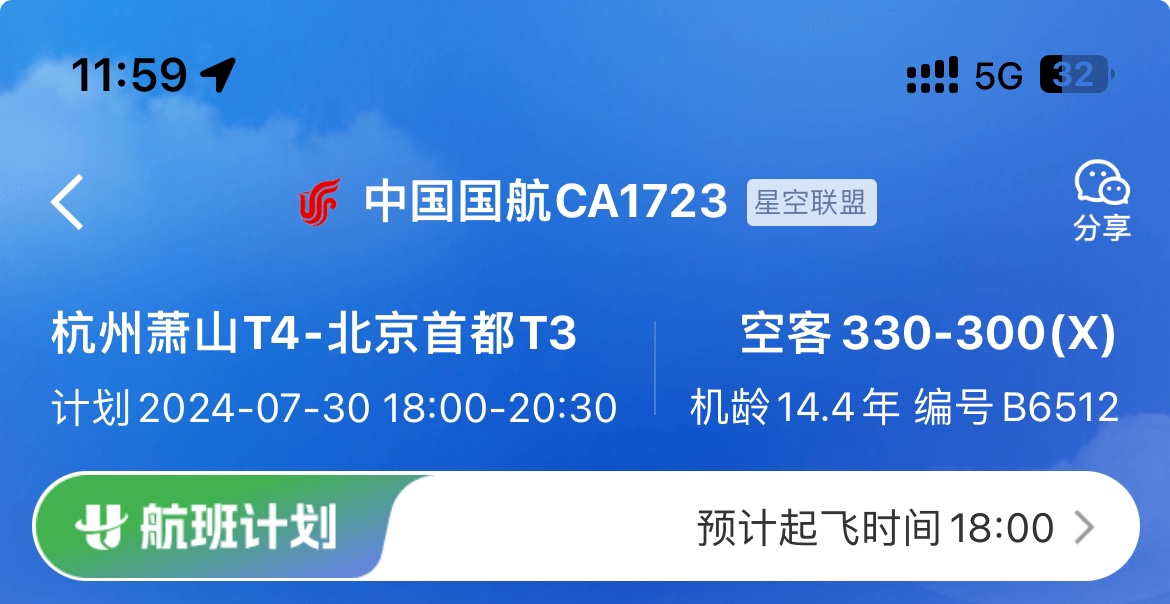}

\begin{CJK*}{UTF8}{gbsn}
几点到达，到的时候目的地天气怎么样

(Translation: What time does this flight arrive? What is the weather like at the destination when it arrives?)
\end{CJK*}
\tcbsubtitle{Function Input}
\begin{CJK*}{UTF8}{gbsn}
北京
\end{CJK*}

\tcbsubtitle{Function Result}
\begin{CJK*}{UTF8}{gbsn}
时间: 2024-07-30 14:00, 风向: 西南风, 风力: 0-3级 微风 <5.4m/s, 温度: 29°C, 天气: 小雨

时间: 2024-07-30 15:00, 风向: 西南风, 风力: 0-3级 微风 <5.4m/s, 温度: 28°C, 天气: 小雨

时间: 2024-07-30 16:00, 风向: 西南风, 风力: 0-3级 微风 <5.4m/s, 温度: 28°C, 天气: 小雨

时间: 2024-07-30 17:00, 风向: 西南风, 风力: 0-3级 微风 <5.4m/s, 温度: 28°C, 天气: 小雨

时间: 2024-07-30 18:00, 风向: 西南风, 风力: 0-3级 微风 <5.4m/s, 温度: 27°C, 天气: 中雨

时间: 2024-07-30 19:00, 风向: 南风, 风力: 0-3级 微风 <5.4m/s, 温度: 27°C, 天气: 中雨

时间: 2024-07-30 20:00, 风向: 南风, 风力: 0-3级 微风 <5.4m/s, 温度: 27°C, 天气: 中雨

时间: 2024-07-30 21:00, 风向: 南风, 风力: 0-3级 微风 <5.4m/s, 温度: 26°C, 天气: 小雨

时间: 2024-07-30 22:00, 风向: 南风, 风力: 0-3级 微风 <5.4m/s, 温度: 26°C, 天气: 小雨

时间: 2024-07-30 23:00, 风向: 南风, 风力: 0-3级 微风 <5.4m/s, 温度: 26°C, 天气: 小雨

（避免显示过长，省略部分信息。）
\end{CJK*}

\tcbsubtitle{Model Response}
\begin{CJK*}{UTF8}{gbsn}
根据航班信息，您乘坐的中国国航CA1723航班预计将于2024年7月30日20:30抵达北京首都机场T3航站楼。

在您到达时，北京的天气预报显示会有中雨，并且温度约为27℃。请注意携带雨具并做好防雨措施。祝您旅途愉快！

(Translation: According to flight information, your Air China flight CA1723 is expected to arrive at Beijing Capital Airport T3 Terminal at 20:30 on July 30, 2024. When you arrive, the weather forecast in Beijing shows moderate rain and the temperature is about 27℃. Please remember to bring rain gear and take rain protection measures. Have a pleasant journey!)
\end{CJK*}
\label{func_call_0}
\end{tcolorbox}
\caption{The model identified the destination and arrival time in the image, called the 24-hour weather query function, correctly input the destination, and extracted the weather at the arrival time from the query results, thus enabling it to answer the user's question.}
\end{figure}

\begin{figure}[htbp]
\centering
\begin{tcolorbox}[colback=black!5!white,colframe=black!75!black,title=Function Calling - Code Interpreter]
\includegraphics[width=14cm]{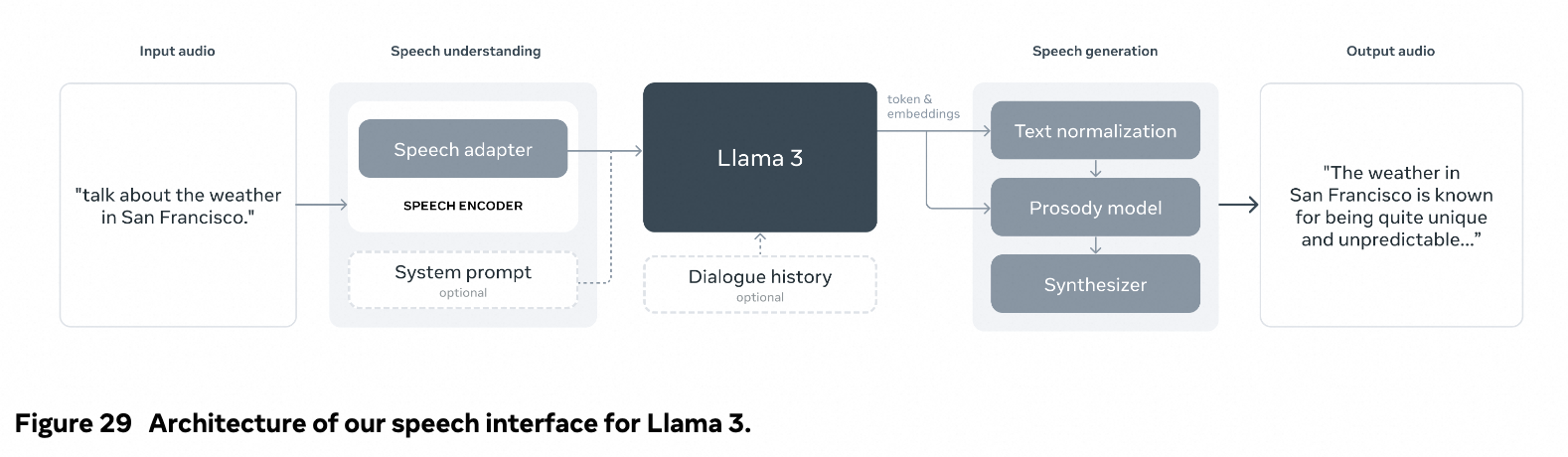}

Implement the flowchart, representing modules as functions without implementation details. Write the code and validate it using the code interpreter.
\tcbsubtitle{Function Input}
\begin{lstlisting}[language=Python]
# Placeholder functions for the modules
def speech_adapter(input_audio):
    # Placeholder for speech adapter implementation
    pass

# To avoid excessive length, 
# the remaining functions without implementation details are omitted here.

# Example usage
input_audio = "talk about the weather in San Francisco."
speech_adapter_output = speech_adapter(input_audio)
speech_encoder_output = speech_encoder(speech_adapter_output)
system_prompt_output = system_prompt()
dialogue_history_output = dialogue_history()
llama_3_output = llama_3(speech_encoder_output, system_prompt_output, \
dialogue_history_output)
text_normalization_output = text_normalization(llama_3_output)
prosody_model_output = prosody_model(text_normalization_output)
output_audio = synthesizer(prosody_model_output)
\end{lstlisting}

\tcbsubtitle{Function Result}
Finished execution.

\tcbsubtitle{Model Response}
The code has been executed without any errors, which means that the structure of the code matches the flowchart and there are no syntax issues. However, since the functions are placeholders and do not perform any actual operations, we cannot validate the correctness of the implementation. The code simply represents the flowchart by defining each module as a function and demonstrating how the functions could be called in sequence to process input audio and produce output audio.
\label{func_call_1}
\end{tcolorbox}
\caption{The model identified the steps in the flowchart, drafted the simulation code as required while omitting the details, and then successfully executed it through the code interpreter. Image source:~\citep{dubey2024llama}}
\end{figure}

\begin{figure}[htbp]
\centering
\begin{tcolorbox}[colback=black!5!white,colframe=black!75!black,title=Function Calling - Code Interpreter]
\includegraphics[width=14cm]{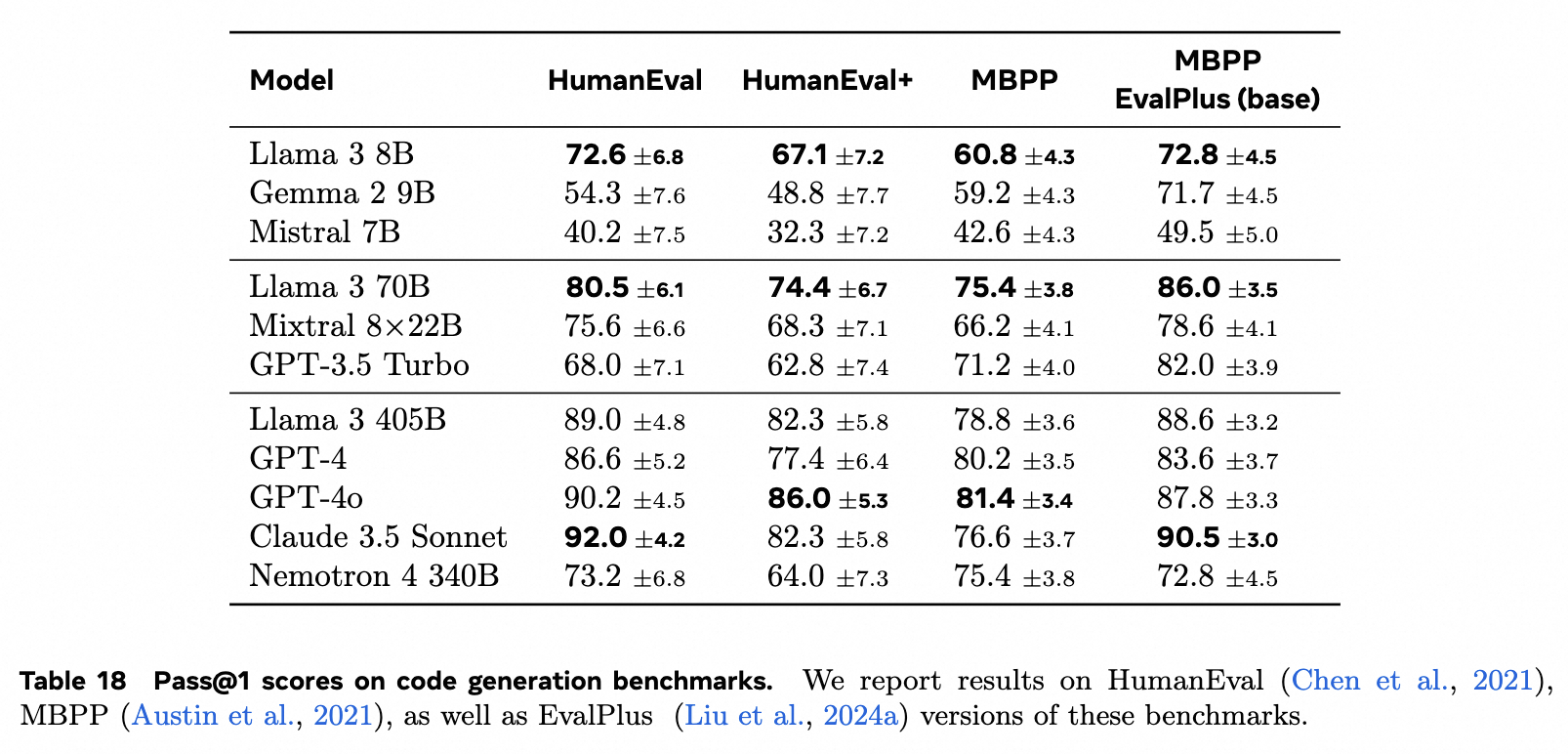}

Count the average value of each row in this chart and draw a bar chart to show it. Write the code and verify it using the code interpreter.
\tcbsubtitle{Function Input}
\begin{lstlisting}[language=Python]
import pandas as pd
import numpy as np
import matplotlib.pyplot as plt

# Extracting the numerical values from the table
data = {
    "Model": ["Llama 3 8B", "Gemma 2 9B", "Mistral 7B", "Llama 3 70B", \
        "Mixtral 8x22B", "GPT-3.5 Turbo", "Llama 3 405B", "GPT-4", \
        "GPT-4o", "Claude 3.5 Sonnet", "Nemotron 4 340B"],
    "HumanEval": [72.6, 54.3, 40.2, 80.5, 75.6, 68.0, 89.0, 86.6, 90.2, \
        92.0, 73.2],
    "HumanEval+": [67.1, 48.8, 32.3, 74.4, 68.3, 62.8, 82.3, 77.4, 86.0, \
        82.3, 64.0],
    "MBPP": [60.8, 59.2, 42.6, 75.4, 66.2, 71.2, 78.8, 80.2, 81.4, 76.6, \
        75.4],
    "MBPP EvalPlus (base)": [72.8, 71.7, 49.5, 86.0, 78.6, 82.0, 88.6, \
        83.6, 87.8, 90.5, 72.8]
}

# Create a DataFrame
df = pd.DataFrame(data)

# Calculate the average for each row
df["Average"] = df.mean(axis=1)

# Plotting the bar chart
plt.figure(figsize=(12, 6))
plt.bar(df["Model"], df["Average"], color="skyblue")
plt.xlabel("Model")
plt.ylabel("Average Score")
plt.title("Average Pass@1 Scores on Code Generation Benchmarks")
plt.xticks(rotation=45, ha="right")
plt.tight_layout()
plt.show()
\end{lstlisting}

\label{func_call_2}
\end{tcolorbox}
\end{figure}

\begin{figure}[htbp]
\centering
\begin{tcolorbox}[colback=black!5!white,colframe=black!75!black,title=Function Result]

\includegraphics[width=14cm]{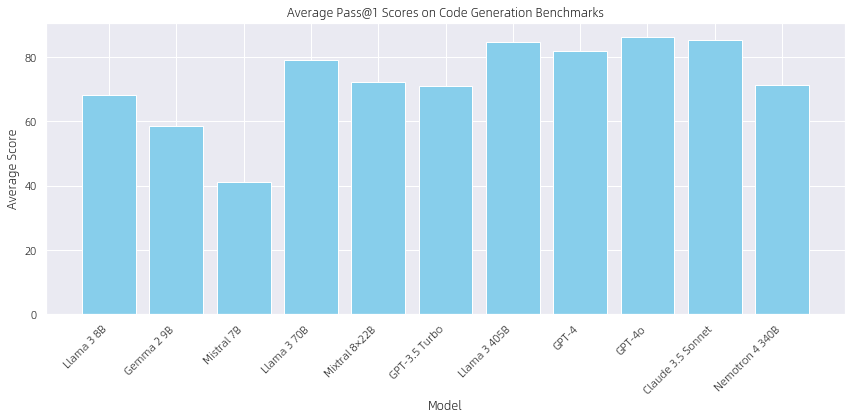}

\tcbsubtitle{Model Response}
The average Pass@1 scores on code generation benchmarks for each model have been calculated and are displayed in the table above. The bar chart shows the average scores for each model, providing a visual comparison of their performance across the different benchmarks.
\label{func_call_2_1}
\end{tcolorbox}
\caption{The model analyzed the table, and wrote the code to calculate the mean of each row in the table and to plot the results as a bar chart, which was successfully executed and visualized by the code interpreter. Image source:~\citep{dubey2024llama}}
\end{figure}

% function call example: formula

\begin{tcolorbox}[colback=black!5!white,colframe=black!75!black,title=Function Calling - Code Interpreter]
\includegraphics[width=10cm]{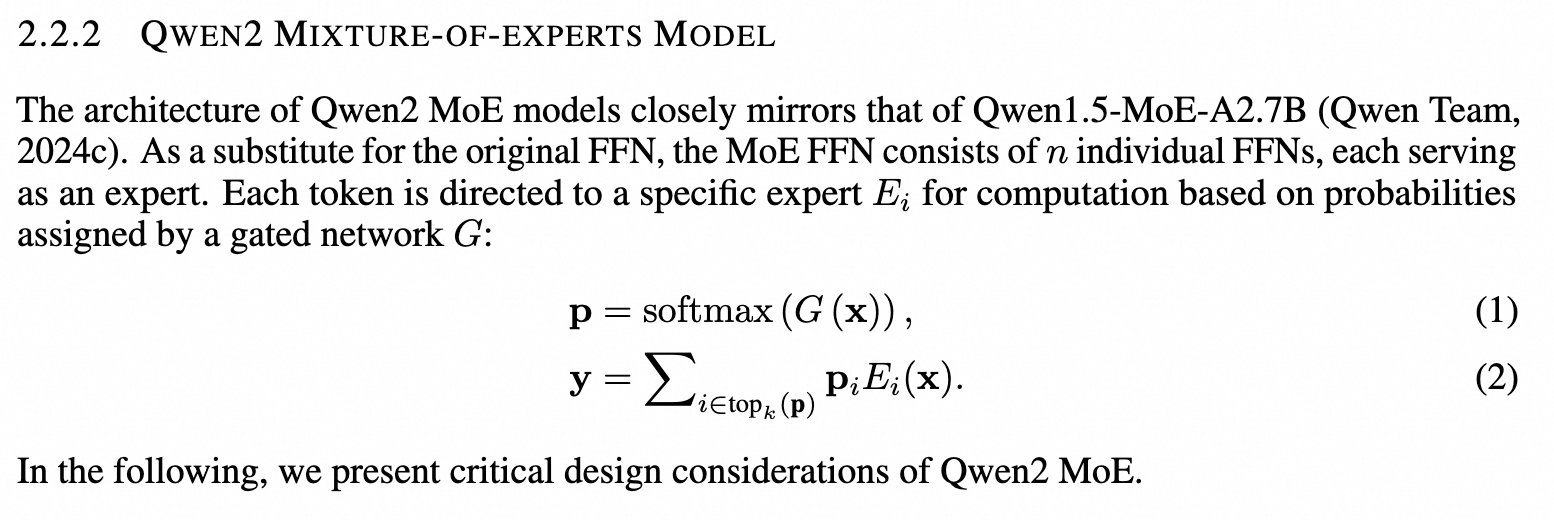}

Implement the formula, write the code and verify it using the code interpreter.
\tcbsubtitle{Function Input}
\begin{lstlisting}[language=Python]
import numpy as np

def softmax(x):
    """Compute softmax values for each sets of scores in x."""
    e_x = np.exp(x - np.max(x))
    return e_x / e_x.sum(axis=0)

def top_k(p, k):
    """Return the indices of the top k elements in the probability vector p."""
    return np.argsort(p)[-k:][::-1]

# Example input
x = np.array([1, 2, 3, 4, 5])  # Example input vector
G = lambda x: x  # Simple gating network (identity function for demonstration)
E = lambda x, i: x + i  # Simple expert functions (add the expert index to the input)
# Compute probabilities
p = softmax(G(x))
# Select top k experts
k = 3
top_experts = top_k(p, k)
# Compute output
y = np.sum([p[i] * E(x, i) for i in top_experts])
p, top_experts, y
\end{lstlisting}

\tcbsubtitle{Function Result}
(array([0.01165623, 0.03168492, 0.08612854, 0.23412166, 0.63640865]),
 array([4, 3, 2]),
 31.451165958047895)

\tcbsubtitle{Model Response}
The probabilities assigned to each expert based on the input vector (\textbf{x}) are: (\textbf{p} = [0.01165623, 0.03168492, 0.08612854, 0.23412166, 0.63640865]).

The top (k = 3) experts selected based on these probabilities are the experts with indices 4, 3, and 2 (in descending order of probability).
The output (\textbf{y}) computed as a weighted sum of the outputs of these top experts is approximately 31.4512.

This example demonstrates a simple implementation of the MoE model's operation. In a real-world scenario, the gating network ($G$) and the experts ($E_i$) would be more complex, involving neural network layers and potentially non-linear transformations.
\label{func_call_3}
\end{tcolorbox}
\captionof{figure}{The model understood the formula, implemented the code as required, and successfully executed it in the code interpreter to obtain the results. Image source: ~\citep{dubey2024llama}}

\begin{tcolorbox}[colback=black!5!white,colframe=black!75!black,title=VL Agent - UI Operations]
% \tcbsubtitle{}
% \begin{center}
\includegraphics[width=0.8\textwidth]{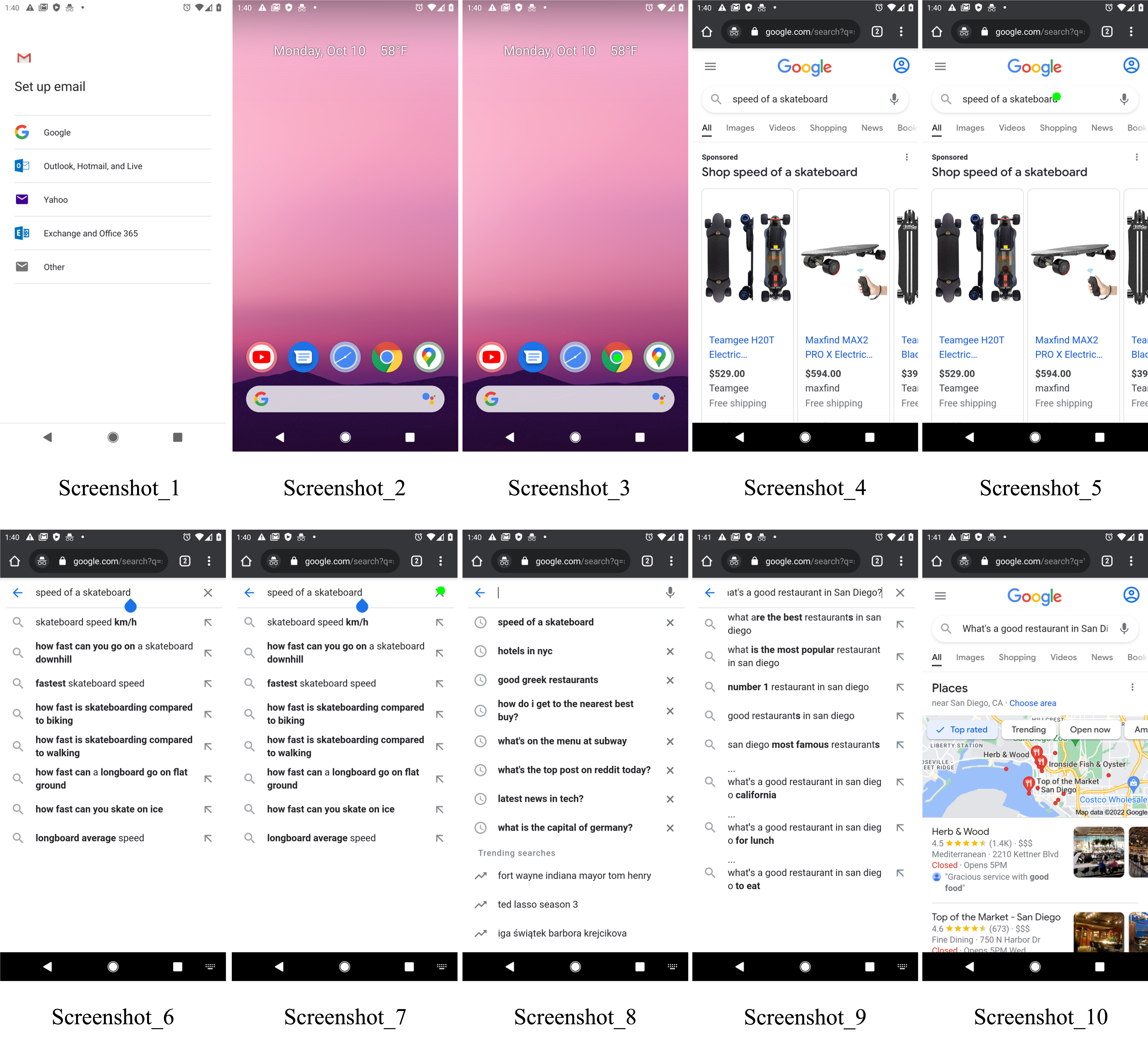}
%\end{center}
\tcbsubtitle{System Message}
You are a helpful assistant. \\
\# Actions \\
\#\# You have the following actions. \\
\#\#\# Click \\
Click: A quick, light fingertip press that commands, selects, or navigates through a phone's user interface. Parameters: [\{"name": "point", "description": "The specific point of interest on the screen, marked by the coordinate (x, y)." "required": true, "schema": "string"\}] \\
\#\#\# Type \\
Type: Engaging with a smartphone's interface by entering text for various purposes like messaging, searching, or command execution. Parameters: [\{ "name": "text", "description": "The text to be typed on a smartphone.", "required": true, "schema": "string"\}] \\
\#\#\# Swipe \\
Swipe: Touch the screen and move your finger to a new location for new content. Parameters: [{"name": "direction", "description": "Determine the swipe route from the cardinal directions: up, down, left, or right." "required": true, "schema": "string", "choices": ["up", "down", "left", "right"]}] \\
\#\#\# Back \\
Back: Go in reverse to the prior interface or exiting the current context Parameters: [] \\
\#\#\# Home \\
Home: Return to the device's home screen from any application or task. Parameters: [] \\
\#\#\# Enter \\
Enter: Execute a command to confirm inputs or selections. Parameters: [] \\
\#\#\# Done \\
Done: Indicate the completion of a task. Parameters: [] \\

\label{agent_ui_operation_part1}
\end{tcolorbox}

\begin{tcolorbox}[colback=black!5!white,colframe=black!75!black,title=VL Agent - UI Operations]
\#\# Continuously take action until the task is completed.\\
*FUNCTION*: The action to take, should be one of \{Click,Type,Swipe,Back,Home,Enter,Done\} \\
*ARGS*: The input of the action. \\
*RESULT*: Action results. \\
*RETURN*: Reply based on action results. 
\tcbsubtitle{User}
Your task is: What's a good restaurant in San Diego? \\![](Screenshot\_1)
\tcbsubtitle{Assistant}
The screen shows the email setup page, which is not relevant to finding a restaurant. Possible actions are to exit the em ail setup and navigate to a search engine or map application where I can look up restaurants in San Diego.
*FUNCTION*: Home *ARGS*: \{\} \\
*RESULT*: ![](Screenshot\_2) \\
*RETURN*: The screen shows the home screen with various apps, including Google Chrome and Maps which could be used to search for a restaurant in San Diego. Possible actions are to tap on the Google Chrome app to open it and then use it to search for a good restaurant in San Diego.\\
*FUNCTION*: Click *ARGS*: \{"point": "(687, 792)"\} \\
*RESULT*: ![](Screenshot\_4) \\
*RETURN*: The current screen shows search results for skateboards, which is not related to the query about finding a good restaurant in San Diego. Possible actions are to click on the Google search bar at the top of the screen and enter the correct search term "good restaurant in San Diego" to find relevant information. \\
*FUNCTION*: Click *ARGS*: \{"point": " (596, 213)"\} \\
*RESULT*: ![](Screenshot\_6)  \\
*RETURN*: The current screen shows search suggestions related to skateboard speeds, which is not relevant to the query about finding a good restaurant in San Diego. Possible actions are to click on the "X" button to he correct query for finding a good restaurant in San Diego. \\
*FUNCTION*: Click *ARGS*: \{"point": "(927, 135)"\} \\
*RESULT*: ![](Screenshot\_8)  \\
*RETURN*: The screen shows the Google search bar with a list of previous searches, none of which are related to finding a restauran t in San Diego. Possible actions are to type "good restaurant in San Diego" into the search bar and then press enter to perform the search. \\
*FUNCTION*: Type *ARGS*: \{"text": "What's a good restaurant in San Diego?"\} \\
*RESULT*: ![](Screenshot\_9)  \\
*RETURN*: The screen shows a list of search suggestions related to finding good restaurants in San Diego, indicating that the user has typed in a query but not yet executed it. Possible actions are to select one of the suggested searches or press enter on the keyboard to execute the current search and view the results. \\
*FUNCTION*: Enter *ARGS*: \{\} \\
*RESULT*: ![](Screenshot\_10)  \\
*RETURN*: The screen shows a list of top-rated restaurants in San Diego with details such as ratings, price range, and opening hours. Possible actions are to stop and set the query as completed. \\
*FUNCTION*: Done *ARGS*: \{\}

\label{agent_ui_operation}
\end{tcolorbox}
\captionof{figure}{Qwen2-VL as an agent understands the query with respect to UI operation, utilizes the pre-defined actions in system message, and fulfill the task step-by-step.}

\begin{tcolorbox}[colback=black!5!white,colframe=black!75!black,title=VL Agent - Card Game]
% \tcbsubtitle{}
% \begin{center}
\includegraphics[width=0.8\textwidth]{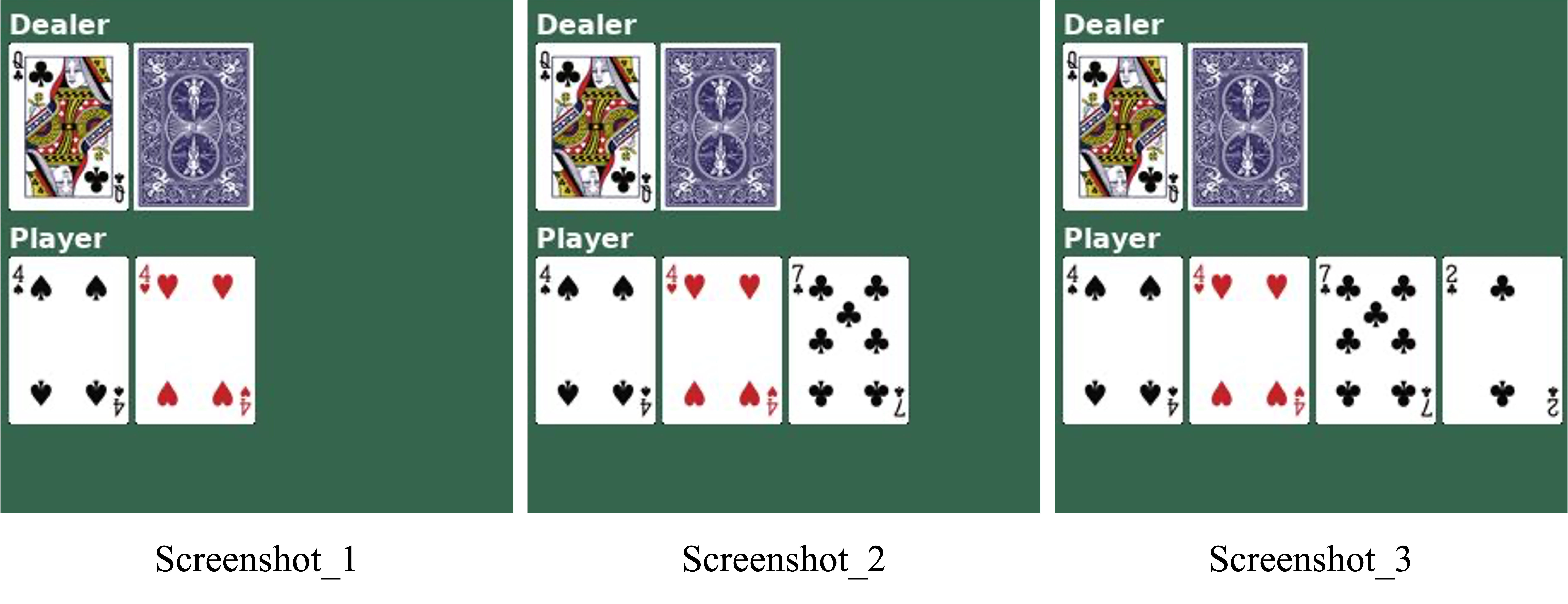}
%\end{center}
\tcbsubtitle{System Message}
You are a helpful assistant. \\
\# Actions \\
\#\# You have the following actions. \\
\#\#\# Hit \\
Hit: Get a new card. Parameters: [] \\
\#\#\# Stand \\
Stand: Stop to get a new card. Parameters: [] \\
\#\# Continuously take action until the task is completed.\\
*FUNCTION*: The action to take, should be one of \{Hit,Stand\} \\
*ARGS*: The input of the action. \\
*RESULT*: Action results. \\
*RETURN*: Reply based on action results. 

\tcbsubtitle{User}
You are a blackjack player. You are observing the current game state in the image. With the given functions, your goal is to dealer by getting as close to 21 points as possible without exceeding it.\\![](Screenshot\_1)

\tcbsubtitle{Assistant}
I have 8 points, and the dealer has 10 points. I should hit. \\
*FUNCTION*: Hit *ARGS*: \{\} \\
*RESULT*: \\![](Screenshot\_2) \\
*RETURN*: I have 15 points, and the dealer has 10 points. I should hit. \\
*FUNCTION*: Hit *ARGS*: \{\} \\
*RESULT*: \\![](Screenshot\_3) \\
*RETURN*: I have 17 points, and the dealer has 10 points. I should stand. \\
*FUNCTION*: Stand *ARGS*: \{\} \\
*RESULT*: Dealer Card: Q, 5, Q. The total point of dealer is 25 > 21. \\
Player Card: 4, 4, 7, 2. The total point of player is 17. \\
Player win!
\label{agent_card_game}
\end{tcolorbox}
\captionof{figure}{Qwen2-VL recognizes these cards and utilizes Hit and Stand to play the blackjack.}

\end{document}